%% file: main.tex
\documentclass{article}

\usepackage{microtype}
\usepackage{graphicx}
\usepackage{booktabs} % for professional tables
\usepackage{multirow}
\usepackage{amssymb, amsmath, mathtools}
\usepackage{adjustbox} 
\usepackage{comment}
\usepackage[percent]{overpic}

\DeclareMathOperator*{\argmax}{arg\,max}
\DeclareMathOperator*{\argmin}{arg\,min}
\DeclareMathOperator{\E}{\mathop{\mathbb{E}}}
\DeclareMathOperator{\R}{\mathop{\mathbb{R}}}

\DeclarePairedDelimiter\ceil{\lceil}{\rceil}
\DeclarePairedDelimiter\floor{\lfloor}{\rfloor}

\usepackage{hyperref}

\usepackage{caption}
\usepackage{subcaption}

\usepackage[accepted]{icml2022}

\icmltitlerunning{Stabilizing Off-Policy Deep Reinforcement Learning from Pixels}

\begin{document}

\twocolumn[
\icmltitle{Stabilizing Off-Policy Deep Reinforcement Learning from Pixels}



\icmlsetsymbol{equal}{*}

\begin{icmlauthorlist}
\icmlauthor{Edoardo Cetin}{equal,kcl}
\icmlauthor{Philip J. Ball}{equal,ox}
\icmlauthor{Steve Roberts}{ox}
\icmlauthor{Oya Celiktutan}{kcl}

% \icmlauthor{Cieua Vvvvv}{goo}
% \icmlauthor{Iaesut Saoeu}{ed}
% \icmlauthor{Fiuea Rrrr}{to}
% \icmlauthor{Tateu H.~Yasehe}{ed,to,goo}
% \icmlauthor{Aaoeu Iasoh}{goo}
% \icmlauthor{Buiui Eueu}{ed}
% \icmlauthor{Aeuia Zzzz}{ed}
% \icmlauthor{Bieea C.~Yyyy}{to,goo}
% \icmlauthor{Teoau Xxxx}{ed}
% \icmlauthor{Eee Pppp}{ed}
\end{icmlauthorlist}

\icmlaffiliation{kcl}{Centre for Robotics Research, Department of Engineering, King's College London}
\icmlaffiliation{ox}{Department of Engineering Science, University of Oxford}
% \icmlaffiliation{ed}{ED}
% \icmlaffiliation{goo}{Googol ShallowMind, New London, Michigan, USA}
% \icmlaffiliation{ed}{School of Computation, University of Edenborrow, Edenborrow, United Kingdom}

\icmlcorrespondingauthor{Edoardo Cetin}{edoardo.cetin@kcl.ac.uk}
\icmlcorrespondingauthor{Philip J. Ball}{ball@robots.ox.ac.uk}
%\icmlcorrespondingauthor{Cieua Vvvvv}{c.vvvvv@googol.com}
%\icmlcorrespondingauthor{Eee Pppp}{ep@eden.co.uk}

\icmlkeywords{Machine Learning, ICML}

\vskip 0.3in
]



\printAffiliationsAndNotice{\icmlEqualContribution} % otherwise use the standard text.

\begin{abstract}
% New version - tried to make it even more 'to the point'
Off-policy reinforcement learning (RL) from pixel observations is notoriously unstable. As a result, many successful algorithms must combine different domain-specific practices and auxiliary losses to learn meaningful behaviors in complex environments. In this work, we provide novel analysis demonstrating that these instabilities arise from performing temporal-difference learning with a convolutional encoder and low-magnitude rewards. We show that this new \textit{visual deadly triad} causes unstable training and premature convergence to degenerate solutions, a phenomenon we name \textit{catastrophic self-overfitting}. Based on our analysis, we propose A-LIX, a method providing adaptive regularization to the encoder's gradients that explicitly prevents the occurrence of \emph{catastrophic self-overfitting} using a dual objective. By applying A-LIX, we significantly outperform the prior state-of-the-art on the DeepMind Control and Atari 100k benchmarks without any data augmentation or auxiliary losses.

%Off-policy reinforcement learning (RL) from pixels is notoriously unstable. As a result, many successful algorithms must combine different heuristics and auxiliary losses in order to improve data-efficiency and learn meaningful behaviors in complex environments. In this work, we provide novel analysis explaining how instabilities arise in the common setting of temporal-difference learning with a convolutional encoder and low-magnitude rewards. Specifically, we show how a \textit{vision-based deadly triad} causes unstable training, resulting in premature convergence to poor solutions. We name this phenomenon \textit{catastrophic self-overfitting}, and illustrate how several popular reinforcement learning methods in fact already mitigate this, however they do so implicitly. To address this, we propose A-LIX, a new method that \emph{explicitly} prevents \emph{catastrophic self-overfitting} by adaptively regularizing the encoder's gradients using a dual objective. By applying A-LIX to existing architectures, we obtain state-of-the-art performance on both DeepMind Control and Atari benchmarks without any data augmentation or auxiliary losses.

\end{abstract}

\input{sections/1intro}
\input{sections/2background}
\input{sections/3analysis}
\input{sections/4method}
\input{sections/5experiments}
\input{sections/6related}
\input{sections/7conclusion}
% \newpage
\section*{Acknowledgments}
%Could improve?
Edoardo Cetin and Oya Celiktutan would like to acknowledge the support from the Engineering and Physical Sciences Research Council [EP/R513064/1] and LISI Project [EP/V010875/1]. Philip J. Ball would like to thank the Willowgrove Foundation for support and funding. Furthermore, support from Toyota Motor Corporation contributed towards funding the utilized computational resources.

\bibliography{main}
\bibliographystyle{icml2022}

\clearpage
\input{appendix}


\end{document}

%% file: sections/1intro.tex
\section{Introduction}

\label{sec:intro}

One of the core challenges in real world Reinforcement Learning (RL) is achieving stable training with sample-efficient algorithms \citep{challenges-rl}. Combining these properties with the ability to reason from visual observations has great implications for the application of RL to the real world \citep{qt-opt, rl_real_world_ingredients}. Recent works utilizing \textit{temporal-difference} (TD-) learning have made great progress advancing sample-efficiency \citep{ddpg, td3, sac, learningpessimism}. However, stability has remained a key issue for these off-policy algorithms \citep{suttonTDlearning, unstable_off, deadly_triad, RL_stability_practical}, making their general applicability limited as compared to their on-policy counterparts \citep{ppo, ppg}. At the same time, using pixel observations has been another orthogonal source of instabilities, with several successful approaches relying on pre-training instead of end-to-end learning \citep{deepspatial-ae, actionable-repr}. In fact, alternative optimization objectives, large amounts of simulation data, and symbolic observations have been common factors in most contemporary large-scale RL milestones \citep{alphazero, alphastar, OpenAI5}.

In this work, we provide novel insights behind why applying successful off-policy RL algorithms designed for proprioceptive tasks to pixel-based environments is generally underwhelming \citep{slac, sacae}. In particular, we provide evidence that three key elements strongly correlate with the occurrence of detrimental instabilities: \textit{i)} \emph{Exclusive} reliance on the \emph{TD-loss}. \textit{ii)} Unregularized end-to-end learning with a \textit{convolutional encoder}. \textit{iii) Low-magnitude sparse rewards}. Using this framework, we are able to motivate the effectiveness of auxiliary losses \citep{curl, spr-atari, sacae} and many domain-specific practices \citep{rainbowDQN, rad} by explaining how they address elements of this new \textit{visual deadly triad}.
\begin{figure}[t]
    \centering
    %\vspace{-1mm}
    \includegraphics[width=0.999\linewidth]{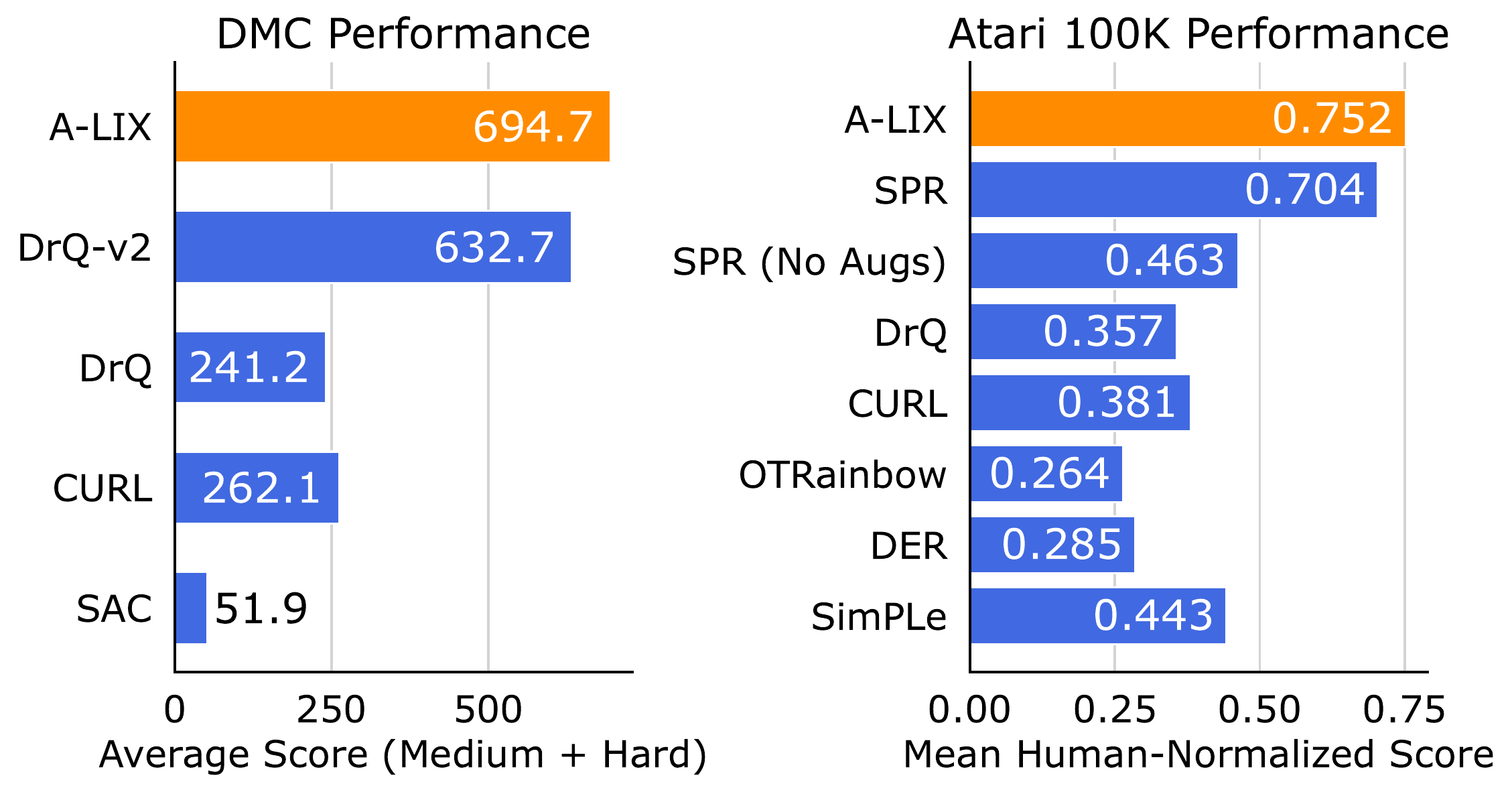} 
    \vspace{-7mm}
    %\vspace{-5mm}
    \caption{\small{{Performance of agents in DMC (\textbf{left}) and Atari 100k (\textbf{right}) benchmarks from 10 seeds. A-LIX outperforms previous methods without using image augmentations or auxiliary losses.}}}
    \vspace{-5mm}
    %\vspace{-4mm}
    \label{fig:mainperfplot}
\end{figure}%

We focus our analysis on the popular DeepMind Control suite (DMC) \citep{dmc}, where the introduction of random shift augmentations has played a key role in recent advances \citep{rad, drq, drqv2}. In this domain, we observe that the presence of the \textit{visual deadly triad} results in the TD-loss gradients through the convolutional encoder's feature maps having high spatial frequencies. We find these gradients are \textit{spatially inconsistent} and result in degenerate optimization landscapes when backpropagated to the encoder's parameters. Furthermore, repeatedly updating the convolutional encoder with these gradients consistently leads to early convergence to degenerate feature representations causing the critic to fit high-variance erroneous targets, a phenomenon we name \textit{catastrophic self-overfitting}. As a way of identifying the direct implications of the visual deadly triad in the gradient signal, we propose a new measure called the \textit{Normalized Discontinuity (ND)} score and show how its value precisely correlates with agent performance. Thus, we explain the effectiveness of shift augmentations by recognizing that they regularize the gradient signal by providing an \textit{implicit} spatial smoothing effect.

Based on our analysis, we propose \textbf{A}daptive \textbf{L}ocal S\textbf{I}gnal Mi\textbf{X}ing (A-LIX) a novel method to prevent catastrophic self-overfitting with two key components: \textit{i)} A new {parameterized} layer (LIX) that \textit{explicitly} enforces smooth feature map gradients. \textit{ii)} A dual objective that ensures learning stability by adapting the LIX parameters based on the estimated \textit{ND} scores. We show that integrating A-LIX with existing off-policy algorithms achieves state-of-the-art performance in both DeepMind Control and Atari 100k benchmarks without requiring image augmentations or auxiliary losses and significantly fewer heuristics. We open-source our code to facilitate reproducibility and future extensions\footnote{https://github.com/Aladoro/Stabilizing-Off-Policy-RL}.

\begin{table}[t]
\centering
%\vspace{-2mm}
\caption{\small{Practices from recent pixel-based TD-learning methods to mitigate elements of the visual deadly triad. $^\dagger$DrQ uses 10-step returns on Atari. *CURL uses 20-step returns on Atari.}}
\vskip 0.1in
\label{tab:deadlytriad}
\adjustbox{max width=\linewidth}{
\begin{tabular}{@{}lccc@{}}
\toprule
\multicolumn{1}{c}{\multirow{2}{*}{\textbf{Algorithm}}} & \multicolumn{3}{c}{\textbf{Visual Deadly Triad Mitigation}}                                     \\ \cmidrule(l){2-4} 
\multicolumn{1}{c}{}                           & TD-Loss            & CNN Overfit          & \multicolumn{1}{l}{Low-Density Reward} \\ \midrule
DrQ/RAD                                        & - & Shift/Jitter Augmentations                & 10-step returns$^\dagger$               \\
DrQ-v2                                         & -        & Shift Augmentations                & 3-step returns                         \\
SAC-AE                                         & VAE Loss  & -          & -                                      \\
SPR                                            & Model-Based Loss & Shift/Jitter Augmentations & 10-step returns                        \\
DER                                            & -  & Non-Overlapping Strides                & 20-step returns                        \\
CURL                                           & Contrastive Loss&  Shift Augmentations         & 20-step returns*                      \\
\bottomrule
\end{tabular}}
\vspace{-4mm}
\end{table}

Our main contribution can be summarized as follows:
\begin{itemize}
\itemsep0em 
    \item We conjecture the existence of a \textbf{\emph{visual deadly triad}} as a principal source of instability in reinforcement learning from pixel observations and provide clear empirical evidence validating our hypothesis.
    \item We show these instabilities affect the gradient signal causing \textbf{\emph{catastrophic self-overfitting}}, a phenomenon that can severely harm TD-learning. As a result, we design the \textbf{\emph{normalized discontinuity score}} to explicitly anticipate its occurrence.
    \item We propose \textbf{\emph{A-LIX}}, a new method that adaptively regularizes convolutional features to prevent catastrophic self-overfitting, achieving state-of-the-art results on two popular pixel-based RL benchmarks.
\end{itemize}

%% file: sections/2background.tex
\section{Background}

\label{sec:background}

We consider problem settings described by Markov Decision Processes (MDPs) \cite{mdp}, defined as the tuple $(S, A, P, p_0, r, \gamma)$. This comprises a state space $S$, an action space $A$, transitions dynamics given by $P$ and $p_0$, and a reward function $r$. The RL objective is then for an agent to recover an optimal policy $\pi^*$, yielding a distribution of trajectories $p_\pi(\tau)$ that maximizes its expected sum of discounted future rewards, $\pi^*=\argmax_\pi\E_{p_\pi(\tau)}\left[ \sum^{\infty}_{t = 0}\gamma^t r (s_t, a_t) \right]$. In off-policy RL, this objective is usually approached by learning a \textit{critic} function to evaluate the effectiveness of the agent's behavior. A common choice for the critic is to parameterize the policy's Q-function $Q^\pi: S \times A \rightarrow \mathbb{R}$, that quantifies the agent's performance after performing a particular action: $Q^\pi(s, a) = \E_{p_\pi(\tau|s_0{=}s, a_0{=}a)}\left[ \sum^{\infty}_{t = 0}\gamma^t r (s_t, a_t)\right]$. Most off-policy algorithms entail storing trajectories in a buffer $D$, and learning parameterized Q-functions by iteratively minimizing a squared temporal difference (TD-) loss:
\begin{equation}\label{q_fn_obj}
\begin{split}
    J_Q(\phi)=\E_{(s, a, s', r) \sim D}\left[(Q^\pi_\phi(s, a) - y)^2\right],\\
    y= r + \gamma\mathbb{E}_{a\sim \pi(s')}\left[\hat{Q}_{\phi'}^\pi(s', a)\right].
\end{split}
\end{equation}
Here, the TD-targets $y$ are computed from a 1-step bootstrap operation with a slowly-changing \emph{target Q-function} $\hat{Q}_{\phi'}^\pi$.
In continuous action spaces, we also learn a separate parameterized policy to exploit the information in the critic. This practically results in alternating TD-learning with maximizing the Q-function's expected return predictions, following the policy gradient theorem \citep{pg-thm}.

%% file: sections/3analysis.tex
\section{Instabilities in TD-Learning from Pixels}
\label{sec:analysis}

\begin{figure}[t]
    %\vspace{-2mm}
    \centering
    \includegraphics[width=0.8\linewidth]{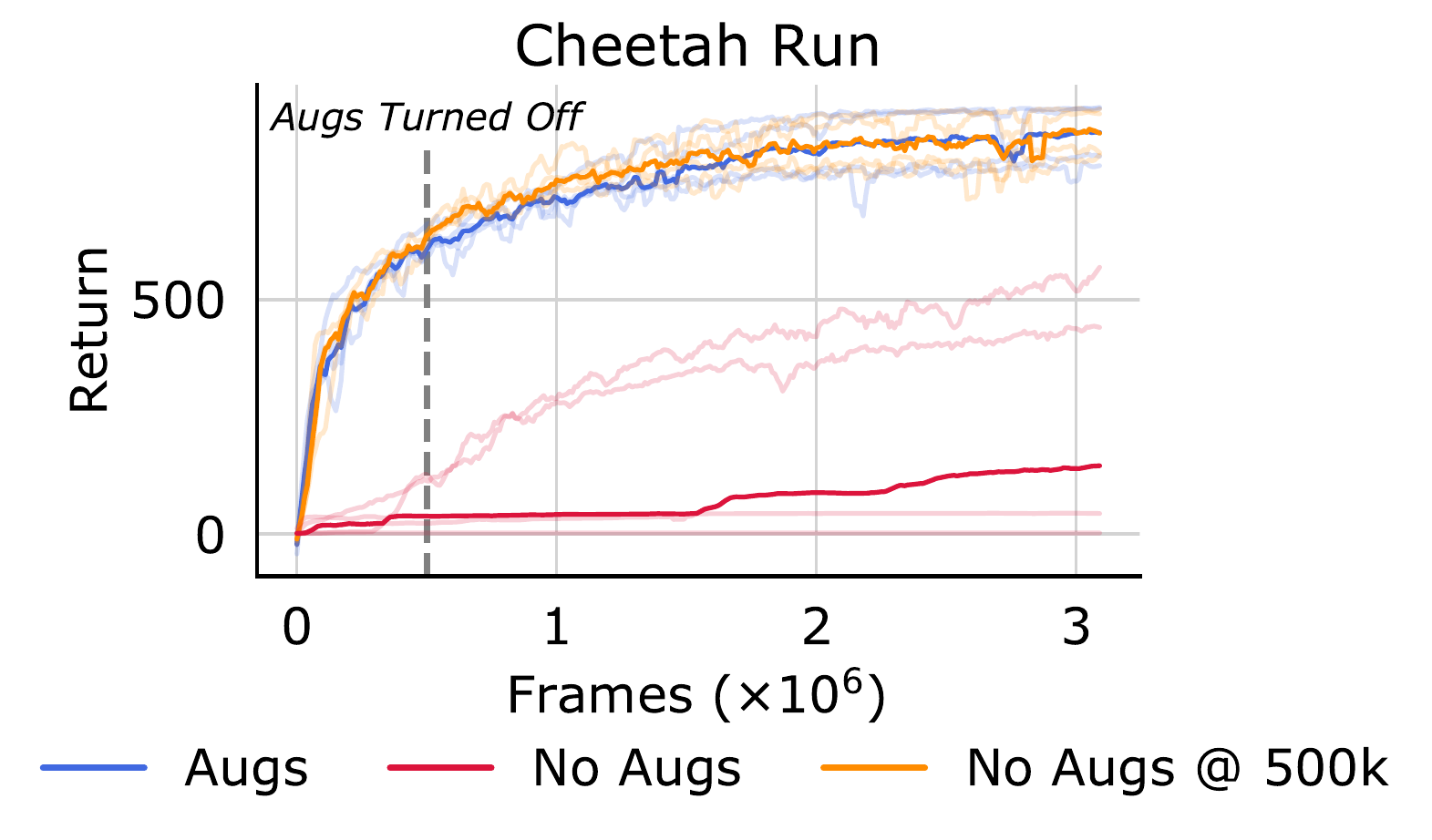} 
    %\vspace{-6.5mm}
    \vspace{-4mm}
    \caption{\small{{Returns of agents over 5 seeds. Solid lines represent median performance, faded lines represent individual runs. The vertical dashed line shows when augmentations are turned off.}}}
    \vspace{-4mm}
    \label{fig:turnoffaug}
\end{figure}%

Unlike proprioceptive observations, off-policy RL from pixel observations commonly requires additional domain-specific practices to ensure stability. In this section, we provide a novel analysis of this phenomenon by focusing on the DeepMind Control Suite \citep{dmc}. In this benchmark, the introduction of random shift data augmentations has been a core component of recent advances in pixel-based off-policy RL \citep{rad, drqv2}, allowing us to isolate and reproduce stable and unstable training regimes. Our analysis suggests the existence of specific elements that cause instabilities and strives to explain their implications on learning dynamics. We validate our findings via thorough empirical experimentation showing numerous results corroborating our hypotheses. Based on our discoveries, in Section~\ref{sec:method} we provide a new interpretation of random shifts and propose a new improved method to isolate and counteract instabilities.

\subsection{Why Do Augmentations Help?}

\begin{figure}[t]
    %\vspace{-2mm}
    \centering
    % \begin{subfigure}[t]{0.49\linewidth}
    %     \centering
    %     \includegraphics[width=0.999\linewidth]{figs/analysis_plots/qvals.pdf} 
    %     %\vspace{-5mm}
    %     \vspace{-5mm}
    %     %\vspace{-6.5mm}
    %     \caption{\small{{Shaded are individual estimates, solid is median.}}}
    %     %\vspace{15pt}
    %     %\vspace{-4mm}
    %     \label{fig:qvals}
    % \end{subfigure}
    % \hfill
    % \begin{subfigure}[t]{0.49\linewidth}
    %     \centering
    %     \includegraphics[width=0.999\linewidth]{figs/analysis_plots/overfit.pdf} 
    %     %\vspace{-5mm}
    %     %\vspace{-6.5mm}
    %     \vspace{-5mm}
    %     \caption{\small{{Q-values Pearson correlation with TD-loss components.}}}
    %     %\vspace{-2mm}
    %     %\vspace{-4mm}
    %     \label{fig:pearson}
    % \end{subfigure}
    % \begin{subfigure}[t]{0.99\linewidth}
        \centering
        \includegraphics[width=0.999\linewidth]{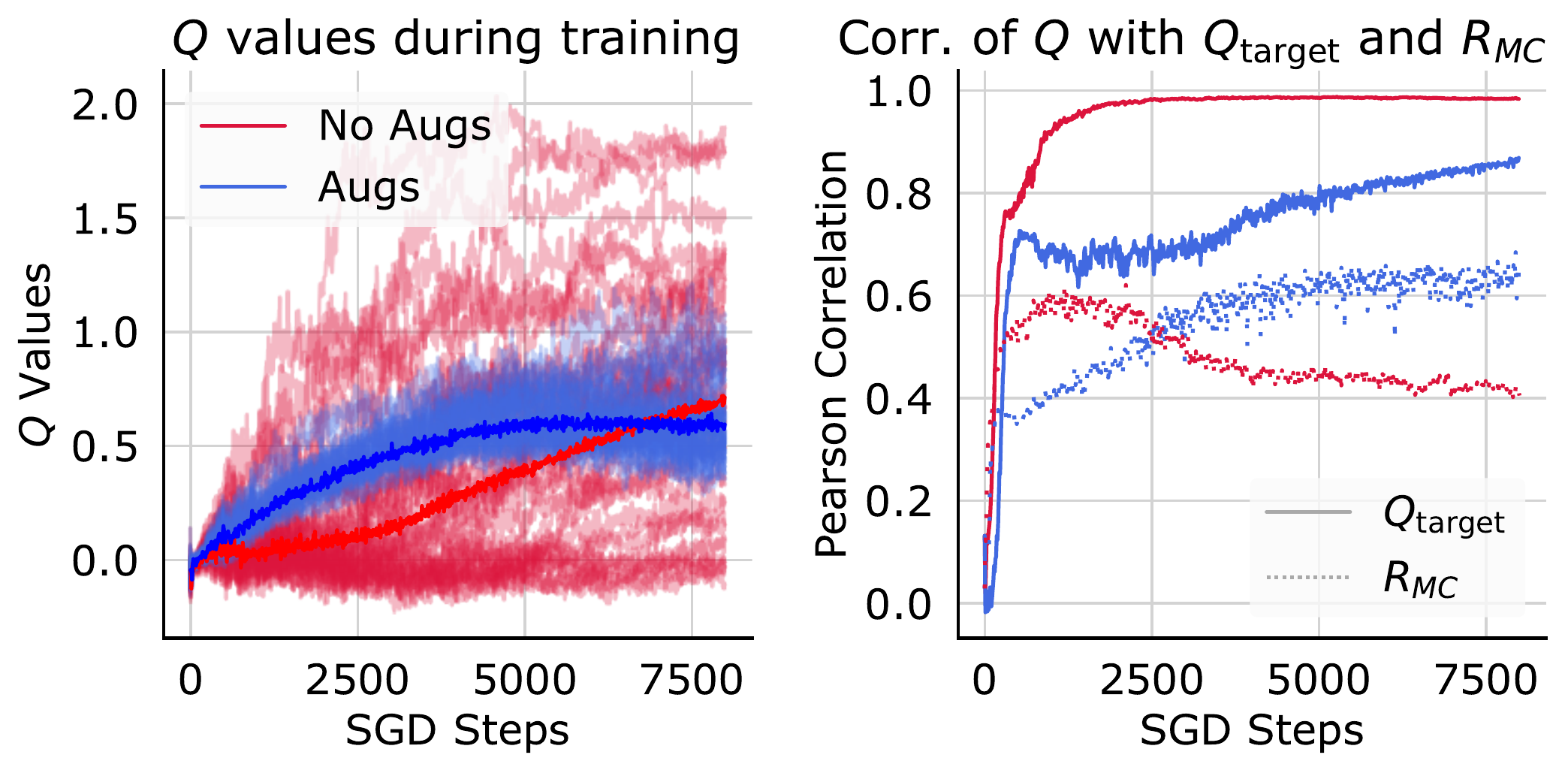} 
        %\vspace{-5mm}
        \vspace{-5mm}
        %\vspace{-6.5mm}
        % \caption{\small{{Shaded are individual estimates, solid is median.}}}
        %\vspace{15pt}
        %\vspace{-4mm}
    \vspace{-2mm}
    \caption{\small{Evidence of overfitting when augmentations are not used. On the \textbf{left}, shaded lines are individual estimates, the solid line represents the median Q-value. On the \textbf{right}, the Q-values Pearson correlation with target values and Monte-Carlo returns ($R_{MC}$).}}
    \label{fig:qvals_pearson}
    %\vspace{-6mm}
    \vspace{-4mm}
\end{figure}%

The underlying mechanism behind the effectiveness of random shifts is not immediately clear. While this augmentation may appear to assist generalization by encoding an invariance \cite{shorten2019augsurvey}, we note that all the environments from DMC employ a camera that is fixed relative to the agent's position. Hence, robustness to shifts does not appear to introduce any useful inductive bias about the underlying tasks. Moreover, prior work successfully learned effective controllers without augmentations~\citep{dreamer,sacae}, suggesting that shift generalization might not be the primary benefit of this method. We analyze the effect of random shifts by training a DrQ-v2 agent \citep{drqv2} on Cheetah Run but turning off augmentations after an initial 500,000 time-steps learning phase. As shown in Fig.\ \ref{fig:turnoffaug}, while training without any shift augmentation fails to make consistent progress, turning off augmentations after the initial learning phase actually appears to slightly improve the performance of DrQ-v2. This result is a clear indication that augmentations are not needed for asymptotic performance, and are most helpful to counteract instabilities present in the \emph{earlier stages of learning}, which we now focus on analyzing (see App.~\ref{sec:behaviorcloning}-\ref{sec:turnoffaug} for complementary experiments validating these claims).

\subsection{Identifying a New Deadly Triad}

\begin{figure}[t]
    %\vspace{-2mm}
    \centering
    \includegraphics[width=0.999\linewidth]{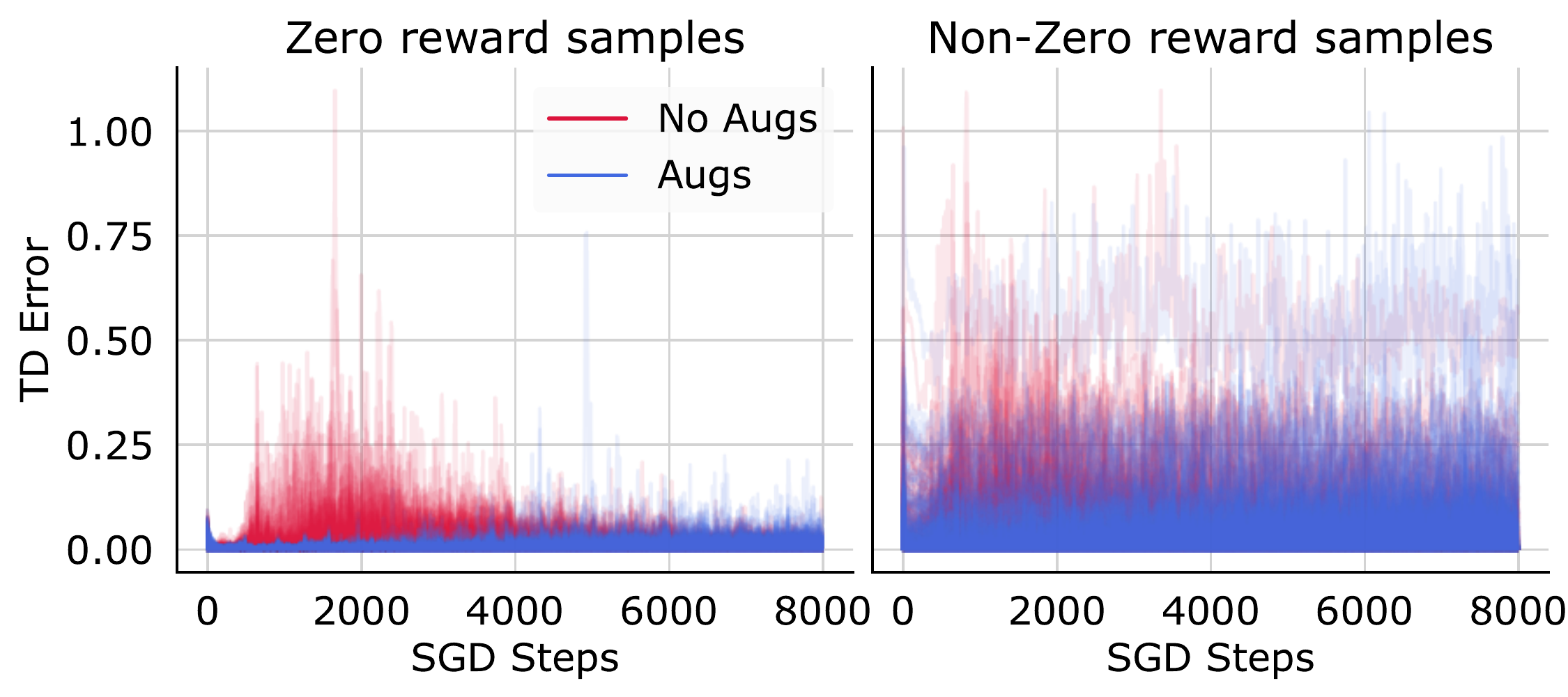} 
    %\vspace{-9.5mm}
    \vspace{-5mm}
    \caption{\small{{TD-loss of offline fixed transitions during training, separated based on having non-zero reward.}}}
    \vspace{-4mm}
    \label{fig:mses}
\end{figure}%

To reduce confounding factors and to disentangle the origin of these instabilities, we design an offline RL experiment \cite{offlinerl_survey}. This experiment isolates three distinct elements affecting off-policy RL: exploration, policy evaluation, and policy improvement. First, we gather a set of 15,000 transitions with pixel observations using a random policy in Cheetah Run. This allows us to ground exploration and analyze learning from fixed data resembling the early stages of online training (when augmentations appear most helpful). We then isolate policy evaluation by training both critic and encoder using SARSA \cite{sarsa} until convergence on this fixed data. Finally, we run policy improvement, training an actor to maximize the expected discounted return as predicted by the converged critic (see App.\ \ref{sec:offlinedescrip} for details). Interestingly, we find that \textit{turning on} augmentations exclusively during exploration or policy improvement has no apparent effect on stability and final performance.
Hence, we focus on the effects that augmentations have on TD-learning and analyze applying augmentations {only during policy evaluation}.

\begin{table}[H]
\centering
\vspace{-4mm}
\caption{\small{Performance and training statistics of different agent types in the offline experiments from 15,000 random transitions.}} %over 4 seeds.}}
\vskip 0.1in
\setlength{\tabcolsep}{3pt}
\label{tab:offline}
\adjustbox{max width=\linewidth}{
\begin{tabular}[b]{lccc}
    \toprule
    \textbf{Agent} & \textbf{Final TD-Loss} & \textbf{Final Policy Loss} & \textbf{Return}\\ \midrule
    Augmented & $0.021$ & $-0.99$ & $86.5 \pm 11.3$\\
    Non-Augmented & $0.002$ & $-1.05$& $9.2 \pm 12.1$\\
    \cmidrule(lr){1-4}
    Proprioceptive & $0.012$ & $-1.14$ & $79.1 \pm 7.7$\\
    Frozen CNN (random) & $0.023$ & $-0.95$ & $43.6 \pm 20.2$\\
    Frozen CNN (pre-trained) & $0.012$ & $-0.99$ & $77.6 \pm 18.5$\\
    \cmidrule(lr){1-4}
    Non-Augmented (norm $r$) & $18.616$ & $3.86$ & $38.6\pm 16.5$\\
    Non-Augmented (10-step returns) & $0.003$ & $-1.24$& $36.5 \pm 20.3$\\
    \bottomrule
\end{tabular}}
    \vspace{-5mm}
\end{table}

As shown in Table \ref{tab:offline}, applying augmentations during policy evaluation enables us to learn policies that achieve a return of 86.5, despite the best trajectory in the offline data achieving only 10.8. In contrast, without augmentations we consistently recover near 0 returns, resembling the failures observed in the online experiments. On the left of Fig.\ \ref{fig:qvals_pearson} we show the evolution of the predicted Q-values for both agents on a fixed batch of offline data. In particular, when performing policy evaluation without augmentations, these predictions display extremely high variance across different state-action pairs. In Table~\ref{tab:offline} we further show that the non-augmented agent displays {significantly} lower loss, despite having higher average Q-values than the augmented agent \citep{schaul2021returnbased}. We argue this is a clear indication of the occurrence of \emph{overfitting}. We corroborate our claim by analyzing the evolution of the Pearson bi-variate correlation between the estimated Q-values and target Q-values on the right of Fig.\ \ref{fig:qvals_pearson}. These results show that the non-augmented agent displays near-perfect correlation with its own target Q-values throughout training, indicating that it immediately \emph{learns to fit its own noisy, randomly-initialized predictions}. We also record the correlation with the actual discounted Monte-Carlo returns, which represent the \textit{true targets} the Q-values should ideally approximate during policy evaluation. For these results, we observe that the relationship between applying augmentations and the recorded correlation is reversed, with the non-augmented agent displaying significantly lower correlation. This dichotomy appears to indicate that fitting the noisy targets severely affects learning the \textit{useful} training signal from the collected transitions regarding the experienced rewards. We confirm this phenomenon by splitting the data into non-zero and zero reward transitions, where the only learning signal propagated in the TD-loss is from the initially random target values. In Fig.\ \ref{fig:mses} we illustrate that the non-augmented agents initially experience much higher TD-errors on zero reward transitions, confirming that they focus on fitting uninformative components of the TD-objective.

In Table~\ref{tab:offline} we provide the results of additional experiments that indicate that TD-learning is not the \textit{only} cause for the observed instabilities.
First, we confirm that the observed overfitting appears to be exclusive to performing \textit{end-to-end} TD-learning with convolutional neural network (CNN) encoders. Concretely, we run the same offline experiment without \textit{training an encoder} in three different settings. First, we consider performing policy evaluation directly from non-augmented \textit{proprioceptive} observations with a fully-connected critic network. Moreover, we also consider \textit{freezing} the encoder weights either to their initial \textit{random values} or to \textit{pre-trained} values from the augmented agent experiments. In all three cases, we attain largely superior performance, almost matching the augmented agent's performance for both the proprioceptive and pre-trained experiments. In addition, we also find that the observed overfitting phenomenon is diminished when simply increasing the magnitude of the reward signal in the TD-loss. We test this through two additional experiments which consider \textit{normalizing} the collected rewards before policy evaluation and incorporating \textit{large n-step returns}~\cite{suttonTDlearning}. As reported, both modifications considerably improve the non-augmented agent's performance. However, we note that both practices introduce further unwanted variance in the optimization, failing to yield the same improvements as augmentations (see App.~\ref{sec:nstep}). 

Taken together, our results appear to strongly indicate that instabilities in off-policy RL from pixel observation come from three key conditions, which we refer to as the \emph{visual deadly triad}: i) Exclusive {reliance on the TD-loss}; ii) Unregularized learning with an {expressive convolutional encoder}; iii) Initial {low-magnitude sparse rewards}. Further evidence arises when considering the ubiquity of particular practices employed in pixel-based off-policy RL. In particular, as summarized in Table \ref{tab:deadlytriad}, most popular prior algorithms feature design choices that appear to counteract at least two elements of this triad, either directly or implicitly. Furthermore, we show these instabilities result in the non-augmented critics {focusing on learning their own noisy predictions}, rather than the actual experienced returns. We observe this ultimately leads to convergence to erroneous and high-variance Q-value predictions, a phenomenon we name \textit{catastrophic self-overfitting}.

\subsection{Anticipating Catastrophic Self-Overfitting}
\label{sec:anticipating}
\begin{figure}[t]
    %\vspace{-2.5mm}
    \centering
    \includegraphics[width=0.9\linewidth]{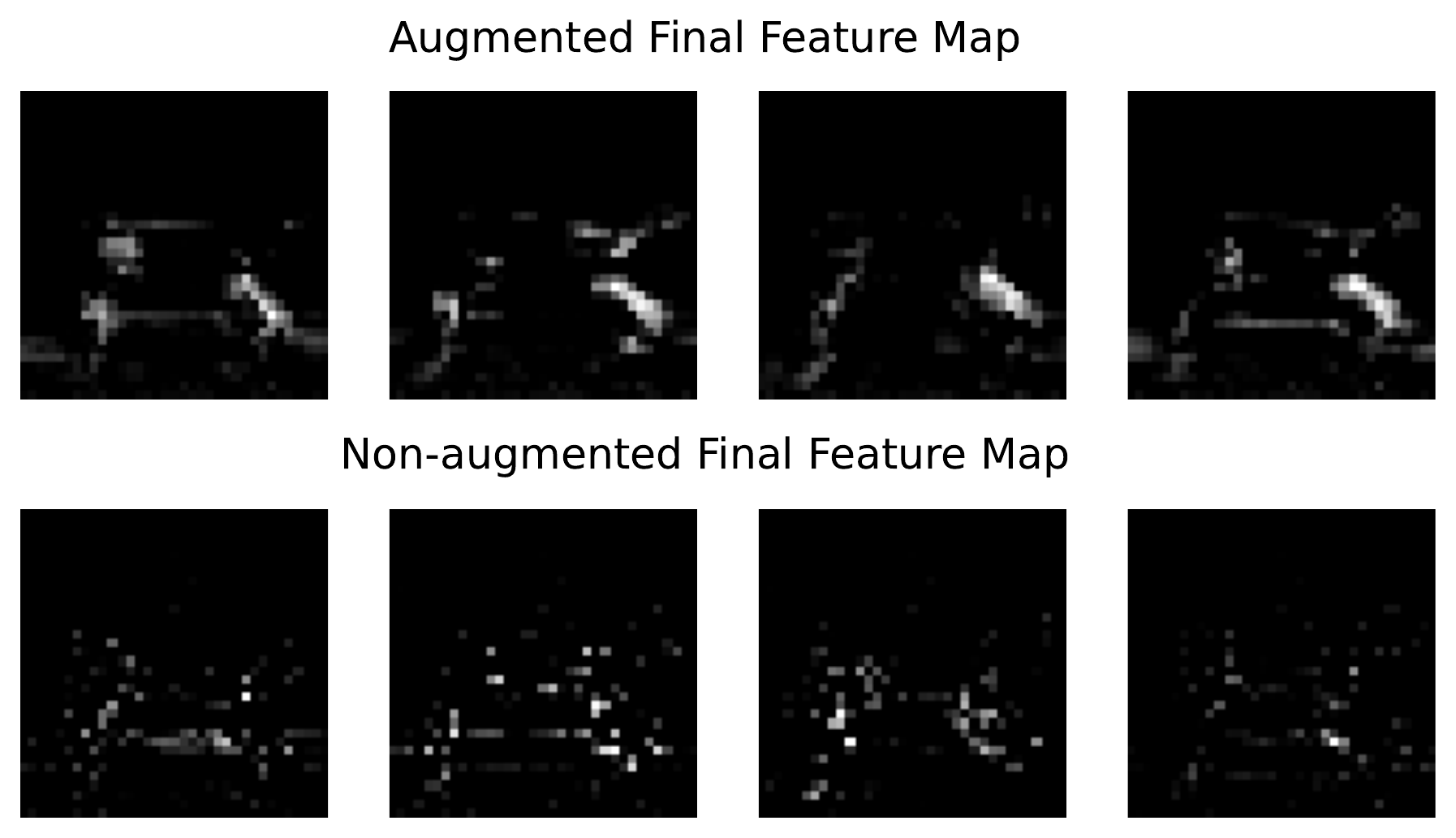} 
    \vspace{-2mm}
    \caption{\small{{Feature maps in the final layer of both augmented (\textbf{top}) and non-augmented (\textbf{bottom}) agent encoders. Non-augmented agents manifest inconsistent, high-frequency feature maps.}}}
    \label{fig:feature_maps}
    \vspace{-4mm}
\end{figure}%
% To understand overfitting in the CNN, we perform a sensitivity analysis over its final feature representation $z\in \R^{C\times H \times W}$. 
We now attempt to unravel the links that connect the \textit{visual deadly triad} with \textit{catastrophic self-overfitting}. We start by observing that catastrophic self-overfitting comes with a significant reduction of the critic's sensitivity to changes in action inputs, implying that the erroneous high-variance Q-value predictions arise primarily due to changes in the observations (see App.~\ref{sec:actvalsurf} for action-value surface plots). Hence, we focus on analyzing the feature representations of the pixel observations, computed by the convolutional encoder, $z\in \R^{C\times H \times W}$. In particular, we wish to quantify the \emph{sensitivity} of feature representations to small perturbations in the input observations. To measure this, we evaluate the Jacobians of the encoder across a fixed batch of offline data for the augmented and non-augmented agents. We then calculate the Frobenius norm of each agent's Jacobians, giving us a measure of how quickly the encoder feature representations are changing locally around an input (see App.~\ref{sec:jacobiananalysis} for details). Our results show a stark difference, with the feature representations of the non-augmented agents being on average \textbf{2.85} times more sensitive. This suggests that overfitting is driven by the CNN encoder's \textit{representations} learning \textit{high-frequency} information about the input observations and, thus, breaking useful inductive biases about this class of models \citep{cnnspectralbias}.

 \begin{figure}[t]
    %\vspace{-2mm}
    \centering
    \includegraphics[width=0.999\linewidth]{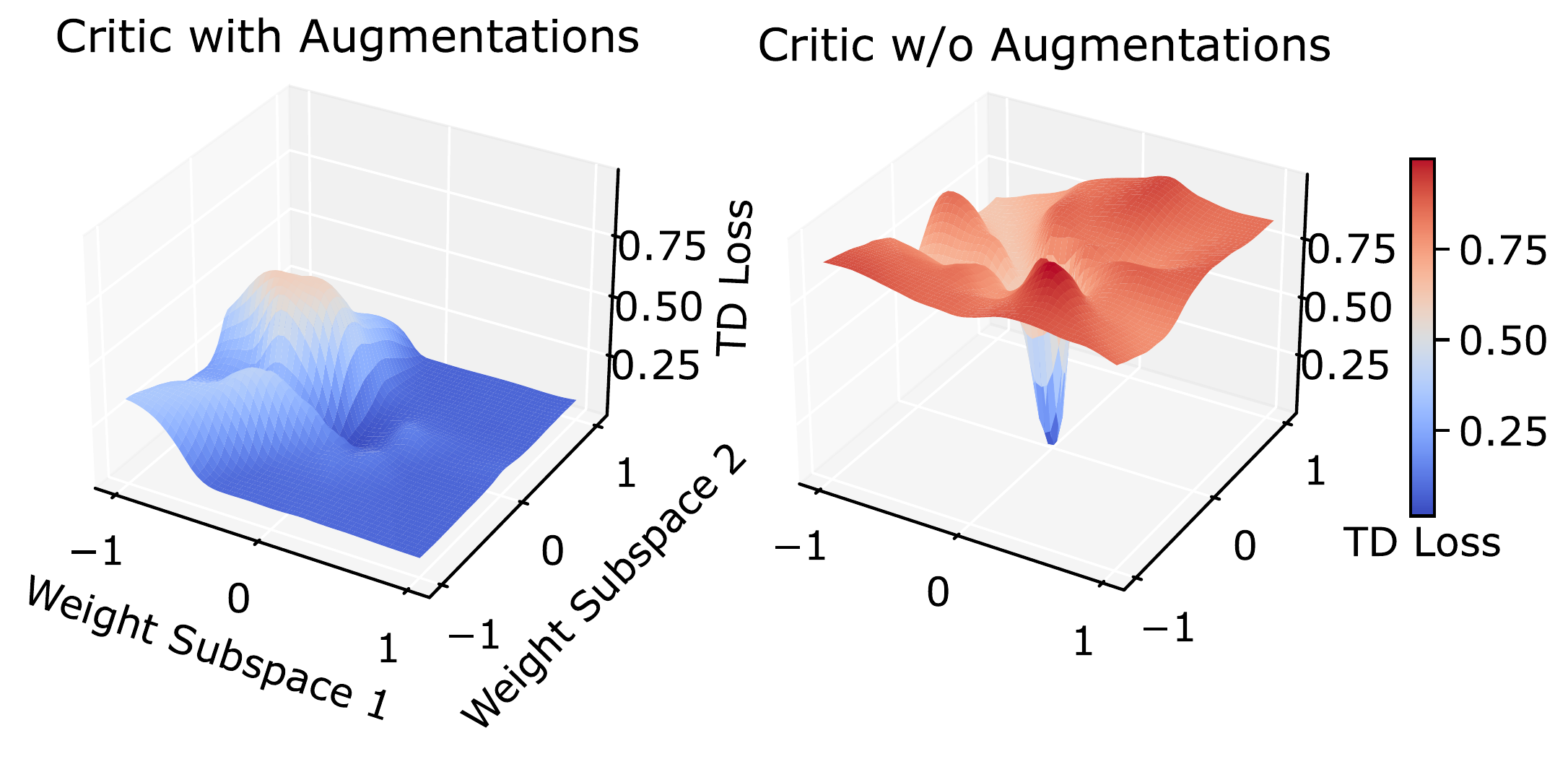} 
    \vspace{-6mm}
    \caption{\small{Critic loss surface plots of augmented (\textbf{left}) and non-augmented (\textbf{right}) agents after 5,000 steps of offline training.}}
    \label{fig:encoder_loss_surf}
    \vspace{-4mm}
\end{figure}%

In App.~\ref{sec:specnorm} we demonstrate that lower sensitivity to random noise, while desirable for optimization \cite{rosca2020a}, is actually a \textit{byproduct} of a stable feature representations, and not its defining factor. Furthermore, observing the actual feature maps of different observations in Fig.~\ref{fig:feature_maps}, we see that augmentations make the encoder produce features that are \emph{spatially consistent}, aligned with common understandings of how natural representations should appear \cite{mindThePad, allenzhu2021feature}. In contrast, the non-augmented agents display high-frequency and \textit{discontinuous} feature maps that do not reflect the spatial properties of their inputs. Hence, our evidence suggests that catastrophic self-overfitting specifically follows from the same learning process that produces highly-sensitive and discontinuous encoder feature maps. Therefore, we turn our focus to analyzing the \textit{gradients} backpropagated to the encoder's features maps and observe one key property: the gradients of the output feature maps \textit{consistently reflect the same spatial properties} of their resulting features. In particular, the \textit{gradients of the feature maps} appear \emph{spatially consistent} for the augmented agent, and \textit{discontinuous} for the non-augmented agent. This optimization property reflects intuitive understandings of backpropagation since discontinuous gradients should push the encoder's weights to encode discontinuous representations. To provide further complementary evidence that discontinuous gradients are the direct cause of catastrophic self-overfitting, we analyze the normalized loss surfaces \textit{when backpropagating these discontinuous gradients} to the encoder's parameters (following \citet{visualloss}). In Fig.\ \ref{fig:encoder_loss_surf}, we see that gradient discontinuities in the non-augmented agent yield extreme peaks in its encoder's loss surface, clearly suggestive of overfitting \citep{sharpLossBadGeneralization} \footnote{Instead, the loss surface with respect to the fully-connected weights is smoother (App.\ \ref{sec:mlpsurface}).}.
 
\begin{figure}[t]
    %\vspace{-2mm}
    \centering
    \includegraphics[width=0.999\linewidth]{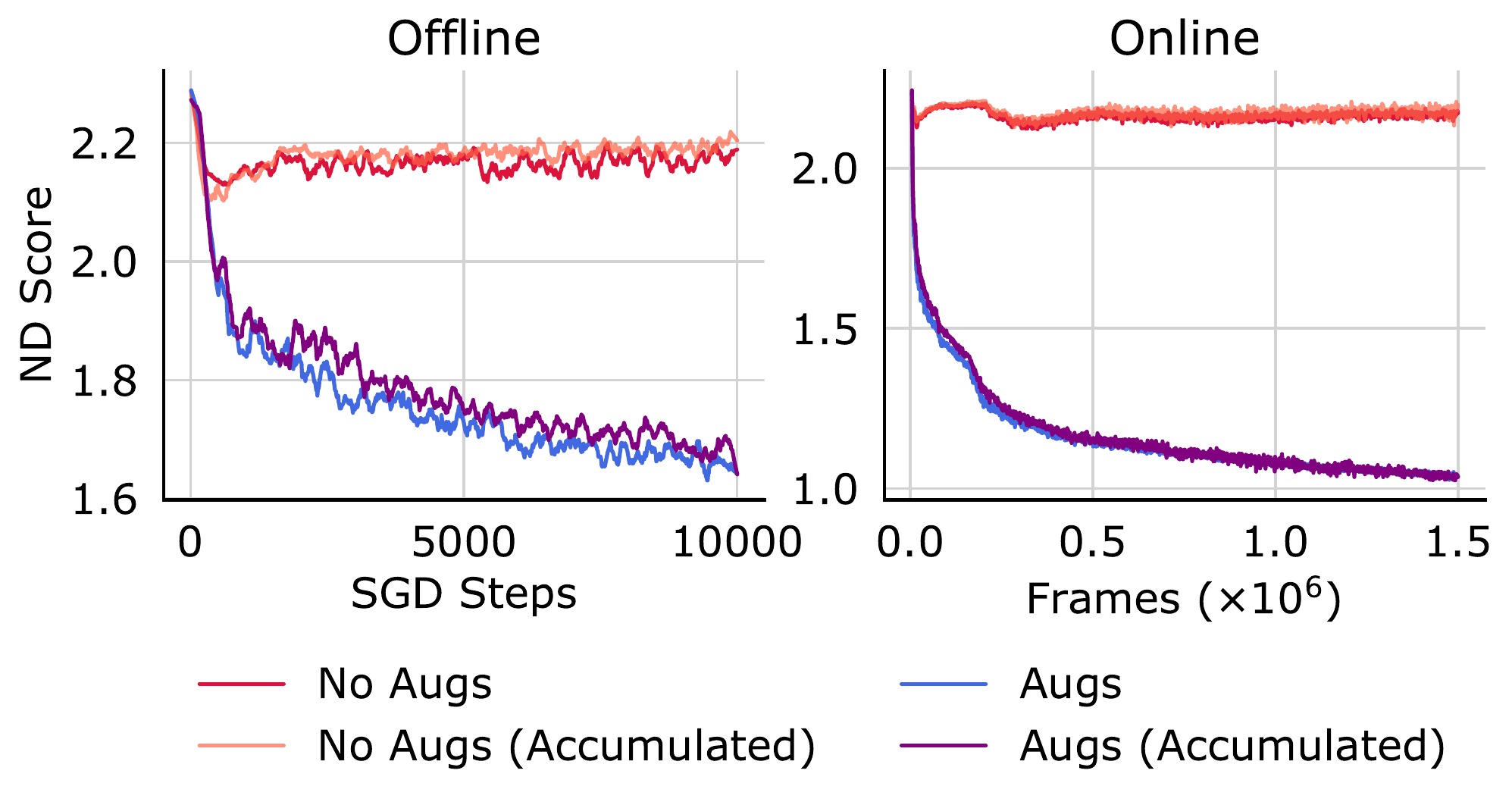} 
    \vspace{-7mm}
    \caption{\small{{Instantaneous (\textcolor{red}{red} and \textcolor{blue}{blue}) and accumulated (\textcolor{orange}{orange} and \textcolor{purple}{purple}) $ND$ scores for the features gradients from offline (\textbf{left}) and online (\textbf{right}) training in Cheetah Run.}}}% Augmented agents consistently score lower.}}}
    \vspace{-4mm}
    \label{fig:ndscore}
\end{figure}

 %but only with respect to \emph{the convolutional parameters}, and observe a signficiant difference in their respective flatness\footnote{Loss with respect to the MLP weights is smoother (App.\ \ref{sec:mlpsurface}).}; these extreme peaks in the non-augmented agent are again suggestive of over-fitting \citep{sharpLossBadGeneralization}. 
 To quantify the level of discontinuity in the features and their gradients, we propose a new metric that encodes the aggregated immediate spatial `unevenness' of each feature location within its relative feature map. In particular, we define $D(z)\in \mathbb{R}^{C\times H\times W}$ as the expected squared local discontinuity of z in any spatial direction, i.e.:
\begin{equation}
D(z)_{ijc} \approx \E_{v\sim S^1}\left[\left(\frac{\partial z_{ijc}}{\partial v}\right)^2\right],
\end{equation}
% In practice, $D(z)$ can be estimated by computing the average square difference between $z$ and its immediate neighbors.
% We provide further discussion of this measure together with connections to the discrete Laplacian filter in App~\edo{app}.
practically estimated via sampling. We then normalize each value in $D(z)$ by its squared input and average over all the feature positions. We name this metric the \textit{normalized discontinuity (ND)} score:
\begin{equation}
\begin{split}
\textit{ND}(z) = \frac{1}{C\times H\times W}\sum_{c=1}^C\sum_{j=1}^H\sum_{i=1}^W \frac{D(z)_{ijc}}{z_{ijc}^2}.
\end{split}
\end{equation}
Intuitively, this score reflects how locally discontinuous $z$ is expected to be at any spatial location. 
In Fig.~\ref{fig:ndscore}, we show how the $ND$ score of $\nabla z$ evolves during training in the offline and an online setting for both augmented and non-augmented agents. We see that augmented agents experience considerably less discontinuous gradients through their features, and that recordings of lower $ND$ scores also appear to be highly correlated with performance improvements. We additionally show an accumulated $ND$ score, using an exponential moving average of $\nabla z$ in each spatial position to calculate this metric. Interestingly, we observe that the $ND$ score over accumulated gradients is almost identical to the instantaneous $ND$ score, showing that similar gradient discontinuities are propagated \emph{persistently} through training in each position of the feature map. This property confirms that the discontinuities are not smoothed by the stochastic sampling of different consecutive training batches, in which case we would expect to observe lower accumulated $ND$ scores. Thus, it suggests that self-overfitting emerges in the non-augmented agents due to repeated gradient steps towards \emph{persistent} feature map discontinuities.

%% file: sections/4method.tex
\section{Counteracting Gradient Discontinuities}
\label{sec:method}

\subsection{Gradient Smoothing and Random Shifts}
\label{sec:rand_shifts}

As analyzed in Section~\ref{sec:analysis}, catastrophic self-overfitting occurs when the gradients in the convolution layers are locally discontinuous.
As a result, we argue that the efficacy of random shifts arises from their downstream effect on feature gradient computation, which counteracts these discontinuities during backpropagation. In particular, while random shifts do not act directly on the latent representation or their respective gradients, they do affect \textit{how} the latent representations are computed. This has an impact on how \emph{persistent} discontinuous components of the gradient are backpropagated to the encoder's parameters during learning. From the approximate shift invariance of convolutional layers, we can view a convolutional encoder as computing each of the feature vectors $\left[z_{1ij}, ..., z_{Cij}\right]^T$ with the same parameterized function, $V_\phi$, that takes as input a subset of each observation $O\in{\R^{C'\times H' \times W'}}$. This subset corresponds to a local neighborhood around some reference input coordinates $i', j'$. Thus, the only factor differentiating features in the same feature map (e.g., $z_{cij}$ and $z_{ckl}$) is some implicit function $f(i, j)=i', j'$ translating each of the output features coordinate into the relative reference input coordinate, 
i.e. $z_{cij}= V_\phi(O, i', j')_c$ (determined by kernel sizes, strides...). Therefore, random shifts are approximately equivalent to further translating each reference coordinate by adding some uniform random variables $\delta'_x, \delta'_y$:

\vspace{-4mm}
%\begin{small}
\begin{equation*}
\begin{split}
& z_{cij}\approx V_\phi(O, i'+\delta'_x, j'+\delta'_y)_c, \\
\mathrm{where}\quad &\delta'_x, \delta'_y\sim U[-s', s'], \quad f(i, j)=i', j'.
\end{split}
\end{equation*}
%\end{small}
\vspace{-4mm}

Due to the employed strides from the convolutional architectures used in DrQ-v2~\citep{drqv2}, the difference in reference coordinates of adjacent features in a feature map is less than the maximum allowable shift employed in the augmentations, i.e., $(i+1)'-i', (j+1)'-j'< s'$ (where $s'$ is the maximum allowable shift). Consequently, shift augmentations effectively turn the deterministic computation graph of each feature $z_{cij}$ into a random variable, whose sample space comprises the computation graphs of all nearby features within its feature map. Hence, applying different random shifts to samples in a minibatch makes the gradient of each feature $\nabla z_{cij}$ backpropagate to a random computation graph, sampled from a set that extends the set of non-augmented computation graphs for all features in a local neighborhood of coordinates $i, j$.
Therefore, aggregating the parameter gradients produced with different $\delta'_x, \delta'_y$, provides a \emph{smoothing effect} on how each discontinuous component of $\nabla z$ affects learning, and prevents {persistent} discontinuities from accumulating. Hence, random shifts break the second condition of the visual deadly triad, by providing effective \emph{implicit} regularization of the convolutional encoder's learning process. At the same time, this minimally disrupts the information content of the resultant features, since discarded observation borders almost exclusively comprise background textures that are irrelevant for performing the task. This interpretation of random shifts aligns with the analysis in Section~\ref{sec:analysis}, showing that implicitly smoothing over the backpropagated gradient maps consistently prevents catastrophic self-overfitting. 

\subsection{Local Signal Mixing}

 We extrapolate our hypotheses about catastrophic self-overfitting and random shifts by proposing a technique that aims to enforce gradient smoothing regularization \emph{explicitly}. We propose \textbf{L}ocal S\textbf{I}gnal Mi\textbf{X}ing, or \textbf{LIX}, a new layer specifically designed to prevent catastrophic self-overfitting in convolutional reinforcement learning architectures. LIX acts on the features produced by the convolutional encoder, $z\in \R^{C\times H \times W}$, by randomly mixing each component $z_{cij}$ with its neighbors belonging to the same feature map. Hence, LIX outputs a new latent representation with the same dimensionality $\hat{z}\in \R^{C\times H \times W}$, whose computation graph minimally disrupts the information content of each feature $z_{cij}$ while smoothing discontinuous components of the gradient signal during backpropagation.

%The persistent discontinuous component of the gradient signal analyzed in Section~\ref{sec:anticipating} is highly entangled with a consistent, useful component that would enable TD-learning to progress and escape its initial unstable regime. 
% Thus, acting directly on the gradients would be impractical, causing potential unexpected consequences on the optimization objectives \phil{this seems to high-level (e.g., `potential', `unexpected'; is there a more concrete statement, perhaps a citation}.
LIX is a regularization layer that acts as a simple random smoothing operation, reducing the expected magnitude of gradient discontinuities by preventing higher frequency signals to persist. In the forward pass, LIX produces a new latent representation where for each of the $C$ feature maps, $\hat{z}_{cij}$ is computed as a randomly weighted average of its spatial neighbors around coordinates $i, j$. We further \emph{parameterize} this stochastic operation using some maximum range radius $S$, and consequently sample two uniform \textit{continuous} random variables $\delta_x, \delta_y \sim U[-S, S]$, representing shifts in the $x$ and $y$ coordinates respectively. Correspondingly, we define $\tilde{i}=i+\delta_x$ and $\tilde{j}=j+\delta_y$ and perform the weighted averaging as a bilinear interpolation with weights determined by the random shifts:

\vspace{-4mm}
\begin{footnotesize}
\begin{equation*}
\begin{split}
\hat{z}_{cij} = &
z_{c\floor{\tilde{i}}\floor{\tilde{j}}}  (\ceil{\tilde{i}} - \tilde{i})(\ceil{\tilde{j}} - \tilde{j}) + 
z_{c\floor{\tilde{i}}\ceil{\tilde{j}}}  (\ceil{\tilde{i}} - \tilde{i})(\tilde{j} - \floor{\tilde{j}})  \\
+ & z_{c\ceil{i}\floor{\tilde{j}}}  (\tilde{i} - \floor{\tilde{i}})(\ceil{\tilde{j}} - \tilde{j}) +
z_{c\ceil{i}\ceil{\tilde{j}}}  (\tilde{i} - \floor{\tilde{i}})(\tilde{j} - \floor{\tilde{j}}).
\end{split}
\end{equation*}
\vspace{-4mm}
\end{footnotesize}

%Since nearby features in a convolutional feature map are computed with very similar receptive fields and their values tend to vary smoothly (as shown in Section~\edo{sec}), LIX should only marginally corrupt the latent signal and the relative information about the original observations. 
Since nearby features in a convolutional feature map are computed with very similar receptive fields, the mixing effect of LIX should have a trivial effect on the information the encoder can convey in its latent representations. In addition, LIX should have a direct regularization effect on the gradients by acting on the feature maps themselves. In particular, since LIX computes each output feature from a weighted average of its neighbors, backpropagation will split each gradient $\nabla \hat{z}_{cij}$, to a random local combination of features within the same feature map, $\{\nabla z_{c\floor{\tilde{i}}\floor{\tilde{j}}},\nabla z_{c\floor{\tilde{i}}\ceil{\tilde{j}}},\nabla z_{c\ceil{i}\floor{\tilde{j}}},\nabla z_{c\ceil{i}\ceil{\tilde{j}}}\}$. Thus, LIX should mostly preserve the consistent component of $\nabla z$, while randomly smoothing its discontinuous component.%, preventing the detrimental effects of its repeated application to the encoder's parameters. \phil{repetitious; we've made this point lots of times that it's to smooth out the persistent discontinuities, i think the reader knows now they're persistent/repeated.}

There are multiple key differences between the regularization from LIX and random shifts. LIX provides a local smoothing effect over the gradients explicitly and exactly, without having to deal with the implications of padding and strided convolutions breaking shift-invariance assumptions. Moreover, LIX smooths the gradient signal not only across different inputs but also within each feature map. In addition, by applying its operation solely at the feature level, the encoder can still learn to entirely circumvent LIX's smoothing effect on the information encoded in the latent representations, given enough capacity. This means that LIX does not forcibly preclude any input information from affecting the computation. Consequently, LIX also does not have to enforce learning invariances which might not necessarily reflect useful inductive biases about the distribution of observations. In contrast, random shifts need to exploit the particular uninformativeness of the observations borders to avoid disrupting the features' information content.

%\edo{be sure to show that the discontinuous component persists throughout batches.}

%\edo{be sure to mention in analysis that its the relative magnitude of the discontinuities to the useful gradient signal.}

\subsection{A-LIX}
\label{sec:alix}

\begin{figure}[t]
    %\vspace{-2mm}
    \centering
    \includegraphics[width=0.99\linewidth]{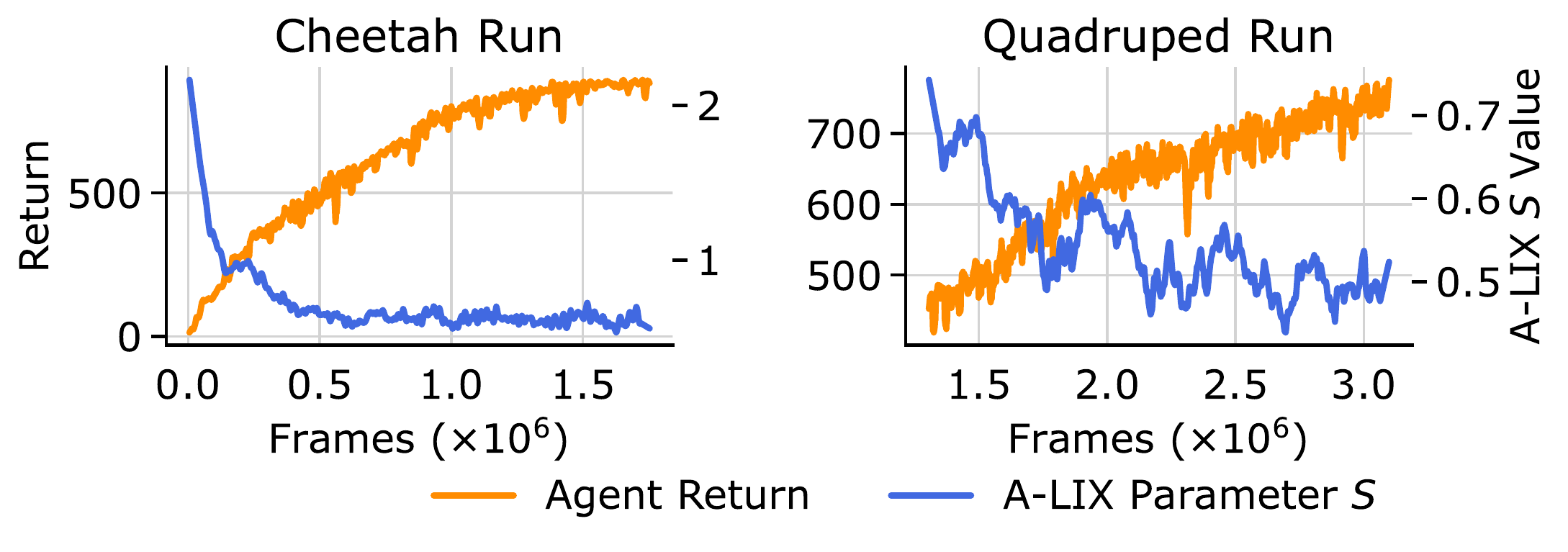}
    \vspace{-4mm}
    \caption{\small{{A-LIX's $S$ parameter evolution during training in Cheetah Run (\textbf{left}) and Quadruped Run (\textbf{right}). As the critic targets become more informative, $S$ falls, improving data efficiency and asymptotic performance.}}}
    \vspace{-4mm}
    \label{fig:adaptive_both}
\end{figure}

LIX introduces a single key parameter: the range radius $S$ used for sampling $\delta_x$ and $\delta_y$. Intuitively, this value should reflect how much we expect gradients to be locally consistent for a given architecture and task. 
Therefore, we argue that the value of $S$ should ideally decrease throughout training as the useful learning signal from the TD-loss becomes stronger. This is consistent with the results illustrated in Figure~\ref{fig:turnoffaug}, showing that turning off random shift augmentations after the TD-targets become informative can improve learning. Hence, we propose an adaptive strategy to learn $S$ throughout training. Utilizing the \textit{normalized discontinuity ($ND$)} score in Section~\ref{sec:anticipating}, we set up a dual optimization objective to ensure a minimum value of local smoothness in the representation gradients, $\overline{ND}$. However, computing the $ND$ score of the gradient signal involves a ratio between potentially very small values. As a result, estimation of these values from a batch of gradient samples can lead to outliers having an extreme impact on this average measure, translating into large erroneous updates of $S$. To overcome this, we propose using a slightly modified version of the $ND$ score with increased robustness to outliers (see App.~\ref{sec:adaptive_dual_objective} for further details):
\begin{equation}
\label{eq:mod-nd}
\widetilde{ND}(\nabla \hat{z}) = \sum_{c=1}^C\sum_{j=1}^H\sum_{i=1}^W \log\left(1 + \frac{D(\nabla \hat{z})_{cij}}{\nabla \hat{z}_{ijc}^2}\right).
\end{equation}
% This alternative score is inspired by recordings of signal-to-noise ratio measurements, we provide further intuition and justification for this modification in Section~\edo{app} of the Appendix.
% We would like to remark that since we set up the optimization of $S$ with a dual objective, changes in the actual target value relating to some appropriate smoothness constraint are mostly irrelevant when considering the optimization's dynamics \phil{Don't think we need this here; can discuss in appendix instead}.
In practice, we set up a dual optimization objective similar to the automatic temperature adjustment from \citet{sac-alg}. This entails alternating the optimization of the TD-learning objectives described in Section~\ref{sec:background} with minimizing a dual objective loss:
\begin{equation}
\label{eq:dual-obj}
    \argmin_{S} -S\times \E_{\hat{z}}\left[\widetilde{ND}(\nabla \hat{z}) - \overline{ND}\right],
\end{equation}
approximating dual gradient descent \citep{convex}. Hence, we call this new layer \textbf{A}daptive \textbf{LIX} (A-LIX). In Fig.~\ref{fig:adaptive_both} we show that A-LIX effectively anneals $S$ as the agent escapes its unstable regimes, in line with our intuition. %, and provide pseudocode of its implementation in App.~\edo{sec}.

%% file: sections/5experiments.tex
\section{Performance Evaluation}

We evaluate the effectiveness of A-LIX in pixel-based reinforcement learning tasks in two popular and distinct domains featuring a diverse set of continuous and discrete control problems. We integrate A-LIX with existing popular algorithms and compare against current state-of-the-art model-free baselines. We provide further details of our integration and full hyperparameters in App.~\ref{sec:app_hyper}.
We also extend this section by providing more granular evaluation metrics in App.~\ref{app:full_res}. Furthermore, we provide ablation studies analyzing different components of A-LIX in App.~\ref{sec:app_abl}. 

\subsection{DeepMind Control Evaluation}

We first evaluate the effectiveness of A-LIX for pixel-based RL on continuous control tasks from the DeepMind Control Suite (DMC) \citep{dmc}. Concretely, we integrate A-LIX with the training procedure and network architecture from DrQ-v2 \citep{drqv2}, but \emph{without using image augmentations}. To show the generality of our method we \emph{do not modify} any of the environment-specific hyperparameters from DrQ-v2 and simply add our A-LIX layer after each encoder nonlinearity. For simplicity, we optimize a shared $S$ for all the A-LIX layers with the dual objective in Eq.~\ref{eq:dual-obj}. %, where we estimate the $ND$ score utilizing the gradients of the final encoder's representation.
Hence, this introduces a {single additional parameter} and negligible computational overhead. We compare A-LIX to DrQ-v2, which represents the current state-of-the-art on this benchmark. We also compare against three further baselines: the original DrQ \citep{drq}, which foregoes n-step returns and includes an entropy bonus; CURL \citep{curl}, which includes an auxiliary contrastive objective; an extension of SAC \citep{sac-alg} with the encoder from \citet{sacae}. These last three baselines have been performant on a prior DMC benchmark that considers fewer tasks with high action repeats, as described by~\citet{planet}. Instead, we evaluate on the more challenging `Medium' and `Hard' benchmarks from \citet{drqv2}, comprising 15 tasks with low action repeats.
% \phil{This algorithmic discussion is great, but I think somewhat gets in the way of the narrative; perhaps we can instead summarize the high level points, and delegate (and extend) this detailed discussion to the appendix; same for Atari. Basically it takes up quite a bit of space talking about \emph{other} algorithms, not \emph{our} algorithm.}

\begin{figure}[t]
    %\vspace{-2mm}
    \centering
    \includegraphics[width=0.999\columnwidth]{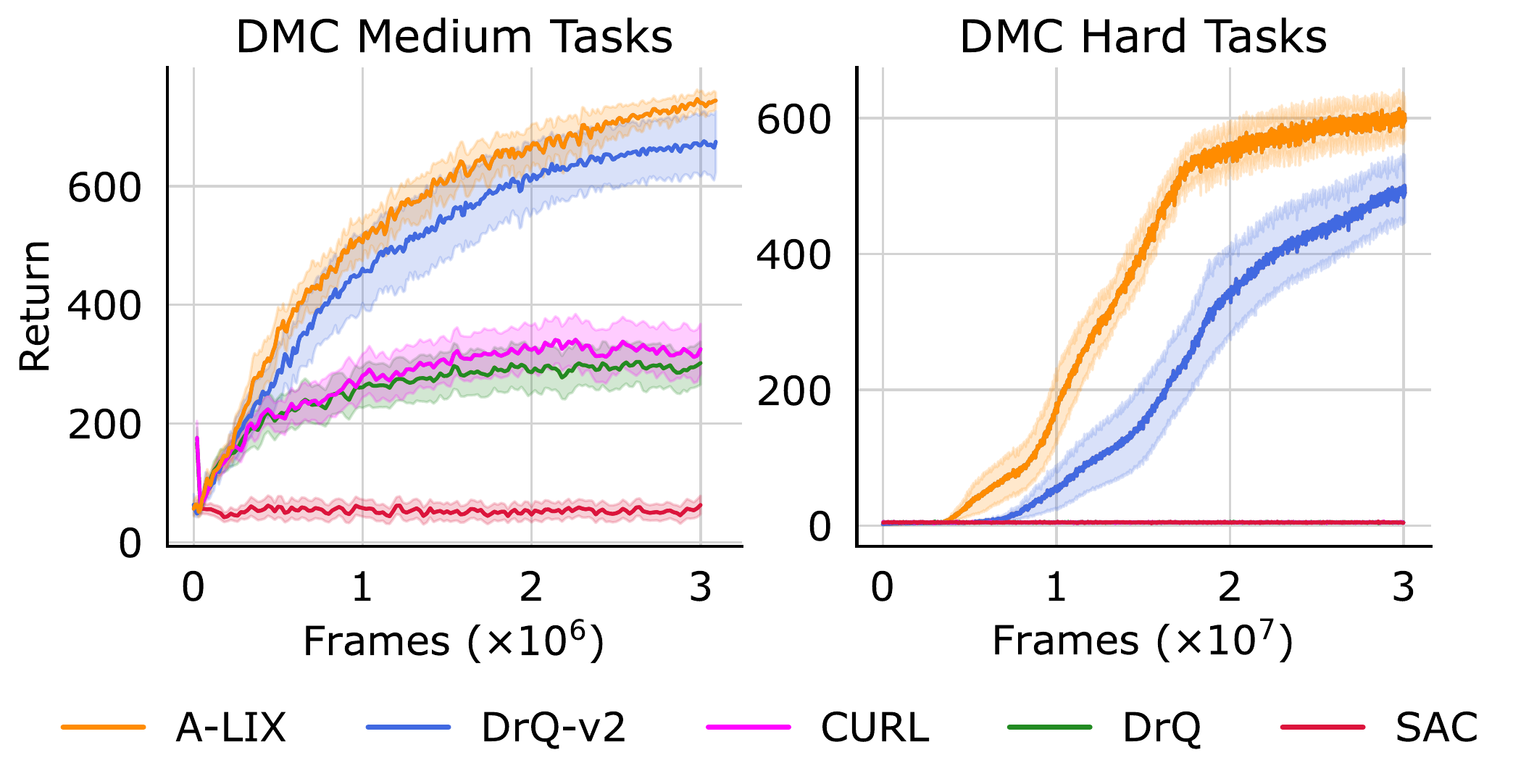}
    \vspace{-8mm}
    \caption{\small{{Average performance in 10 seeds for DMC Medium (\textbf{left}) and Hard tasks (\textbf{right}). Shaded regions represent $\pm1$ SE.}}}
    \vspace{-6mm}
    \label{fig:mean_perf_dmc}
\end{figure}%

\textbf{Results.} We summarize the results in Figure~\ref{fig:mean_perf_dmc}, showing the mean performance curves for both medium and hard benchmark tasks. We provide further details and the full list of results across all 15 environments in App.~\ref{app:dmc_full_res}. Overall, A-LIX surpasses all prior methods with clear margins, both in terms of efficiency and final performanc. This is particularly notable in the more complex `Hard' tasks. As highlighted in prior work \citep{learningpessimism}, DrQ-v2 appears to yield inconsistent results on some of the harder exploration tasks with sparse rewards. This likely indicates that the gradient regularization induced by random shifts (described in Section~\ref{sec:rand_shifts}) is unable to consistently prevent catastrophic self-overfitting in scenarios where the initial learning signal from TD-learning is particularly low. Finally, DrQ, CURL, and SAC fail to make consistent meaningful progress on this harder benchmark. This performance gap corroborates the third component of the visual deadly triad, showing how lower magnitude rewards due to harder exploration and lower action-repeats further destabilize TD-learning based algorithms, and explains the gains seen in DrQ-v2 when incorporating n-step returns. We believe these results emphasize the challenge of overcoming the visual deadly triad in continuous control problems and the particular effectiveness of A-LIX to counteract its direct implications.

\subsection{Atari 100k Evaluation}

\begin{table}[t]
\vspace{-3mm}
\caption{Results summary for the Atari 100k benchmark. %A-LIX achieves the highest normalized mean performance, comparable normalized median performance to SPR, and state-of-the-art results in 11 out of 26 individual Atari games.
 The reported performance of A-LIX is from 10 seeds.} 
% \vskip 0.1in
\label{tab:atari_perf}
\tabcolsep=0.04cm
%\vspace{-5mm}
\begin{center}
%\tiny
\adjustbox{max width=0.98\columnwidth}{
\begin{tabular}{@{}lccccccc@{}}
\toprule
\textit{\textbf{Metrics}}            & SimPLe & DER   & OTRainbow & CURL  & DrQ   & SPR            & \textbf{A-LIX} \\ \midrule
\textit{\textbf{Norm. Mean}}   & 0.443  & 0.285 & 0.264     & 0.381 & 0.357 & 0.704          & \textbf{0.753}        \\
\textit{\textbf{Norm. Median}} & 0.144  & 0.161 & 0.204     & 0.175 & 0.268 & \textbf{0.415} & 0.411                 \\ \midrule
\textit{\textbf{\# SOTA}}            & 7      & 1     & 1         & 1     & 1     & 4              & \textbf{11}           \\
\textit{\textbf{\# Super}}           & 2      & 2     & 1         & 2     & 2     & \textbf{7}     & \textbf{7}            \\ 
\textit{\textbf{Average Rank}}      & 3.92                & 5.00              & 5.21            & 3.92              & 4.85               & 2.88      & \textbf{2.21}            \\ \bottomrule
\end{tabular}}
\end{center}
\vspace{-5mm}
\end{table}

We perform a second set of experiments in an entirely different setting, discrete control. We make use of the popular Atari Learning Environment (ALE) \citep{atariALE} and consider the 100k evaluation benchmark from \citet{simple}. In particular, this benchmark comprises evaluating performance for 26 tasks after only two hours of play-time (100k interaction steps), following the evaluation protocol in \citet{machadoALEprotocol}.
We integrate A-LIX with Data-Efficient Rainbow (DER) \citep{der-rainbow}, a simple extension to Rainbow \citep{rainbowDQN} with improved data-efficiency. We would like to note that our integration has key differences to DER, designed to highlight the generality of our method in tackling the visual deadly triad. In particular, we reduce the n-step returns to 3 (from 20), and we maintain the \emph{same} encoder architecture as in DrQ-v2. To speak to the latter point, this means we do not require the highly regularized encoders with large convolutional filters and strides, used ubiquitously in off-policy learning for Atari environments. Instead, to stabilize learning we simply apply our A-LIX layer after the final encoder nonlinearity. We compare against three algorithms that, like A-LIX, do not employ data-augmentation: Data-Efficient Rainbow (DER); Overtrained Rainbow (OTRainbow) \citep{otrainbow}; and Simulated Policy Learning (SimPLe) \citep{simple} (model-based). %, an alternative extension of Rainbow for data-efficiency.
Moreover, we also compare with additional state-of-the-art off-policy baselines that make use of data augmentations: the aforementioned CURL and DrQ; and Self-Predictive Representations (SPR) \citep{spr-atari}, the current state-of-the-art TD-learning based algorithm on this benchmark. SPR combines data augmentation with numerous additional algorithmic design choices, such as an auxiliary self-supervised loss for learning a latent dynamics model.% from the encoder's representations.%

\begin{figure*}[t]
%\label{fig:rliable_res}
    %\vspace{-1mm}
    \centering
    \begin{subfigure}[t]{0.49\linewidth}
        \begin{overpic}[width=0.99\linewidth]{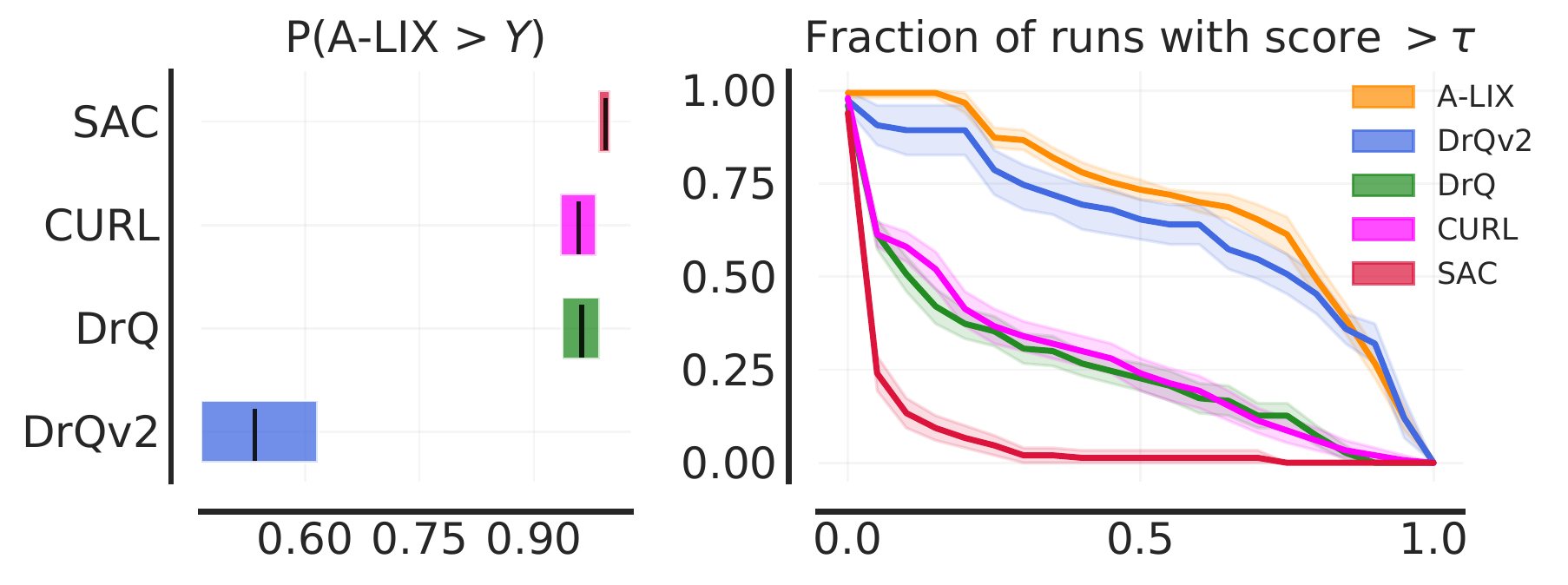}
            % \put (32,20) {\textcolor{red}{\textit{WIP: Need 5 seeds for Hard}}}
        \end{overpic}
        \vspace{-2mm}
        \caption{\small{DeepMind Control: Medium and Hard Tasks}}
        \label{fig:dmcmainrliable}
    \end{subfigure}
    % \vspace{2mm}
    \begin{subfigure}[t]{0.49\linewidth}
        \includegraphics[width=0.95\linewidth]{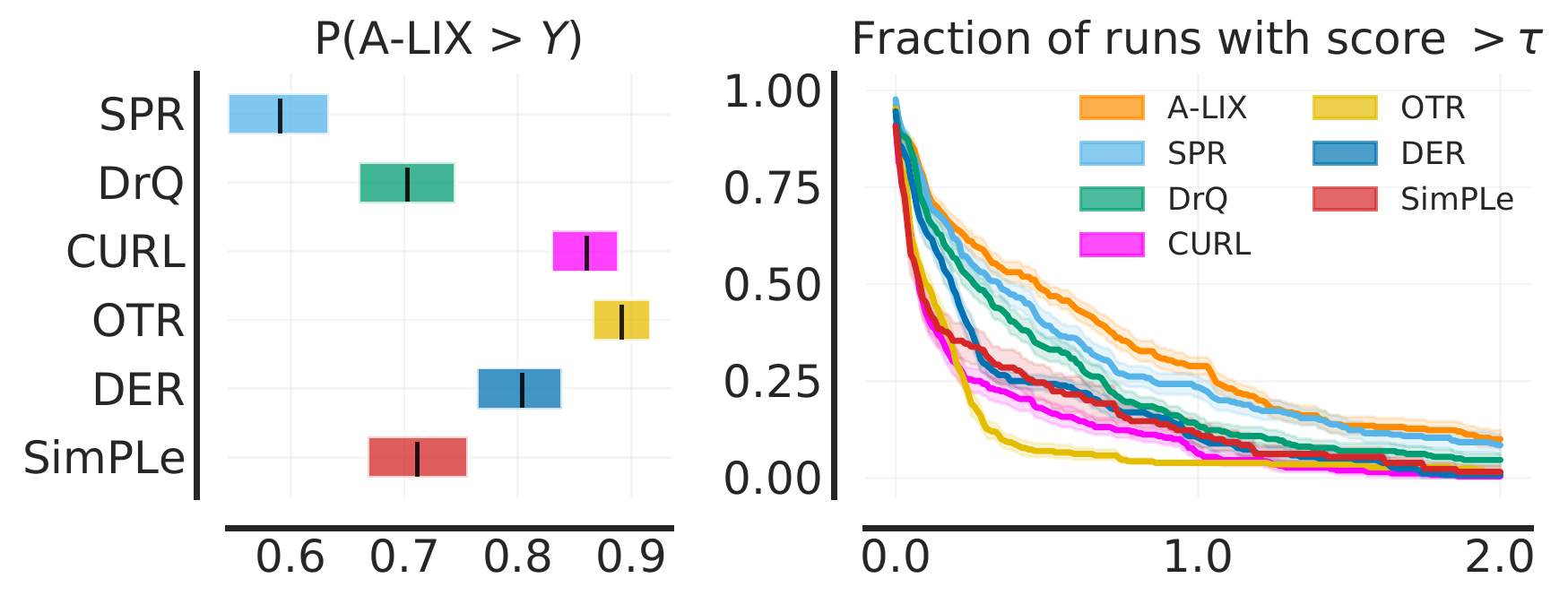}
        \vspace{-2mm}
        \caption{\small{Atari 100k}}
        \label{fig:atarimainrliable}
    \end{subfigure}
    %\vspace{-2.5mm}
    \caption{\small{{\emph{Probability of improvement} and \emph{performance profiles} obtained from the recorded results in DMC (\textbf{left}) and Atari 100k (\textbf{right}). A-LIX displays statistically significantly improvements and stochastically dominates most prior algorithms.}}}
    \vspace{-4mm}
    \label{fig:rliable_res}
\end{figure*}%

\textbf{Results.} We summarize the results in Table~\ref{tab:atari_perf}, showing the mean and median human-normalized scores together with the number of environments where each algorithm either achieves state-of-the-art or super-human performance. We include the full per-environment results in App.~\ref{app:atari_full_res}. Remarkably, A-LIX obtains a substantially higher human-normalized mean performance than all other considered algorithms. While the recorded normalized median performance is slightly inferior to SPR, we argue that such difference is not particularly significant since this metric depends on the performance obtained in just two environments. Moreover, A-LIX achieves super-human performance in 7 games (the same as SPR), and state-of-the-art performance in 11 games, considerably more than all other algorithms. These results corroborate how tuned architectures, data augmentation, and auxiliary losses used on ALE mostly serve the purpose of counteracting the direct implications of the visual deadly triad and show that A-LIX enables us to learn powerful models without relying on these design choices.

\subsection{Statistical Significance}

To validate the significance of our improvements, we statistically analyze our results using the \textit{Rliable} tools and practices from \citet{agarwal2021deep}. We summarize some of our key findings in Fig.~\ref{fig:rliable_res}, showing the \textit{probability of improvements} of A-LIX over prior methods (computed with the Mann-Whitney U statistic \citep{mannwhitneyUstat}) and the relative normalized \textit{performance profiles} \citep{perf-profile}. The ranges correspond to 95\% \textit{stratified bootstrap confidence intervals} \citep{bootstrap-cis}. In both DMC and Atari benchmarks, we find that our improvements are statistically significant (lower confidence intervals $>$0.5) and observe `stochastic dominance' of our algorithm against almost all considered baselines \citep{stoch-dom}. %These results 
We provide further results and details of the employed statistical analysis in App.~\ref{app:dmc_full_res} and App.~\ref{app:rliableinfo} respectively.

%% file: sections/6related.tex
\section{Related Work}

\label{sec:related}

% Our work focuses on model-free pixel-based methods, and in particular recent developments in continuous control \citep{curl, sacae, rad, drq, yarats2021drqv2, drac}.
% We also focus on generalization, in particular as a result of augmentations \citep{pitis2020counterfactual, ball2021augwm, sinha2021s4rl}.

% Recent image based RL (e.g., DrQ, RAD, CURL, Edo's work) and Augmentation based (e.g., AugWM, S4RL). LET'S SELF CITE!
Previous works have characterized several optimization issues related to performing RL via TD-learning \citep{gradient-TD-Baird, gradient-TD-Baird-2}. In this work, we instead focus on the \textit{empirical analysis} of modern TD-learning algorithms, specific to the pixel-based RL setting.  
We also observe links with recent work studying \textit{observational overfitting} \citep{Song2020Observational}. Our work differs by focusing on memorization effects particular to the combination of CNNs and TD-learning. There are also connections with existing feature-level augmentation work, such as Dropout \citep{dropout} and DropBlock \citep{dropblock}. In particular, the latter also applies structured transformations directly to the feature maps and introduces a heuristic to adjust this transformation over training, validating our findings on the utility of adaptivity. Outside RL, there is a rich body of work on implicit regularization and memorization in CNNs \cite{sharpLossBadGeneralization, exploringgeneralization, memorizationCNNs, badglobalminimaexist, randomlabels}. \citet{cnnspectralbias} show that higher frequency data manifolds cause CNNs to learn higher spectral frequency terms, aligning with our analysis of higher frequency representations. \citet{coherentgrads} show generalization arises when similar examples induce similar gradients during learning (i.e., coherence). Their work supports our findings since inconsistent feature gradients are a manifestation of non-coherence, explaining their poor generalization. Finally, our dual objective falls under automatic tuning methods in RL (AutoRL) \cite{parkerholder2022automated}. These approaches have been applied very successfully to manage non-stationary trade-offs, such as exploration and exploitation \cite{rp1} and optimism \cite{moskovitz2021tactical, learningpessimism}. Finally, we note links with recent work concerning implicit regularization in TD-learning \citep{kumar2021implicit}. However, while \citet{kumar2021implicit} observe an implicit `underfitting' phenomenon in later training stages, we analyze an opposed `overfitting' phenomenon occurring during the first training steps, which we find to be specific to learning from visual inputs.

%% file: sections/7conclusion.tex
\section{Conclusion}

\label{sec:conclusion}

%We introduce A-LIX.
% In future, we wish to explore the ability of A-LIX to mitigate warm-start issues in CNNs; as shown in \cite{mobahiselfdistillation}, the incorporation of data augmentation can substantially addresses this issue. We believe similar benefits can be drawn from applying A-LIX, and potentially to a greater degree due to the more explicit regularization. This presents exciting opportunities in continual learning and life-long learning \cite{continuallearning}, where generalization in the face of non-stationary data is a necessary property.
In this work, we provide a novel analysis demonstrating that instabilities in pixel-based off-policy RL come specifically from performing TD-learning with a convolutional encoder in the presence of a sparse reward signal. We show this \emph{visual deadly triad} affects the encoder's gradients, causing the critic to \textit{catastrophically self-overfit} to its own noisy predictions. Therefore, we propose \textbf{A}daptive \textbf{L}ocal S\textbf{I}gnal Mi\textbf{X}ing (A-LIX), a powerful regularization layer to explicitly counteract this phenomenon. Applying A-LIX enables us to outperform prior state-of-the-art algorithms on popular benchmarks without relying on image augmentations, auxiliary losses, or other notable design choices.

% Going forward, we believe that A-LIX can be applied to supervised learning and continual learning \citep{continuallearning}, and specifically to mitigate warm-start issues that may affect these problem settings \citep{warmstart}.

%% file: appendix.tex
\appendix
\onecolumn

\input{sectionsApp/1full_results}
\input{sectionsApp/2experiments_description}
\input{sectionsApp/3addition_analysis}
\input{sectionsApp/4implementation}
\input{sectionsApp/5ablations}
\input{sectionsApp/6offline_analysis}

%% file: sectionsApp/1full_results.tex
\section{Detailed Results}
\label{app:full_res}

\subsection{DMC Medium and Hard Tasks}
\label{app:dmc_full_res}

In Table \ref{tab:dmc_perf}, we show the performance in each of the evaluated 15 DMC environments by reporting the mean and standard deviations over the cumulative returns obtained midway and at the end of training for the medium and hard benchmark tasks, respectively. A-LIX attains state-of-the-art performance in the majority of the tasks at \emph{both} reported checkpoints, while still closely matching DrQ-v2's performance on the remaining tasks. On the other hand, DrQ-v2 struggles to consistently solve some of the harder exploration tasks such as Cartpole Swingup Sparse and Humanoid Run, as shown by the high standard deviations. Interestingly, unlike in the simpler DMC benchmark from \citet{planet} with higher action repeat, CURL appears have a slight edge over DrQ. In particular, the self-supervised signal from CURL appears to aid precisely in the sparse reward environments where DrQ-v2 struggles. Hence, this appears to suggest that including an additional self-supervised signal to the TD-loss, lessens the hindering effects of a lower-magnitude reward signal. We interpret this result as additional evidence showing how addressing any individual component of the deadly triad helps counteracting the catastrophic self-overfitting phenomenon.

We also test the significance of our results by performing a Wilcoxon signed-rank test \citep{wilcoxon} between A-LIX and DrQ-v2. We perform a paired rank test across both seeds and tasks, allowing us to obtain an $p$-value that takes into account both population size and relative performance gains across all tasks. The choice of Wilcoxon signed-rank test also does not presume normality in the distributions of performance which we believe is a more appropriate assumption than for instance a paired $t$-test \citep{studentttest}, despite a potential loss of statistical power. To ensure correct population pairing, A-LIX and DrQ-v2 seeds were identical, resulting in the same initially collected data and network initialization. Performing this test over all 15 tasks and 5 seeds, we achieve a $p$-value of $\mathbf{0.0057}$ at 50\% total frames (1.5M and 15M for Medium and Hard respectively) and $\mathbf{0.0053}$ at 100\% total frames (3.0M and 30M for Medium and Hard Respectively), much lower than the typical rejection criteria of $p>0.05$. We therefore believe this shows clear evidence that our results in DMC are \emph{strongly statistically significant}.

\begin{table}[H]
\caption{Full results for the DeepMind Control Suite benchmark. Each displayed return is averaged over 10 random seeds and from 10 evaluation runs collected at each experience checkpoint.} \label{tab:dmc_perf}
\vspace{-3mm}
\vskip 0.15in
\tabcolsep=0.05cm
\begin{center}
%\tiny
\adjustbox{max width=0.98\linewidth}{
%\begin{center}
\begin{tabular}{@{}lcccccccccc@{}}
\toprule
\textbf{}                        & \multicolumn{5}{c}{1.5M frames}                                   & \multicolumn{5}{c}{3.0M frames}                                  \\ \cmidrule(lr){2-6} \cmidrule(lr){7-11}
\textit{Medium tasks}            & SAC     & CURL    & DrQ     & DrQv2            & 	\textbf{A-LIX (Ours)}             & SAC     & CURL    & DrQ     & DrQv2           & 	\textbf{A-LIX (Ours)}             \\ \midrule
\textit{Acrobot Swingup}         & 8±9     & 6±5     & 24±27   & 256±47           & \textbf{270±99}  & 12±11   & 6±5     & 28±25   & \textbf{442±64}          & 402±100  \\
\textit{Cartpole Swingup Sparse} & 118±233 & 479±329 & 318±389 & 485±396          & \textbf{718±250}  & 185±295 & 499±349 & 316±389 & 505±412         & \textbf{742±250} \\
\textit{Cheetah Run}             & 9±8     & 507±114 & 788±59  & 792±29           & \textbf{806±78}  & 7±8     & 590±95  & 835±45  & \textbf{873±60}          & {864±78}   \\
\textit{Finger Turn Easy}        & 190±137 & 297±150 & 199±132 & \textbf{854±73}  & 546±101          & 200±155 & 309±176 & 216±158 & \textbf{934±54}          & {901±109}  \\
\textit{Finger Turn Hard}        & 79±73   & 174±106 & 100±63  & 491±182          & \textbf{587±109} & 100±78  & 146±95  & 86±70   & 902±77 & \textbf{906±101}          \\
\textit{Hopper Hop}              & 0±0     & 184±127 & 268±91  & 198±102          & \textbf{287±48}  & 0±0     & 224±135 & 285±96  & 240±123         & \textbf{372±48}  \\
\textit{Quadruped Run}           & 68±72   & 164±91  & 129±97  & 419±204          & \textbf{528±107}  & 63±45   & 175±104 & 130±59  & 523±271         & \textbf{759±107}  \\
\textit{Quadruped Walk}          & 75±65   & 134±53  & 144±149 & 591±256          & \textbf{776±37}  & 48±32   & 168±49  & 142±67  & 920±36          & \textbf{900±37}  \\
\textit{Reach Duplo}             & 1±1     & 8±10    & 8±12    & \textbf{220±7}   & 212±3           & 2±3     & 7±10    & 9±9     & \textbf{228±2}  & 221±3            \\
\textit{Reacher Easy}            & 52±64   & 707±142 & 600±201 & \textbf{971±4}   & 887±19           & 115±98  & 667±182 & 612±181 & 940±50          & \textbf{966±19}  \\
\textit{Reacher Hard}            & 3±2     & 463±196 & 320±233 & \textbf{727±172} & 720±83          & 10±23   & 678±350 & 397±273 & \textbf{935±49} & 855±83           \\
\textit{Walker Run}              & 26±4    & 379±234 & 474±148 & 571±276          & \textbf{691±10}  & 25±3    & 447±224 & 547±143 & 616±297         & \textbf{756±10}  \\ \midrule
\textit{\textbf{Average score}}  & 52.28   & 291.73  & 281.03  & 547.96           & \textbf{585.67}  & 63.80   & 326.45  & 300.27  & 671.40          & \textbf{720.30}  \\ \midrule
                                 & \multicolumn{5}{c}{15.0M frames}                                  & \multicolumn{5}{c}{30.0M frames}                                 \\ \cmidrule(lr){2-6} \cmidrule(lr){7-11}
\textit{Hard tasks}              & SAC     & CURL    & DrQ     & DrQv2            & 	\textbf{A-LIX (Ours)}             & SAC     & CURL    & DrQ     & DrQv2           & 	\textbf{A-LIX (Ours)}             \\ \midrule
\textit{Humanoid Walk}           & 7±3     & 5±3     & 3±2     & 243±162          & \textbf{476±79} & 4±3     & 4±3     & 5±3     & 675±86          & \textbf{754±79} \\
\textit{Humanoid Stand}          & 5±3     & 6±3     & 4±3     & 167±159          & \textbf{519±94} & 6±3     & 6±2     & 6±2     & 588±63          & \textbf{781±94}  \\
\textit{Humanoid Run}            & 5±3     & 6±2     & 5±3     & 22±30            & \textbf{122±59}  & 3±3     & 4±3     & 4±2     & 170±122         & \textbf{242±59}  \\ \midrule
\textit{\textbf{Average score}}  & 5.64    & 5.74    & 4.02    & 144.16           & \textbf{372.78}  & 4.30    & 4.89    & 4.90    & 477.74          & \textbf{592.48}  \\ \bottomrule
\end{tabular}
%\end{center}
}
\end{center}
%\vspace{-10pt}
\end{table}

We now compare our results using the \textit{Rliable} framework introduced in \citet{agarwal2021deep} (see App.~\ref{app:rliableinfo} for a detailed explanation about the metrics introduced).

\begin{figure}[H]
    % \vspace{-3mm}
    \centering
    \begin{overpic}[width=0.6\linewidth]{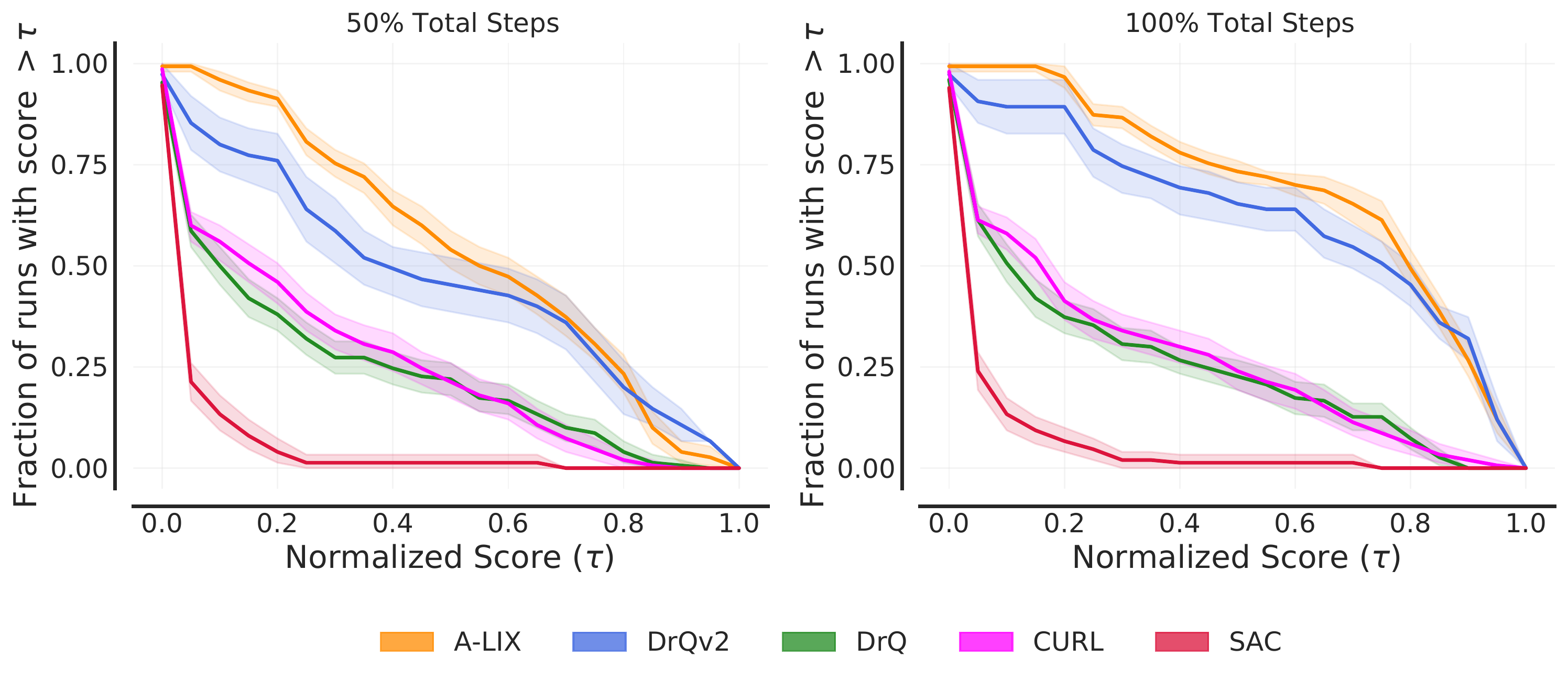} 
        \put (32,20) {\textcolor{red}{\textit{}}}
    \end{overpic}
    \vspace{-4mm}
    \caption{\small{Performance profiles at 50\% (\textbf{left}) and 100\% (\textbf{right}) of the total steps in Medium and Hard DMC Tasks.}}
    \vspace{-5mm}
    \label{fig:dmcperformanceprofiles}
\end{figure}%

We plot performance profiles in Fig.~\ref{fig:dmcperformanceprofiles} at both 50\% and 100\% the total training steps in DMC, which aim to represent sample efficiency and asymptotic performance respectively. We see that in almost all cases, A-LIX improves upon DrQ-v2.

\begin{figure}[H]
    \centering
    \vspace{-2mm}
    \begin{subfigure}[t]{0.27\linewidth}
        \begin{overpic}[width=0.99\linewidth,trim={0 0 10mm 0}]{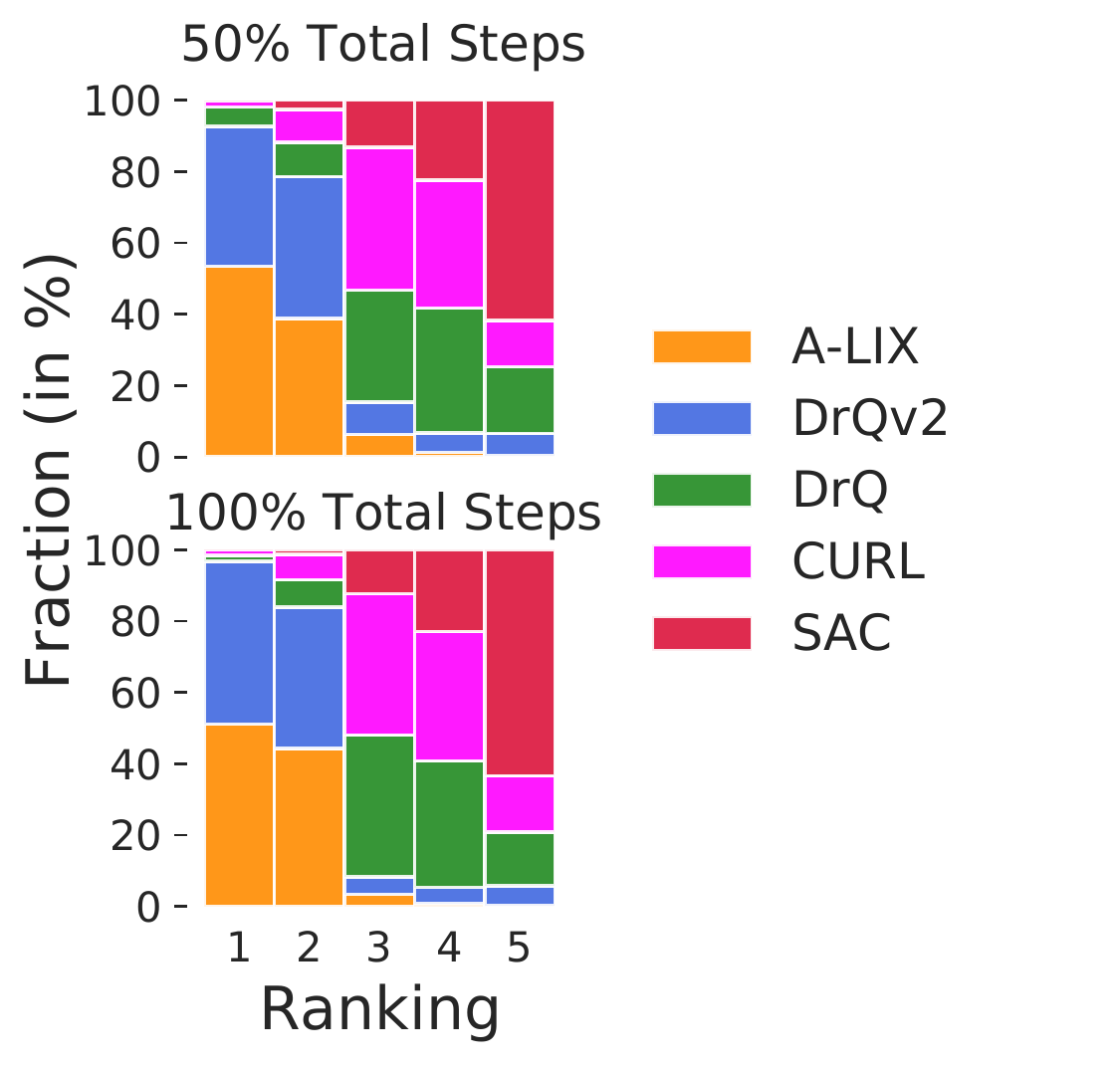} 
            \put (32,55) {\textcolor{red}{\textit{}}}
        \end{overpic}
        \vspace{-2mm}
        \caption{\small{Overall ranking statistics at 50\% (\textbf{top}) and 100\% (\textbf{bottom}) of the total steps in Medium and Hard DMC Tasks.}}
        \label{fig:dmcranking}
    \end{subfigure}
    \hspace{10mm}
    \begin{subfigure}[t]{0.5\linewidth}
        \begin{overpic}[width=0.99\linewidth]{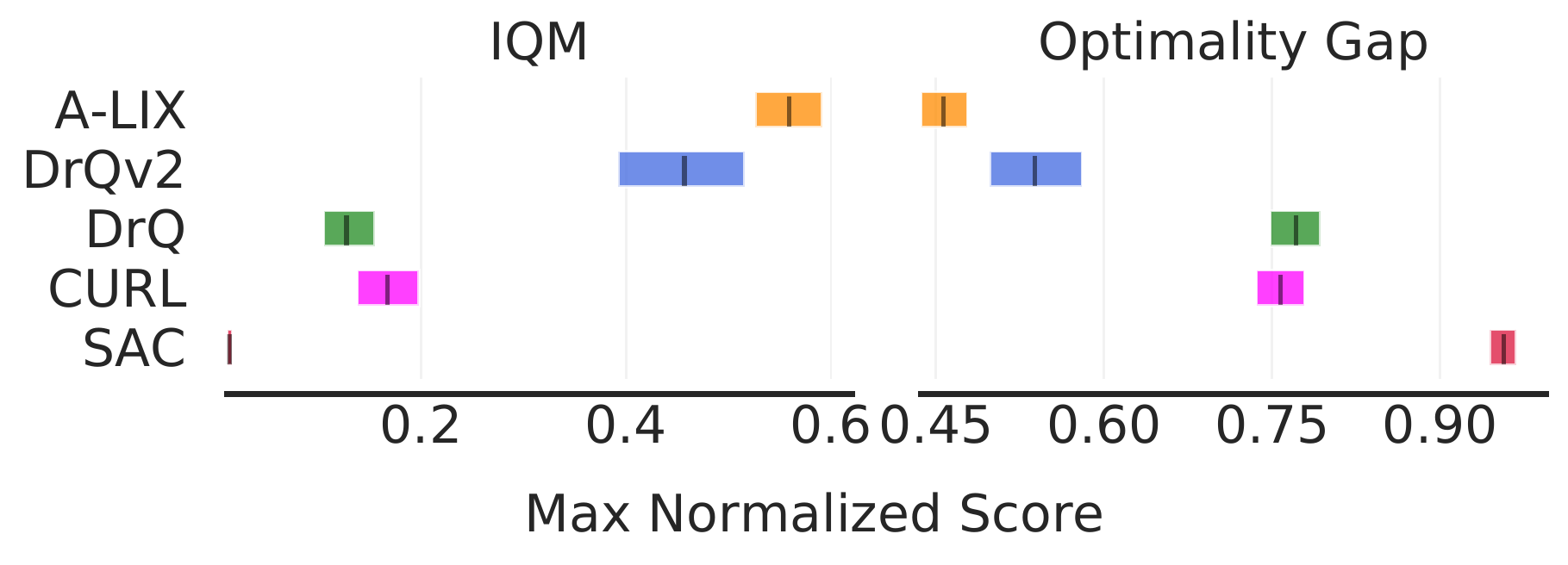} 
            \put (32,25) {\textcolor{red}{\textit{}}}
        \end{overpic}
        \vspace{-2mm}
        \caption{\small{Aggregate IQM (\textbf{left}) and Optimality Gap (\textbf{right}) metrics at 50\% of the total steps in Medium and Hard DMC Tasks.}}
        \label{fig:dmc50p_aggregate}
    \end{subfigure}
    \vspace{-5mm}
\end{figure}%

We plot ranking statistics in Fig.~\ref{fig:dmcperformanceprofiles} at both 50\% and 100\% the total training steps in DMC. We see that A-LIX clearly appears most in the 1st ranked column, and rarely appears in lower ranked (i.e., $>3$), suggesting strong performance across all environments in DMC Medium and Hard. We also provide a further aggregated statistics plot in Fig.~\ref{fig:dmc50p_aggregate} (this time at 50\% the total steps), which shows A-LIX is particularly sample-efficient and consistent (i.e., low error bars) across all environments.

\begin{figure}[H]
    %\vspace{-3mm}
    \centering
    \begin{subfigure}[t]{0.3\linewidth}
        \begin{overpic}[width=0.99\linewidth]{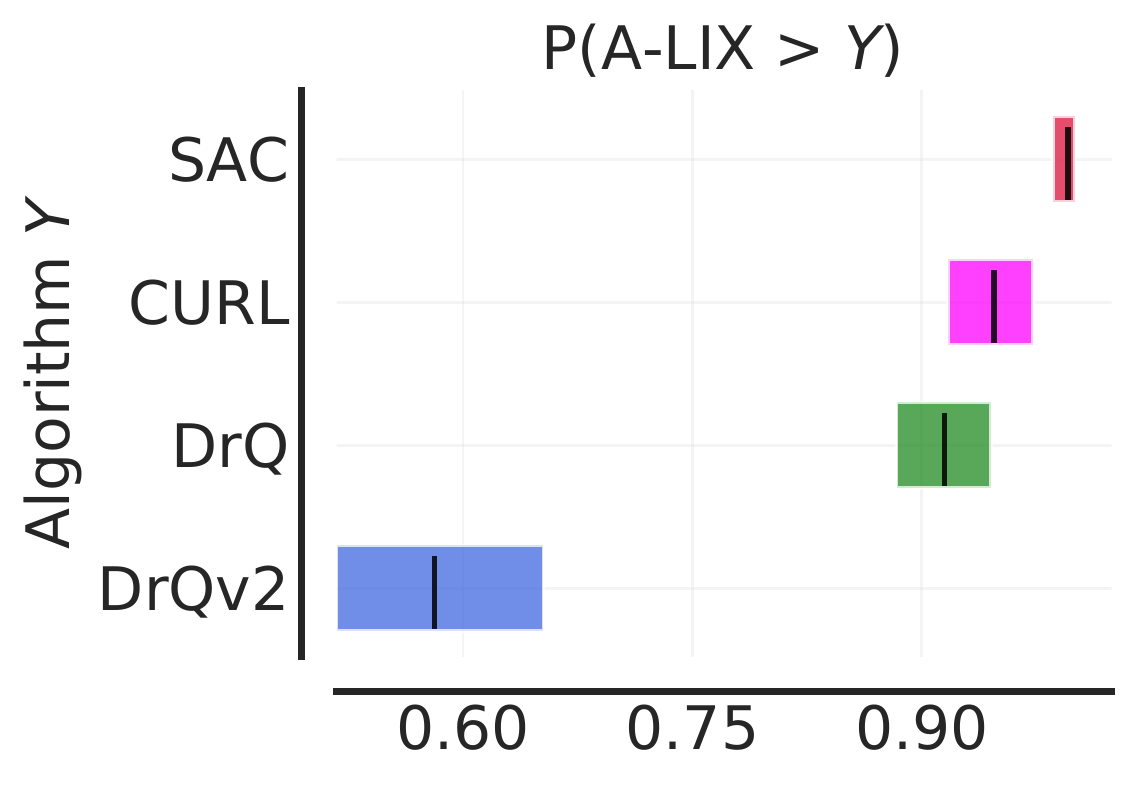} 
            \put (32,25) {\textcolor{red}{\textit{}}}
        \end{overpic}
        \vspace{-6mm}
        \caption{\small{50\% total steps.}}
    \end{subfigure}
    \hspace{15mm}
    \begin{subfigure}[t]{0.3\linewidth}
        \begin{overpic}[width=0.99\linewidth]{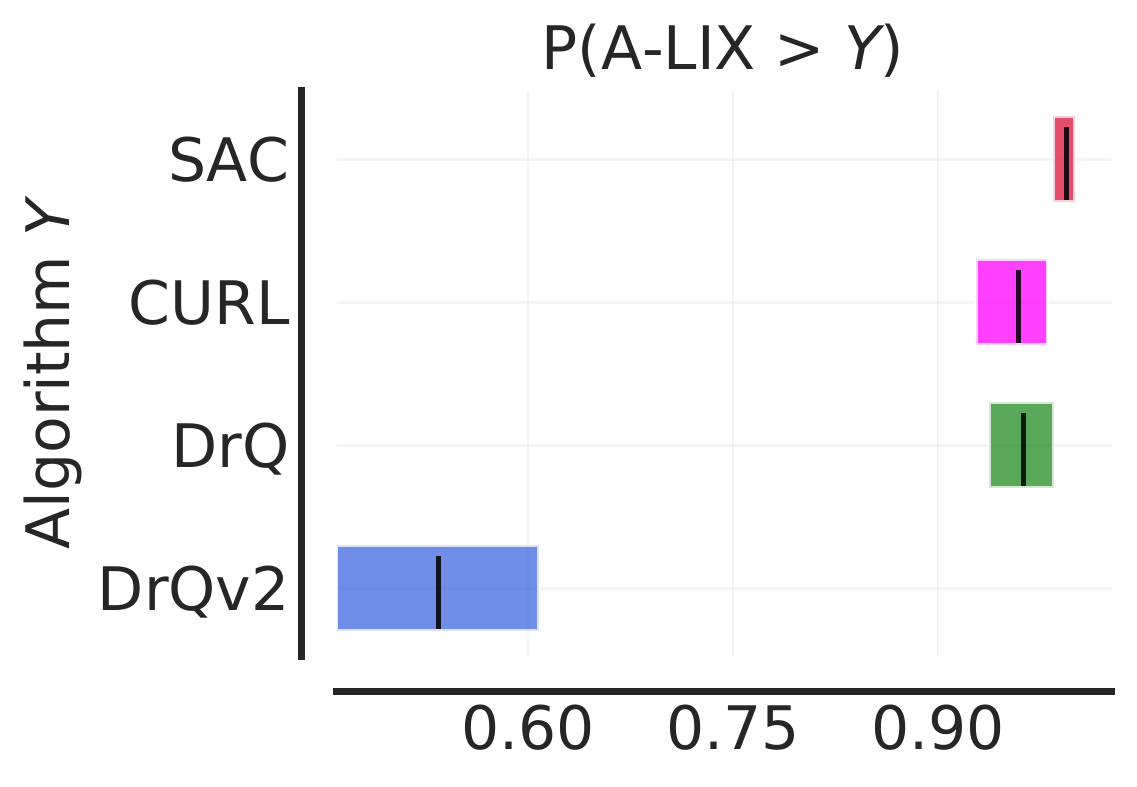} 
            \put (32,25) {\textcolor{red}{\textit{}}}
        \end{overpic}
        \vspace{-6mm}
        \caption{\small{100\% total steps.}}
    \end{subfigure}
    \vspace{-2mm}
    \caption{\small{Probability of Improvement statitistics at both 50\% (\textbf{left}) and 100\% (\textbf{right}) of the total timesteps in Medium and Hard DMC Tasks.}}
    \vspace{-5mm}
    \label{fig:dmc_pimprove}
\end{figure}%

 In Fig.~\ref{fig:dmc_pimprove} we observe that A-LIX likely improves over prior work, and note that whilst the improvement probability over DrQ-v2 may seem slightly low at $\sim$60\%, we note that this value is in line with statistics in prior works that achieve significant gains (as seen in \citet{agarwal2021deep}), and furthermore it does not take into account absolute performance values, and instead only compares relative values, which explains why the gains of A-LIX appear larger when evaluated under IQM and OG. Furthermore, the lower CI for 50\% total steps does not fall below 0.5, which means improvements are indeed statistically significant.

%\newpage
\subsection{Atari 100k}

In Table \ref{tab:full_atari_res}, we show the final average performance for all the evaluated algorithms in each of the twenty-six tasks in the Atari 100k benchmark. A-LIX outperforms SPR, the previous state-of-the-art off-policy algorithm on this benchmark, on 16 out of 26 tasks. Moreover, it attains comparatively similar performance on most of the remaining tasks despite using no augmentation, auxiliary losses, or model-based elements.

\label{app:atari_full_res}
\begin{table}[H]
\caption{Full results for the Atari 100k benchmark, following the evaluation protocol from \citet{machadoALEprotocol}. We report the results collected from 10 random seeds.} \label{tab:figarparam}
\label{tab:full_atari_res}
%\vskip 0.15in
\begin{center}
%\tiny
\adjustbox{max width=0.98\linewidth}{
\begin{tabular}{@{}lccccccccc@{}}
\toprule
\textit{\textbf{Tasks}}              & Random   & Human    & SimPLe           & DER            & OTRainbow    & CURL           & DrQ             & SPR              & \textbf{A-LIX (Ours)} \\ \midrule
\textit{Alien}                       & 227.80   & 7127.70  & 616.9            & 739.9          & 824.7        & 558.2          & 771.2           & 801.5            & \textbf{902}          \\
\textit{Amidar}                      & 5.80     & 1719.50  & 88               & \textbf{188.6} & 82.8         & 142.1          & 102.8           & 176.3            & 174.27                \\
\textit{Assault}                     & 222.40   & 742.00   & 527.2            & 431.2          & 351.9        & 600.6          & 452.4           & 571              & \textbf{660.53}       \\
\textit{Asterix}                     & 210.00   & 8503.30  & \textbf{1128.3}  & 470.8          & 628.5        & 734.5          & 603.5           & 977.8            & 809.5                 \\
\textit{Bank Heist}                  & 14.20    & 753.10   & 34.2             & 51             & 182.1        & 131.6          & 168.9           & 380.9            & \textbf{639.4}        \\
\textit{Battle Zone}                 & 2360.00  & 37187.50 & 5184.4           & 10124.6        & 4060.6       & 14870          & 12954           & \textbf{16651}   & 14470                 \\
\textit{Boxing}                      & 0.10     & 12.10    & 9.1              & 0.2            & 2.5          & 1.2            & 6               & \textbf{35.8}    & 21.5                  \\
\textit{Breakout}                    & 1.70     & 30.50    & 16.4             & 1.9            & 9.8          & 4.9            & 16.1            & 17.1             & \textbf{23.52}        \\
\textit{Chopper Command}             & 811.00   & 7387.80  & \textbf{1246.9}  & 861.8          & 1033.3       & 1058.5         & 780.3           & 974.8            & 747                   \\
\textit{Crazy Climber}               & 10780.50 & 35829.40 & \textbf{62583.6} & 16185.3        & 21327.8      & 12146.5        & 20516.5         & 42923.6          & 53166                 \\
\textit{Demon Attack}                & 152.10   & 1971.00  & 208.1            & 508            & 711.8        & 817.6          & \textbf{1113.4} & 545.2            & 888.15                \\
\textit{Freeway}                     & 0.00     & 29.60    & 20.3             & 27.9           & 25           & 26.7           & 9.8             & 24.4             & \textbf{31.04}        \\
\textit{Frostbite}                   & 65.20    & 4334.70  & 254.7            & 866.8          & 231.6        & 1181.3         & 331.1           & 1821.5           & \textbf{1845.7}       \\
\textit{Gopher}                      & 257.60   & 2412.50  & 771              & 349.5          & \textbf{778} & 669.3          & 636.3           & 715.2            & 500.6                 \\
\textit{Hero}                        & 1027.00  & 30826.40 & 2656.6           & 6857           & 6458.8       & 6279.3         & 3736.3          & 7019.2           & \textbf{7185.85}      \\
\textit{Jamesbond}                   & 29.00    & 302.80   & 125.3            & 301.6          & 112.3        & 471            & 236             & \textbf{365.4}   & 341.5                 \\
\textit{Kangaroo}                    & 52.00    & 3035.00  & 323.1            & 779.3          & 605.4        & 872.5          & 940.6           & 3276.4           & \textbf{6507}         \\
\textit{Krull}                       & 1598.00  & 2665.50  & 4539.9           & 2851.5         & 3277.9       & 4229.6         & 4018.1          & 3688.9           & \textbf{4884.04}      \\
\textit{Kung Fu Master}              & 258.50   & 22736.30 & \textbf{17257.2} & 14346.1        & 5722.2       & 14307.8        & 9111            & 13192.7          & 16316                 \\
\textit{Ms Pacman}                   & 307.30   & 6951.60  & \textbf{1480}    & 1204.1         & 941.9        & 1465.5         & 960.5           & 1313.2           & 1258.4                \\
\textit{Pong}                        & -20.70   & 14.60    & \textbf{12.8}    & -19.3          & 1.3          & -16.5          & -8.5            & -5.9             & 6.03                  \\
\textit{Private Eye}                 & 24.90    & 69571.30 & 58.3             & 97.8           & 100          & \textbf{218.4} & -13.6           & 124              & 100                   \\
\textit{Qbert}                       & 163.90   & 13455.00 & 1288.8           & 1152.9         & 509.3        & 1042.4         & 854.4           & 669.1            & \textbf{2974}         \\
\textit{Road Runner}                 & 11.50    & 7845.00  & 5640.6           & 9600           & 2696.7       & 5661           & 8895.1          & 14220.5          & \textbf{17471}        \\
\textit{Seaquest}                    & 68.40    & 42054.70 & \textbf{683.3}   & 354.1          & 286.9        & 384.5          & 301.2           & 583.1            & 654.6                 \\
\textit{Up N Down}                   & 533.40   & 11693.20 & 3350.3           & 2877.4         & 2847.6       & 2955.2         & 3180.8          & \textbf{28138.5} & 5011.7                \\ \midrule
\textit{\textbf{Human Norm. Mean}}   & 0.000    & 1.000    & 0.443            & 0.285          & 0.264        & 0.381          & 0.357           & 0.704            & \textbf{0.753}        \\
\textit{\textbf{Human Norm. Median}} & 0.000    & 1.000    & 0.144            & 0.161          & 0.204        & 0.175          & 0.268           & \textbf{0.415}   & 0.411                 \\ \midrule
\textit{\textbf{\# SOTA}}                       & N/A      & N/A      & 7                & 1              & 1            & 1              & 1               & 4                & \textbf{11}           \\
\textit{\textbf{\# Super}}           & N/A      & N/A      & 2                & 2              & 1            & 2              & 2               & \textbf{7}       & \textbf{7}            \\
\textit{\textbf{Average Rank}} & N/A      & N/A      & 3.92                & 5.00              & 5.21            & 3.92              & 4.85               & 2.88      & \textbf{2.21}            \\ \bottomrule
\end{tabular}
%\end{center}
}
\end{center}
\vspace{-3mm}
\end{table}
% \vfill
% \newpage
We now present additional evaluations under the \textit{Rliable} framework, continuing on from the analysis in Fig.~\ref{fig:atarimainrliable}.

\begin{figure}[H]
    \vspace{-1mm}
    \centering
    \begin{overpic}[width=0.6\linewidth]{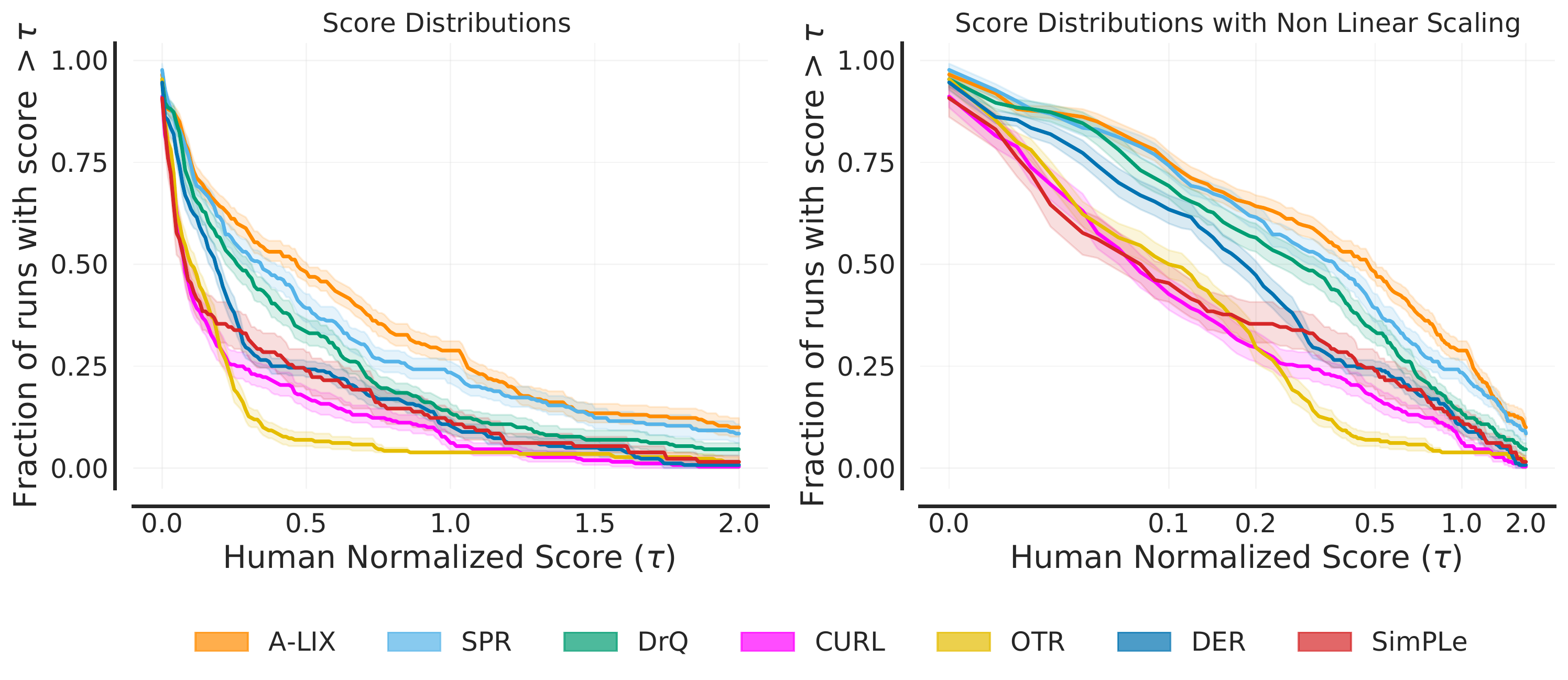} 
    \end{overpic}
    \vspace{-4mm}
    \caption{\small{Performance profiles with linear (\textbf{left}) and logarithmic (\textbf{right}) scaling in Atari 100k.}}
    \vspace{-5mm}
    \label{fig:atari100kperformanceprofile}
\end{figure}%

In Fig.~\ref{fig:atari100kperformanceprofile} A-LIX performs noticeably better than previous work, and performs at least as well as SPR over all settings of normalized scores.

\begin{figure}[H]
    \centering
    \begin{subfigure}[t]{0.33\linewidth}
        \begin{overpic}[width=0.99\linewidth]{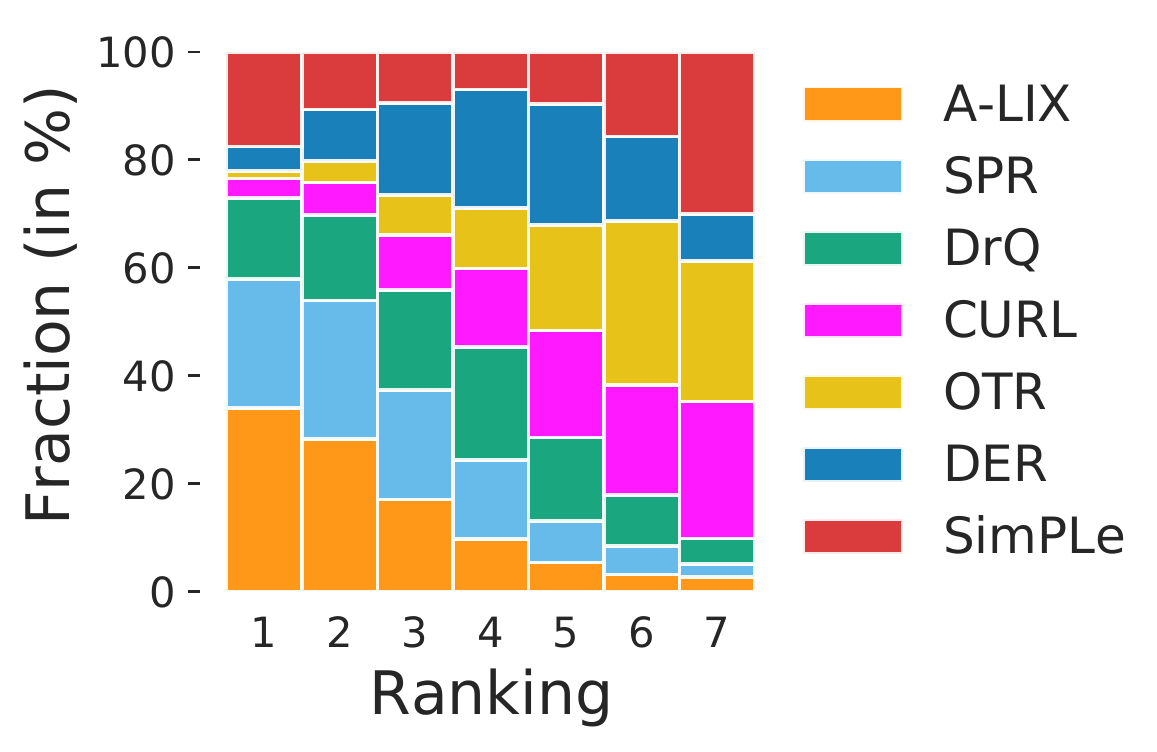}
        \end{overpic}
        \vspace{-6mm}
        \caption{\small{Ranking statistics.}}
        \label{fig:atari100k_ranking}
    \end{subfigure}
    \hspace{15mm}
    \begin{subfigure}[t]{0.33\linewidth}
        \begin{overpic}[width=0.99\linewidth]{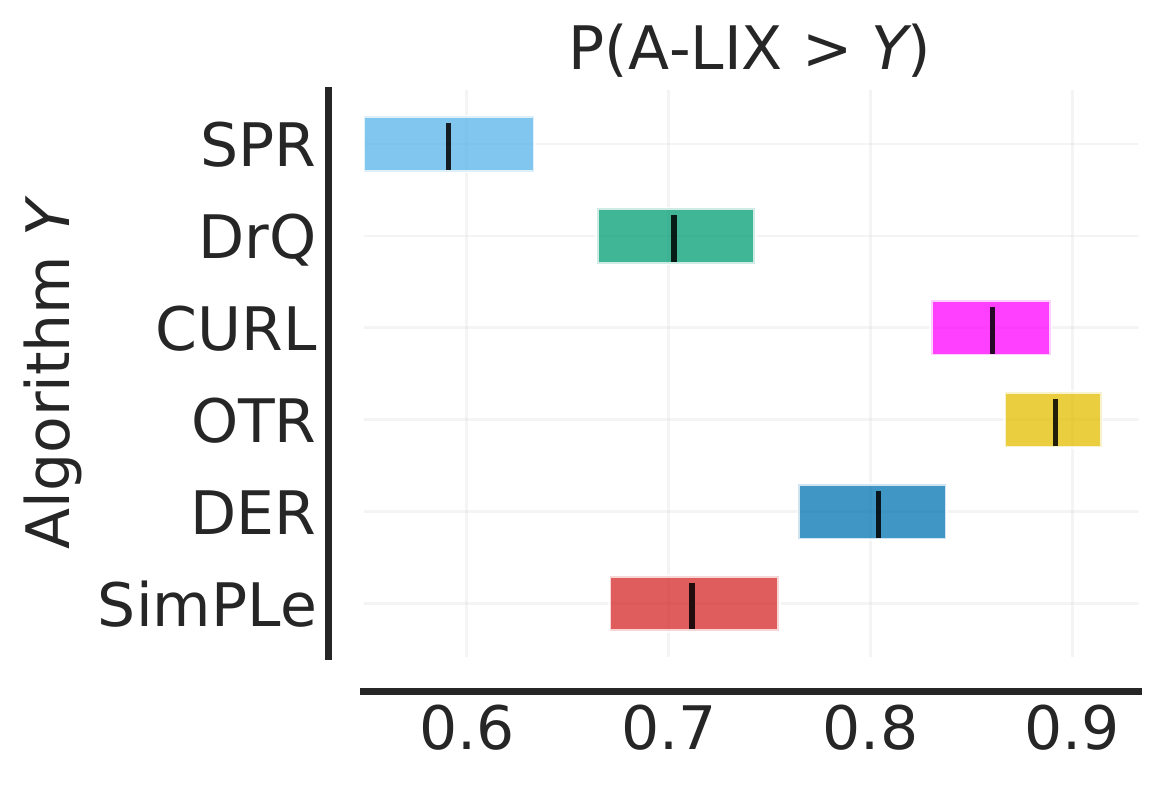}
        \end{overpic}
        \vspace{-6mm}
        \caption{\small{Probability of improvement statistics.}}
        \label{fig:atari100k_pimprove}
    \end{subfigure}
    \vspace{-2mm}
    \caption{\small{Bootstrapped ranking statistics (\textbf{left}) and probability of improvement plots (\textbf{right}) on Atari 100k.}}
    \vspace{-5mm}
\end{figure}%

In Fig.~\ref{fig:atari100k_ranking} A-LIX constitutes the majority of the algorithms ranked in 1st, and shows far fewer instances of being ranked in lower positions (i.e., $>4$). In Fig.~\ref{fig:atari100k_pimprove} we observe A-LIX likely improves upon prior work. Similar to Fig.~\ref{fig:dmc_pimprove}, while the $\sim$60\% improvement value over SPR may seem low, this is justified due to shortcomings in this metric, such as not taking into account actual performance values, and instead relative improvements. Furthermore, the lower CI does not fall below 0.5, which means improvements due to A-LIX are statistically significant.

\subsection{Rliable: A Primer}
\label{app:rliableinfo}

In addition to providing traditional methods of evaluation (e.g., performance tables, significance testing), we use robust metrics and evaluation strategies introduced in \textit{Rliable} \cite{agarwal2021deep}. Rliable advocates for computing aggregate performance statistics not just across many seeds, but also across the many tasks within a benchmark suite.

We give details on how these metrics achieve reliable performance evaluation in RL, denoting number of seeds as $N$ and number of tasks as $M$. We follow \citet{agarwal2021deep} as closely as possible; please refer to their paper for further details.

\subsubsection{Seed and Task aggregation}
In order to aggregate performances across different tasks in the same benchmark suite, we must first normalize each benchmark to the same range. In Atari, this is usually done by normalizing scores with respect to those achieved by humans, and in DMC this is done with respect to the maximum achievable score (i.e., $1,000$). We refer to this normalized score as $\tau$.

\subsubsection{IQM and OG}
Interquartile Mean (IQM) takes the middle 50\% of the runs across seeds and benchmarks (i.e., $[NM/2]$) and then calculates its mean score, improving outlier robustness whilst maintaining statistical efficiency. Optimality Gap (OG) calculates the proportion of performances ($NM$) that fail to meet a minimum threshold $\gamma$, with the assumption that improvements beyond $\gamma$ are not important. In both cases, stratified bootstrap sampling is used to calculate confidence intervals (CIs).

\subsubsection{Performance Profiles}
Performance profiles are a form of empirical CDF, but with stratified bootstrap sampling to produce confidence bands that account for the underlying variability of the score. We can also establish `stochastic dominance' by observing whether one method's performance profile is consistently above another's for all normalized performance values $\tau$.

\subsubsection{Ranking}
Ranking shows the proportion of times a given algorithm ranks in a given position across \emph{all} tasks, with distributions produced using stratified bootstrap sampling having $200,000$ repetitions.

\subsubsection{Probability of Improvement}
Probability of improvement is calculated by calculating the Mann-Whitney U-statistic \cite{mannwhitneyUstat} across all $M$ tasks. The distribution is then plotted as a boxplot, and if the lower CI $>$ 0.5, the improvement is \emph{statistically significant}.

\newpage

%% file: sectionsApp/2experiments_description.tex
\section{Experiments Description}
\subsection{Offline Experiments}
\label{sec:offlinedescrip}

% In online RL algorithms, it is difficult to isolate the effect of a particular design choice due to confounding effects during learning. For instance, assume augmentations somehow improve how agents explore; in this case, acquiring more diverse data will help in learning better representations, which in turn helps learn value functions that extrapolate better. Alternatively, assume augmentations give rise to representations that are less aliased \cite{nachum2021provable}; this would imbue agents with the ability to take diverse actions across states, promoting exploration, and so on. In both cases, we will observe an improvement in final performance, but it is difficult to say where the actual benefit lies. This serves as the motivation behind our offline experiments, which allows us to effectively isolate the effect of augmentations as discussed in Section~\ref{sec:analysis}.

We follow the original training hyperparameters of DrQ-v2, and run policy evaluation and policy improvement until we saw convergence in the TD-loss, which would occur at similar points in all agents (i.e., between 10-20k and 5-10k steps of SGD in policy evaluation and policy iteration respectively). For the proprioceptive experiments, we keep everything consistent, except the input to the critic and actor MLP layers are now the proprioceptive states from the DMC simulator, not the latent representation $z$ from the encoder. That is to say we do not modify the MLP architectures nor their learning rates in the interests of a fair comparison. Furthermore, for any given seed of the offline experiment, we also instantiate all networks in the agents identically and train on the same random offline data, with minibatches presented in the same order.

We also note that a similar algorithm is described in \citet{brandfonbrener2021offline}, but in the context of minimizing extrapolation errors. 

Now we present some additional analysis to provide further context to our offline experiments. First, we see that the proprioceptive statistics mirror those of the augmented agent, further illustrating the crucial role of CNN regularization for successful TD-learning from pixels:
\begin{figure}[H]
    \vspace{-4mm}
    \centering
    \includegraphics[width=0.6\linewidth]{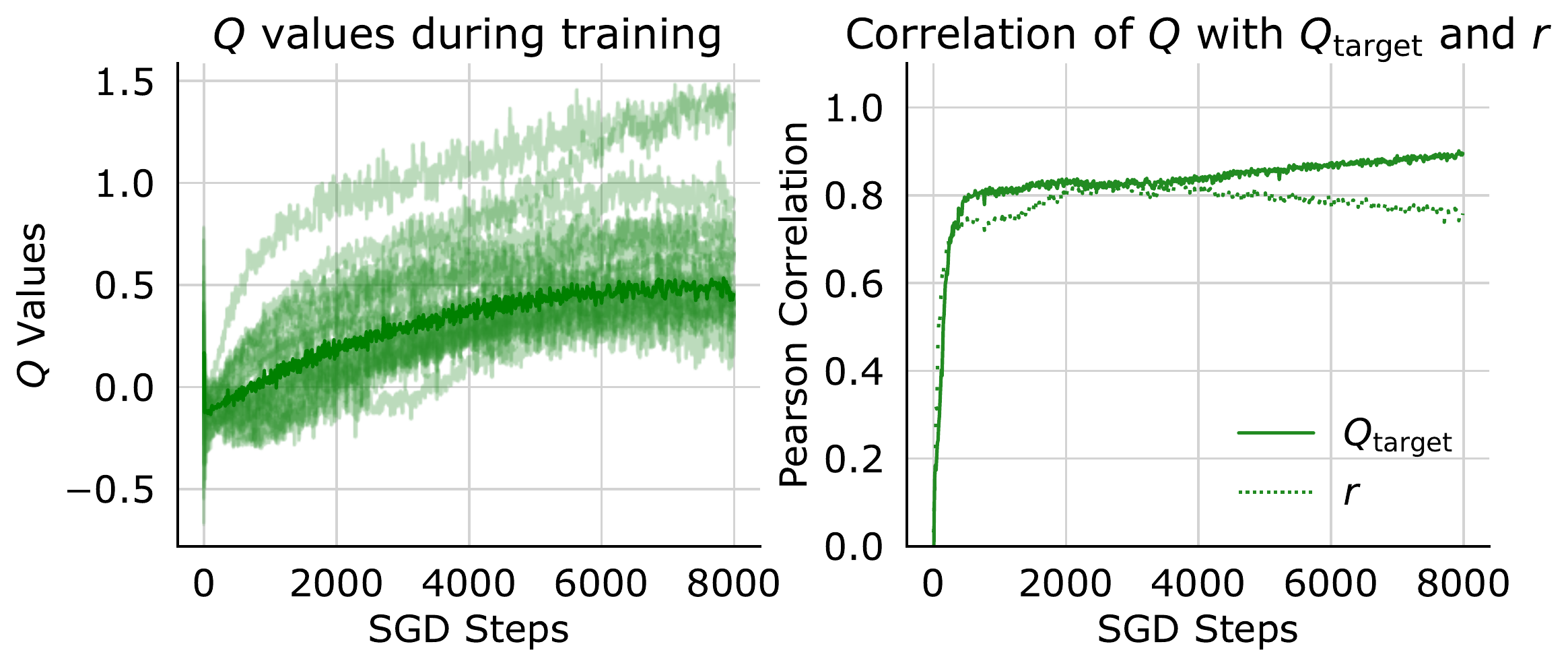} 
    \vspace{-3mm}
    \caption{\small{Q values and Pearson Correlation of the offline Proprioceptive agent on an offline fixed batch.}}
    \vspace{-5mm}
    %\label{fig:nstep}
\end{figure}%

Secondly, we observe that the exact same self-overfit also manifests in the online setting by plotting the Pearson correlation values over the initial stages of training in 5 seeds, confirming that phenomena of our offline analysis applies to the online RL problem:
\begin{figure}[H]
    \vspace{-4mm}
    \centering
    \includegraphics[width=0.6\linewidth]{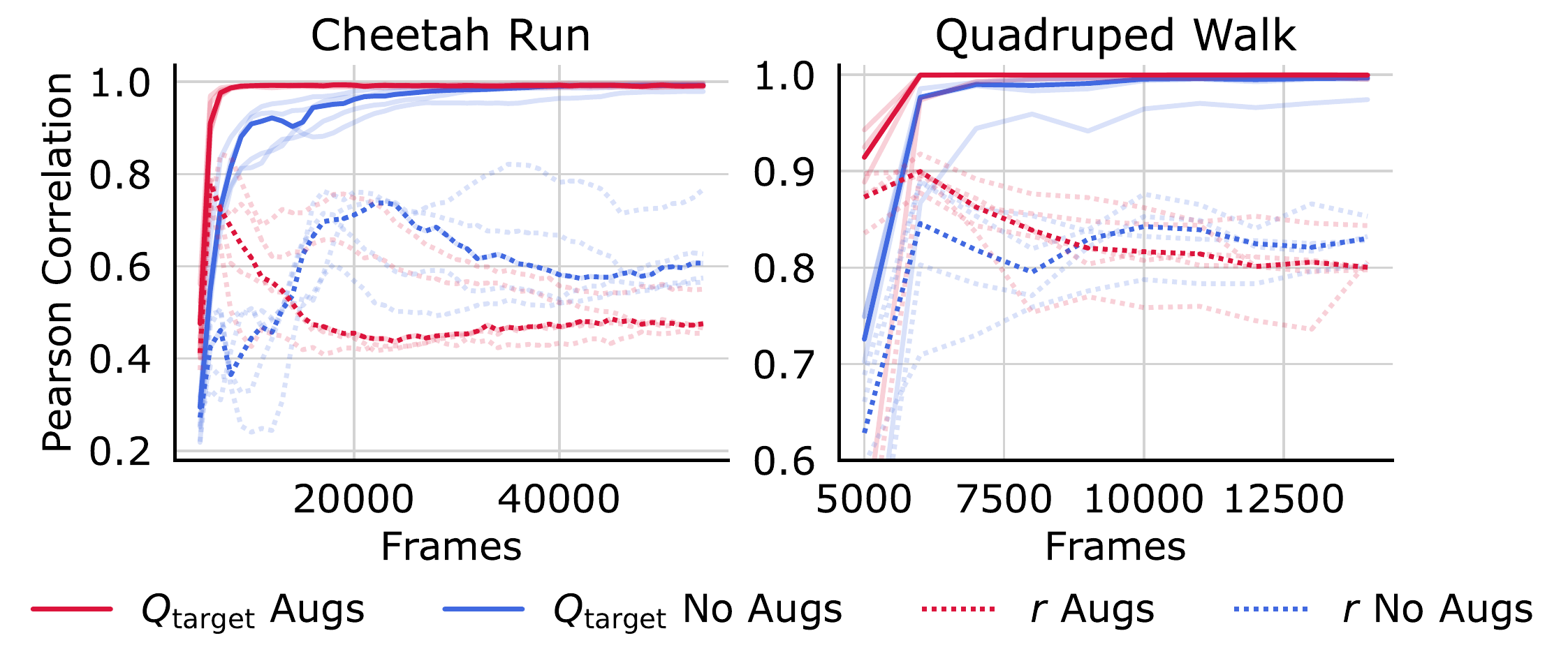} 
    \vspace{-6mm}
    \caption{\small{Pearson Correlation of augmented and non-augmented online agents in Cheetah Run and Quadruped Walk across 5 seeds. Shaded lines represent individual runs, and solid lines represent the median. We see that augmented agents do not immediately overfit to their target networks, and become correlated only after useful signal is learned.}}
    \vspace{-3mm}
    %\label{fig:nstep}
\end{figure}%

% Finally, as further evidence of the memorization effect, if we choose to regularize the outputs of the the target values by augmenting the next state images, but do not augment the current state images to the encoder (thus regularizing through the effect of output/label noise~\citep{labelnoise}). We find that the critic learns to overfit to the rewards, not its own target network:

\subsection{Jacobian Analysis}
\label{sec:jacobiananalysis}

In order to measure local sensitivity, we linearize the encoder around its input using a Taylor series expansion. Consider an $N$-dimensional input $\mathbf{x}\in\mathbb{R}^{N}$ and perturbation $\boldsymbol{\epsilon}\in\mathbb{R}^N$, an $M$-dimensional output $\mathbf{y}\in\mathbb{R}^M$, and a function $\mathbf{F}: \mathbb{R}^N \rightarrow \mathbb{R}^M$. Now, performing a Taylor series expansion around $\tilde{\mathbf{x}}$:
\begin{align}
    \mathbf{F}(\tilde{\mathbf{x}} + \boldsymbol{\epsilon}) &= \mathbf{F}(\tilde{\mathbf{x}}) + \boldsymbol{\epsilon} \mathbf{F}(\mathbf{x})\nabla^T|_{\mathbf{x}=\tilde{\mathbf{x}}} + \frac{\boldsymbol{\epsilon}^2}{2}\nabla\mathbf{F}(\mathbf{x})\nabla^T|_{\mathbf{x}=\tilde{\mathbf{x}}} + \dots \\
    &\approx \mathbf{F}(\mathbf{\tilde{\mathbf{x}}}) + \mathbf{J}(\tilde{\mathbf{x}})\boldsymbol{\epsilon}\\
    &= \tilde{\mathbf{y}}
\end{align}
where we make the approximation in the second line by dropping the second order/Hessian and higher terms under the assumption the perturbation vector $\boldsymbol{\epsilon}$ is small. This allows us to write $\mathbf{F}$ in the form of a local linear system: ${\mathbf{y}} = \mathbf{F}(\mathbf{\mathbf{x}}) + \mathbf{J}(\mathbf{x})\boldsymbol{\epsilon}$. It is straightforward to see that if the entries of the Jacobian matrix $\mathbf{J}$ are larger, then small perturbations $\boldsymbol{\epsilon}$ will cause larger changes in the output ${\mathbf{y}}$. To measure the magnitude of the Jacobian entries, we take the Frobenius norm:
\begin{align}
    ||\mathbf{J}(\mathbf{x})||_F = \sum_n \sum_m \left(\frac{\partial \mathbf{F}_m({\mathbf{x}})}{\partial \mathbf{x}_n}\right)^2
\end{align}
where $\mathbf{x}_n$ is the `$n$'th entry of $\mathbf{x}$ and $\mathbf{F}_m$ is the `$m$'th entry of the codomain of $\mathbf{F}$. The calculation of the Jacobian is trivial through the use of an automatic differentiation framework.

In our analysis we calculate the Jacobians of both agents on of a fixed batch of 128 frame stacked images taken from the offline training dataset, and compare the corresponding ratios of their Frobenius norms, and take this average ratio over the batch across 4 seeds.

%% file: sectionsApp/3addition_analysis.tex
\section{Additional Analysis}

\subsection{Adaptive ND Dual Objective Optimization}
\label{sec:adaptive_dual_objective}

The alternative $ND$ score with increased outlier robustness, $\widetilde{ND}$, proposed in Section~\ref{sec:alix} is inspired by recordings of signal-to-noise ratio measurements. In particular, by passing the individual normalized $D(z)$ terms through a $\log(1+x)$ smoothing function we downweight the effect that large individual outliers might have on this aggregated metric. We would like to remark that since we set up the optimization of $S$ with a dual objective, changes in the actual target value relating to some appropriate smoothness constraint are mostly irrelevant when considering the optimization's dynamics. Therefore, we argue that tuning $S$ with the actual $ND$ should not considerably diverge from tuning $S$ based on a re-scaled appropriate target for $\widetilde{ND}$. 

We provide further plots comparing agent performance and respective adaptive parameter $S$ during training:

\begin{figure}[H]
    \vspace{-3mm}
    \centering
    \includegraphics[width=0.9\linewidth]{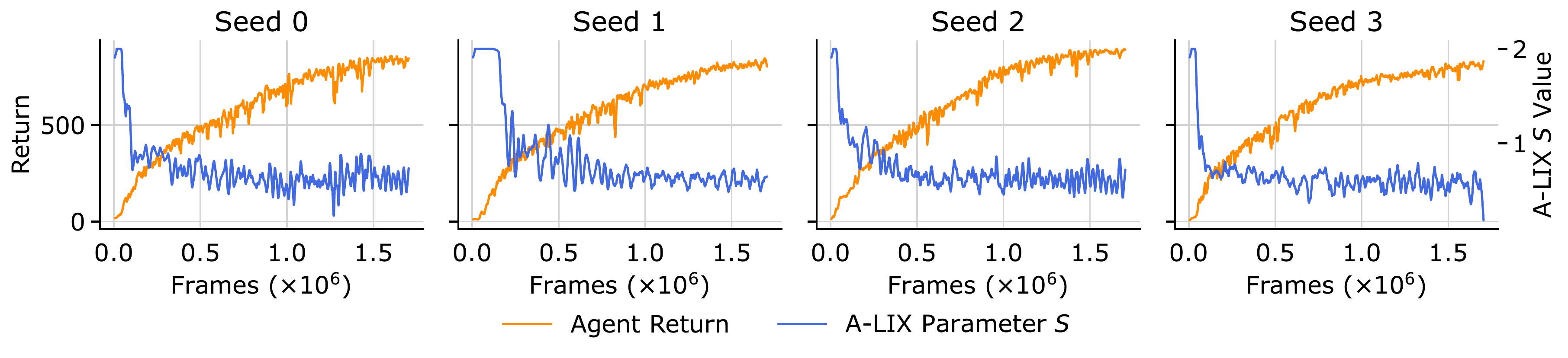}
    \vspace{-6mm}
    \caption{\small{{Performance of agents across 4 different seeds of the Cheetah Run environment and their adaptive scalar parameter $S$. We observe that initially, $S$ is high until agents learn useful behaviors, whereupon it drops to maintain ND due to presence of useful signal in the feature gradients.}}}
    \vspace{-3mm}
    \label{fig:adaptive_cheetah_run}
\end{figure}%

\begin{figure}[H]
    \vspace{-3mm}
    \centering
    \includegraphics[width=0.9\linewidth]{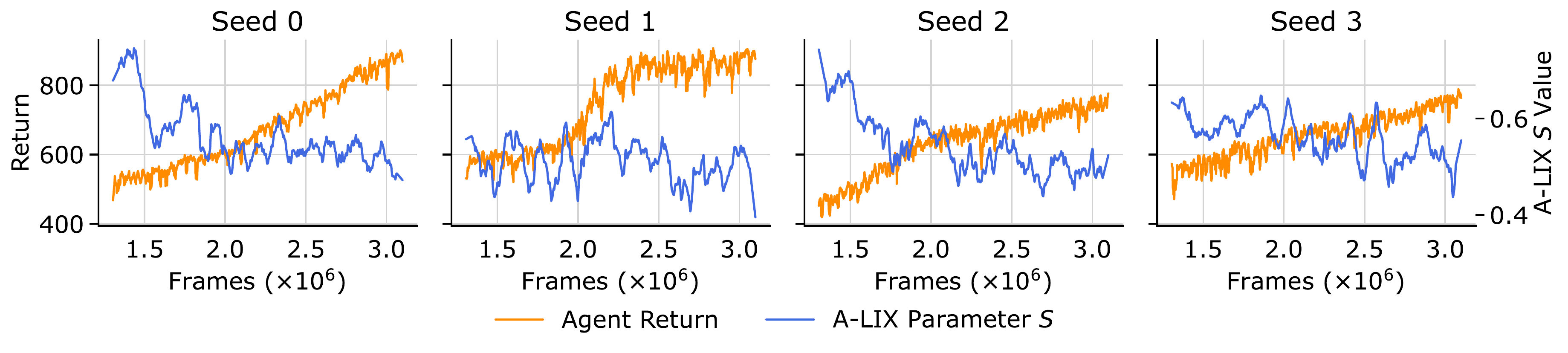}
    \vspace{-6mm}
    \caption{\small{{Performance of agents across 4 different seeds of the Quadruped Run environment and their adaptive scalar parameter $S$. We observe that as meaningful behaviors are learned in agents towards the end of training, $S$ falls accordingly, whereupon it drops to maintain ND due to presence of useful signal in the feature gradients.}}}
    \vspace{-3mm}
    \label{fig:adaptive_quadruped_run}
\end{figure}%

We see the same effect in these two contrasting environments; in Cheetah Run, where learning is more stable due to more predictable initializations and fewer degrees of freedom, we see the A-LIX parameter $S$ drop almost immediately as the TD-targets quickly become more accurate. In the less stable Quadruped Run, we also notice this annealing effect, however this occurs later on in training, when the agent can consistently recover from poor initializations.

\subsection{N-Step Returns}
\label{sec:nstep}
Large n-step rewards have become an important part of many algorithms that use TD-learning from visual observations.
As motivated in Section \ref{sec:analysis}, large n-step rewards can help towards mitigating self-overfitting by densifying the reward and downweighting the contribution of the inaccurate target critic, especially early in training; indeed as shown in \cite{drqv2}, using 1-step learning has a significant negative impact on performance. However, it is known that there is a bias-variance trade-off with multi-step approaches~\citep{nstepBias}, and furthermore, almost all approaches using this method do not apply off-policy bias correction when sampling from a replay buffer. While we motivate the use of n-step returns as a way to mitigate self-overfitting through incurring fewer 0 reward tuples (especially common in sparse reward environments early in training), we believe there is evidence to show that this introduces bias when n is sufficiently large, despite prior work suggesting this is not the case~\citep{nstepStudy}. 
\begin{figure}[H]
    \vspace{-2mm}
    \centering
    \includegraphics[width=0.6\linewidth]{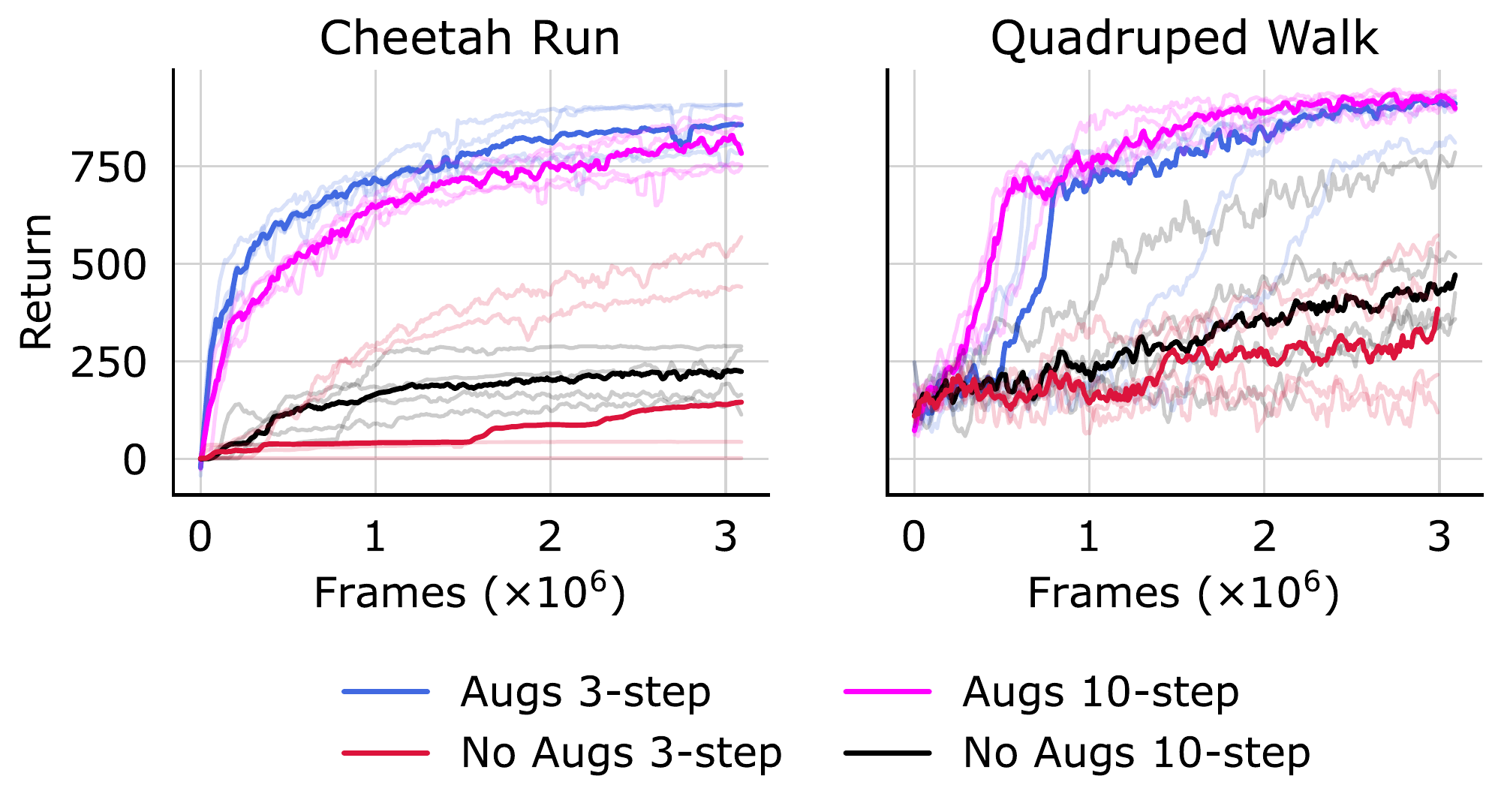} 
    \vspace{-3mm}
    \caption{\small{{Returns of agents over 5 seeds. Solid lines represent median performance, faded lines represent individual runs.}}}
    \vspace{-3mm}
    \label{fig:nstep}
\end{figure}%
We show in Fig.~\ref{fig:nstep} that 10-step (as is commonly done in algorithms used to solve Atari) returns can mitigate failure seeds as predicted under the visual deadly triad framework (indeed in Cheetah Run there are no seeds that completely flat-line when 10-step returns are used). However, we also see evidence that applying 10-step returns can have negative impacts on convergence and asymptotic performance in Cheetah Run when the deadly triad is sufficiently managed, such as using augmentations; in Quadruped Run we see moderate benefit initially, but note that asymptotically the 10-step and 3-step agents converge to the same performance. We also provide further evidence in App.~\ref{sec:alixablations}, where applying 10-step returns to an A-LIX agent generally has a laregely negative impact on performance. Finally, we note that trying 20-step returns, as is done in some algorithms that solve Atari~\cite{curl}, caused significant performance reductions in DMC. In conclusion, this provides evidence that we should consider using lower values of `n' in multi-step returns, and achieve this through addressing other elements of the deadly triad. 
\newpage

%% file: sectionsApp/4implementation.tex
\section{Implementation Details}

In Tables \ref{tab:alix_drq_hyper} and \ref{tab:alix_atari_hyper} we provide the full list of hyperparameters used in our implementations for DMC and Atari 100k, respectively. We show significant differences from standard practices in \textbf{bold}. In particular, A-LIX uses \textit{the same encoder architecture and n-step returns} for both benchmarks, highlighting its lower reliance to environment-specific heuristics. Moreover, unlike prior state-of-the-art algorithms it \textit{does not employ any data augmentation or auxiliary loss function}. These factors show the effectiveness of our adaptive method in counteracting instabilities from the visual deadly triad without any additional help, highlighting its applicability.

\label{sec:app_hyper}
\vfill
\begin{table}[H]
\caption{Full hyperparameters list used for the DeepMind Control A-LIX experiments. \textbf{Bolded} values represent significant differences from canonical implementations.} \label{tab:alix_drq_hyper}

\vskip 0.15in
\begin{center}
%\tiny
%\adjustbox{max width=0.98\linewidth}{
%\begin{center}
\begin{tabular}{@{}ll@{}}
\toprule
\multicolumn{2}{c}{DDPG-integration hyperparameters (following \citep{drqv2})}                                                                                                                                \\ \midrule
Replay data buffer size                                   & $1000000$ ($100000$ for \textit{Quadruped Run})                                                                                                      \\
Batch size                                                & $256$ ($512$ for \textit{Walker Run})                                                                                                                \\
Minimum data before training                              & $4000$                                                                                                                                               \\
Random exploration steps                                  & $2000$                                                                                                                                               \\
Optimizer                                                 & \textit{Adam} \citep{adam}                                                                                                                           \\
Policy/critic learning rate                               & \begin{tabular}[c]{@{}l@{}}medium: $0.0001$\\ hard: $0.00008$\end{tabular}                                                                           \\
Policy/critic $\beta_1$                                   & $0.9$                                                                                                                                                \\
Critic UTD ratio                                          & $0.5$                                                                                                                                                \\
Policy UTD ratio                                          & $0.5$                                                                                                                                                \\
Discount $\gamma$                                         & $0.99$                                                                                                                                               \\
Polyak coefficient $\rho$                                 & $0.99$                                                                                                                                               \\
\textit{N}-step returns                                   & $3$ ($1$ for \textit{Walker Run})                                                                                                                    \\
Hidden dimensionality                                     & $1024$                                                                                                                                               \\
Feature dimensionality                                    & \begin{tabular}[c]{@{}l@{}}medium: $50$ \\ hard: $100$\end{tabular}                                                                                  \\
Nonlinearity                                              & ReLU                                                                                                                                                 \\
Exploration stddev. clip                                  & $0.3$                                                                                                                                                \\
Exploration stddev. schedule                              & \begin{tabular}[c]{@{}l@{}}medium: linear: $1 \rightarrow 0.1$ in $500000$ steps\\ hard: linear: $1 \rightarrow 0.1$ in $2000000$ steps\end{tabular} \\
\textbf{Augmentations}                                    & \textbf{OFF}                                                                                                                                        \\ \midrule
\multicolumn{2}{c}{A-LIX-specific hyperparameters}                                                                                                                                                              \\ \midrule
\textbf{Initial maximum sampling shift $S$}               & $1.0$                                                                                                                                                \\
\textbf{Normalized discontinuity targets $\overline{ND}$} & $0.635$                                                                                                                                              \\
\textbf{Maximum sampling shift learning rate}             & \textbf{$0.003$}                                                                                                                                     \\
\textbf{Maximum sampling shift $\beta_1$}                 & $0.5$                                                                                                                                                \\ \bottomrule
\end{tabular}
%\end{center}
%}
\end{center}
%\vspace{-10pt}
\end{table}
\vfill
\newpage

% \vspace{30mm}
\begin{table}[H]
\vspace{30mm}
\caption{Full hyperparameters list used for the Atari 100k A-LIX experiments. \textbf{Bolded} values represent significant differences from canonical implementations.} \label{tab:alix_atari_hyper}

\vskip 0.15in
\begin{center}
%\tiny
%\adjustbox{max width=0.98\linewidth}{
%\begin{center}
\begin{tabular}{@{}ll@{}}
\toprule
\multicolumn{2}{c}{DER-integration hyperparameters}                                          \\ \midrule
Gray-scaling                                              & True                              \\
Down-sampling                                             & $84\times 84$                     \\
Frames stacked                                            & $4$                               \\
Action repetitions                                        & $4$                               \\
Reward clipping                                           & $[-1, 1]$                         \\
Max episode frames                                        & $108000$                          \\
Replay data buffer size                                   & $100000$                          \\
Replay period every                                       & $1$                               \\
Batch size                                                & $32$                              \\
Minimum data before training                              & $1600$                            \\
Random exploration steps                                  & $1600$                            \\
Optimizer                                                 & \textit{Adam} \citep{adam}        \\
Critic learning rate                                      & $0.0001$                          \\
Critic $\beta_1$                                          & $0.9$                             \\
Critic $\epsilon$                                             & $0.000015$                        \\
Max gradient norm                                         & $10$                              \\
Critic UTD ratio                                          & $2$                               \\
Discount $\gamma$                                         & $0.99$                            \\
Target update period                                      & $1$                               \\
\textbf{\textit{N}-step returns}                          & $\mathbf{3}$                               \\
\textbf{Feature maps}                                     & $\mathbf{32, 32, 32}$                      \\
\textbf{Filter sizes}                                     & $\mathbf{3\times 3, 3\times 3, 3\times 3}$ \\
\textbf{Strides}                                          & $\mathbf{2,1,1}$                           \\
Hidden dimensionality                                     & $256$                             \\
Feature dimensionality                                    & $50$                              \\
Nonlinearity                                              & ReLU                              \\
Exploration noisy nets parameter                          & $0.1$                             \\
\textbf{Augmentations}                                    & \textbf{OFF}                      \\ \midrule
\multicolumn{2}{c}{A-LIX-specific hyperparameters}                                           \\ \midrule
\textbf{Initial maximum sampling shift $S$}               & $1.0$                             \\
\textbf{Normalized discontinuity targets $\overline{ND}$} & $0.75$                            \\
\textbf{Maximum sampling shift learning rate}             & \textbf{$0.0001$}                 \\
\textbf{Maximum sampling shift $\beta_1$}                 & $0.5$                             \\ \bottomrule
\end{tabular}
%\end{center}
%}
\end{center}
%\vspace{-10pt}
\end{table}

\newpage

%% file: sectionsApp/5ablations.tex
\section{Additional Ablations}
\label{sec:app_abl}

\subsection{Smoothness Regularization through Spectral Normalization}
\label{sec:specnorm}
To distinguish between general smoothness contraints in convolutional features, and the smoothness that arises as a result spatial consistency, we apply spectral normalization~\citep{spectralNorm} to the final convolutional layer in the encoder to represent the former class of constraints. Spectral normalization operates on the parameters of a network and constrains its outputs to be 1-Lipschitz and has shown benefits in prior work~\citep{specnormDQN}, but does not explicitly enforce a \emph{spatial regularization} in the features. We train agents without augmentations using spectral normalization.
\begin{figure}[H]
    \vspace{-2mm}
    \centering
    \includegraphics[width=0.6\linewidth]{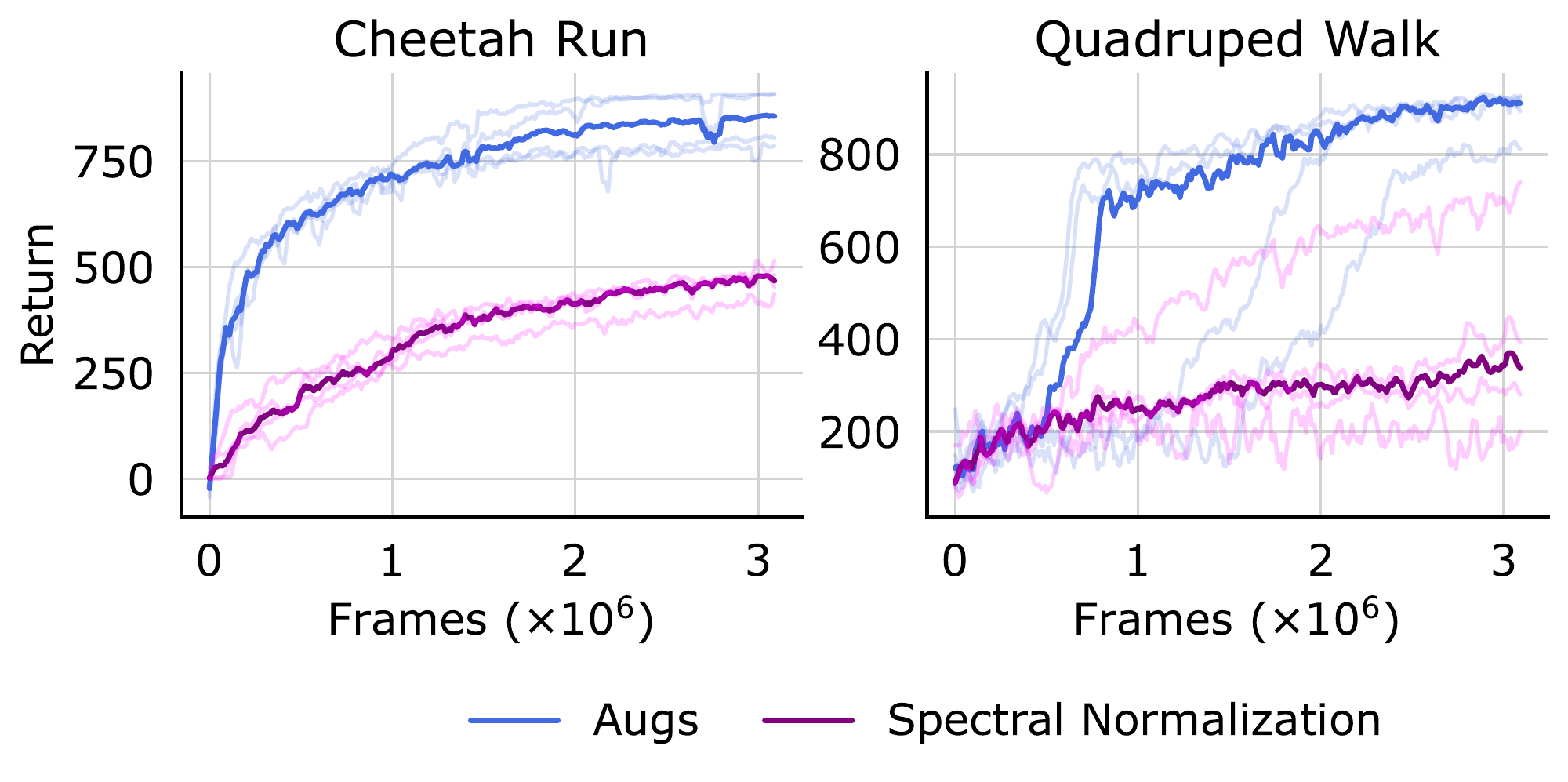} 
    \vspace{-6mm}
    \caption{\small{{Returns of agents over 5 seeds. Solid lines represent median performance, faded lines represent individual runs.}}}
    \vspace{-4mm}
    \label{fig:specnorm}
\end{figure}%

We see that whilst there is clear improvement above the original non-augmented agents in some cases, the performance is still lower than agents that use spatial consistency regularization, such as random shift augmentations.

\subsection{Is Gradient Smoothing All We Need?}
\label{sec:agauss}

Following the argument in Section \ref{sec:rand_shifts}, we can view augmentations as a gradient smoothing regularizer. This naturally leads us to ask the following: can we replace the stochastic shifting mechanism with a fixed smoothing mechanism? To test this, we instead apply a Gaussian smoothing kernel to the feature gradients in the CNN, and utilize our $ND$ score to vary the width of the kernel adaptively through training; we call this method A-Gauss (\textbf{A}daptive \textbf{Gauss}ian Feature Gradient Kernel).
\begin{figure}[H]
    \vspace{-2mm}
    \centering
    \includegraphics[width=0.6\linewidth]{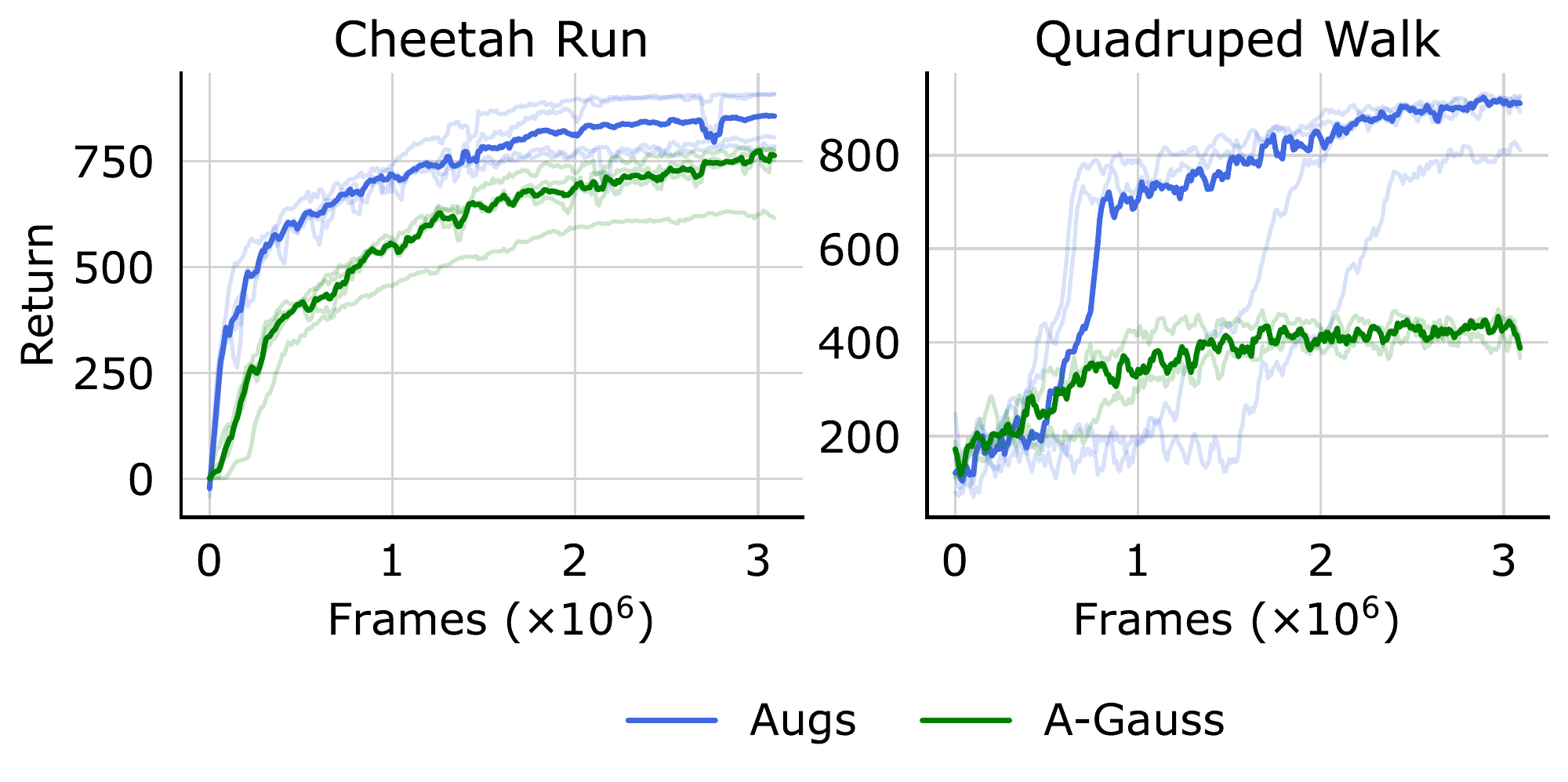} 
    \vspace{-6mm}
    \caption{\small{{Returns of agents over 5 seeds. Solid lines represent median performance, faded lines represent individual runs.}}}
    \vspace{-4mm}
    %\label{fig:specnorm}
\end{figure}%

We see that while there is improvement over non-augmented agents, overall performance is still lower than even simple non-adaptive augmentation. We believe this is due to the Gaussian kernel having too significant an effect on the information contained in the feature gradients during backpropagation, causing information to be lost. We believe this explains the effectiveness of shift-augmentations in reinforcement learning, which is that they effectively balance the information contained in the gradients, as well as ensuring their smoothness to reduce overfitting. 

\subsection{Ablations to A-LIX}
\label{sec:alixablations}

We now provide a set of ablations on both DMC and Atari, assessing the impact of individual components in A-LIX.

\begin{figure}[H]
    \vspace{-2mm}
    \centering
    \hfill
    \begin{subfigure}[t]{0.5\linewidth}
        \centering
        \includegraphics[width=0.999\linewidth]{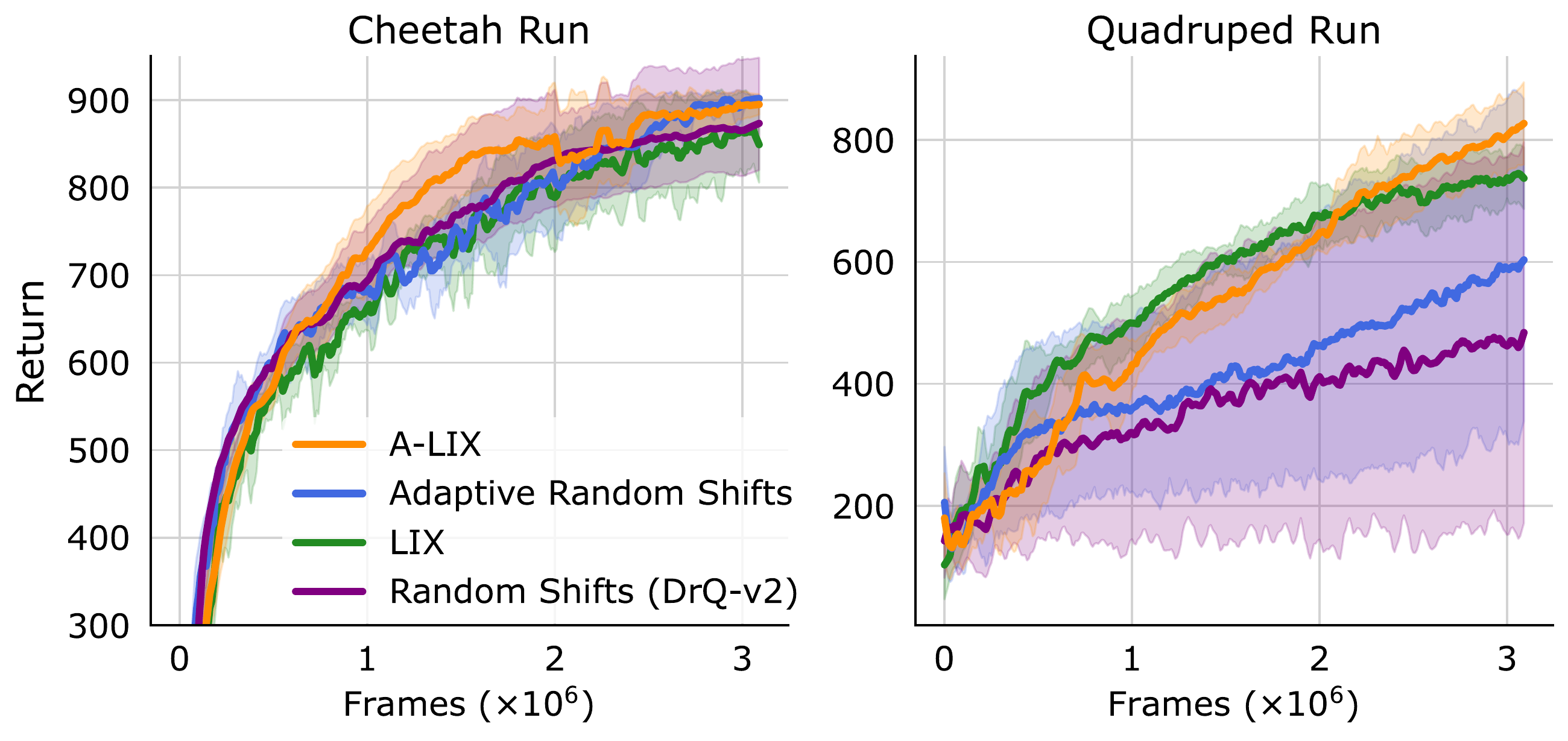}
        \vspace{-5mm}
        \caption{\small{{DMC Control ablations in Cheetah Run (\textbf{left}) and Quadruped Run (\textbf{right}) evaluated over 4 seeds.}}}
        \vspace{15pt}
        \label{fig:dmc_ablations}
    \end{subfigure}
    \hspace{10mm}
    \begin{subfigure}[t]{0.4\linewidth}
        \centering
        \includegraphics[width=0.999\linewidth]{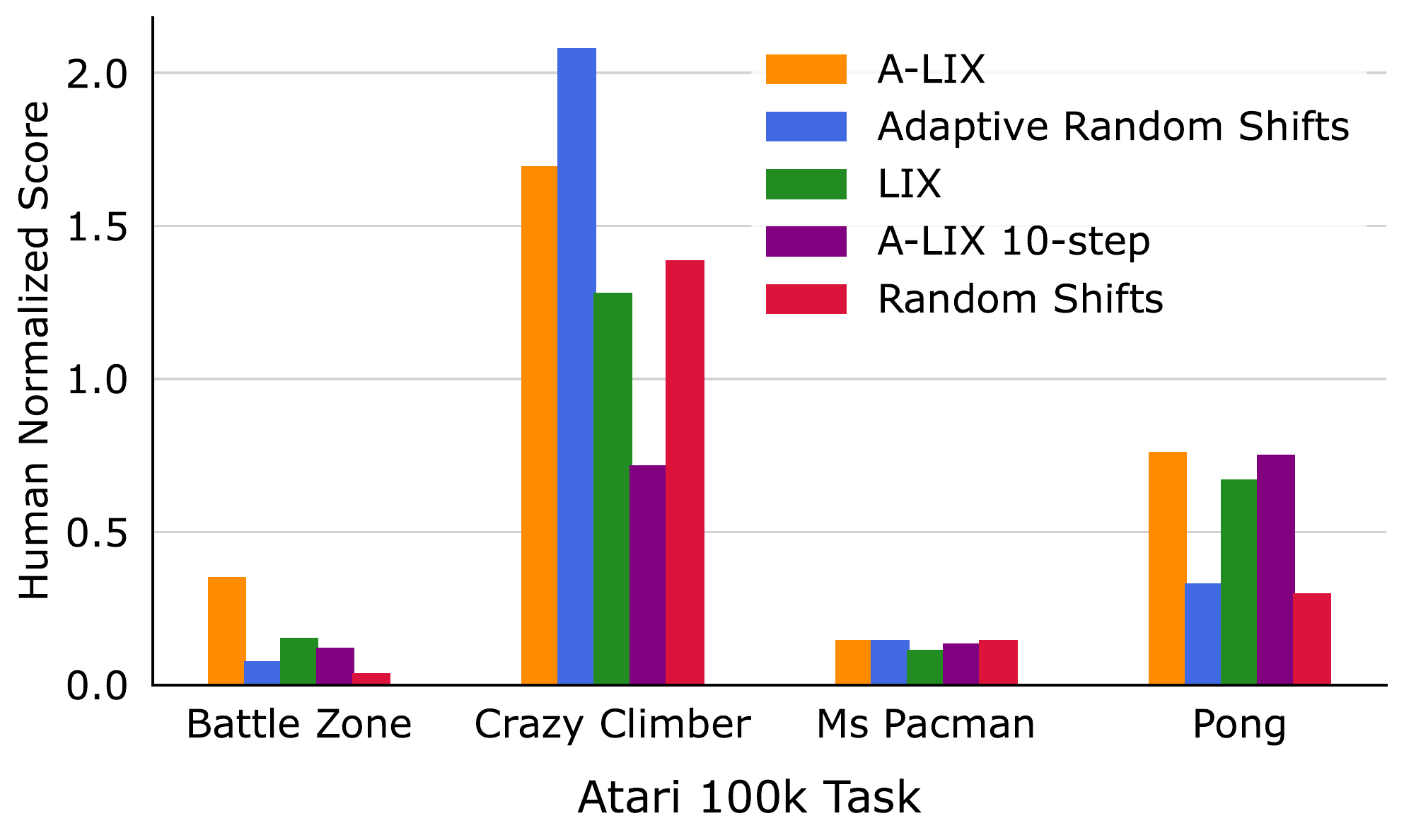}
        \vspace{-5mm}
        \caption{\small{{Atari 100k ablations evaluated over 4 seeds in 4 different Atari 100k tasks.}}}
        \vspace{-2mm}
        \label{fig:atari100k_ablations}
    \end{subfigure}
    \vspace{-6mm}
    \caption{\small{An ablation study of A-LIX, showing the contribution of its individual components to ultimate performance in DMC and Atari 100k.}}
    \vspace{-3mm}
\end{figure}%
In Fig.~\ref{fig:dmc_ablations} we choose the following ablations for DMC:
\vspace{-3mm}
\begin{itemize}
    \itemsep0em 
    \item A-LIX
    \item Adaptive Random Shifts (where the magnitude of the random shift image augmentation is adjusted using the dual ND objective)
    \item LIX
    \item Random Shifts (i.e., DrQ-v2)
\end{itemize}
\vspace{-3mm}
While we see a slight asymptotic performance improvement in Cheetah Run by using LIX layers instead of random shifts, we notice significant differences in the less stable Quadruped Run environment. Concretely, we see much greater stability in both LIX approaches compared with image augmentation approaches, with the former having no failure seeds. Furthermore, we observe stronger asymptotic performance with the inclusion of the adaptive dual objective for both approaches. As motivated in Fig.~\ref{fig:adaptive_quadruped_run}, this is likely a result of reducing the shift parameter as the signal in the target values increases.

In Fig.~\ref{fig:atari100k_ablations}, we choose the following ablations for Atari 100k on a subset of environments that represent a diverse set of tasks and performances with baseline algorithms:
\vspace{-3mm}
\begin{itemize}
    \itemsep0em 
    \item A-LIX
    \item Adaptive Random Shifts (as before)
    \item LIX
    \item A-LIX with 10-step returns
    \item Random Shifts
\end{itemize}
\vspace{-3mm}
We see that A-LIX performs consistently strongly across the environments tested, always placing in the top 2 with regards to Human Normalized Score. We also notice that generally, LIX layer methods outperform random shift methods apart from in Crazy Climber, where the opposite is true. We believe this may be due to random shift augmentations actually reflecting the inductive biases concerning generalization in this environment, and believe this merits further investigation. Finally, we observe that using 10-step returns instead of 3 generally harms performance with A-LIX, with justification given in App.~\ref{sec:nstep}.

%% file: sectionsApp/6offline_analysis.tex
\section{Additional Offline Experiment Analysis}
\subsection{Behavior Cloning without Augmentations}
\label{sec:behaviorcloning}

\begin{figure}[H]
    \vspace{-2mm}
    \centering
    \includegraphics[width=0.4\linewidth]{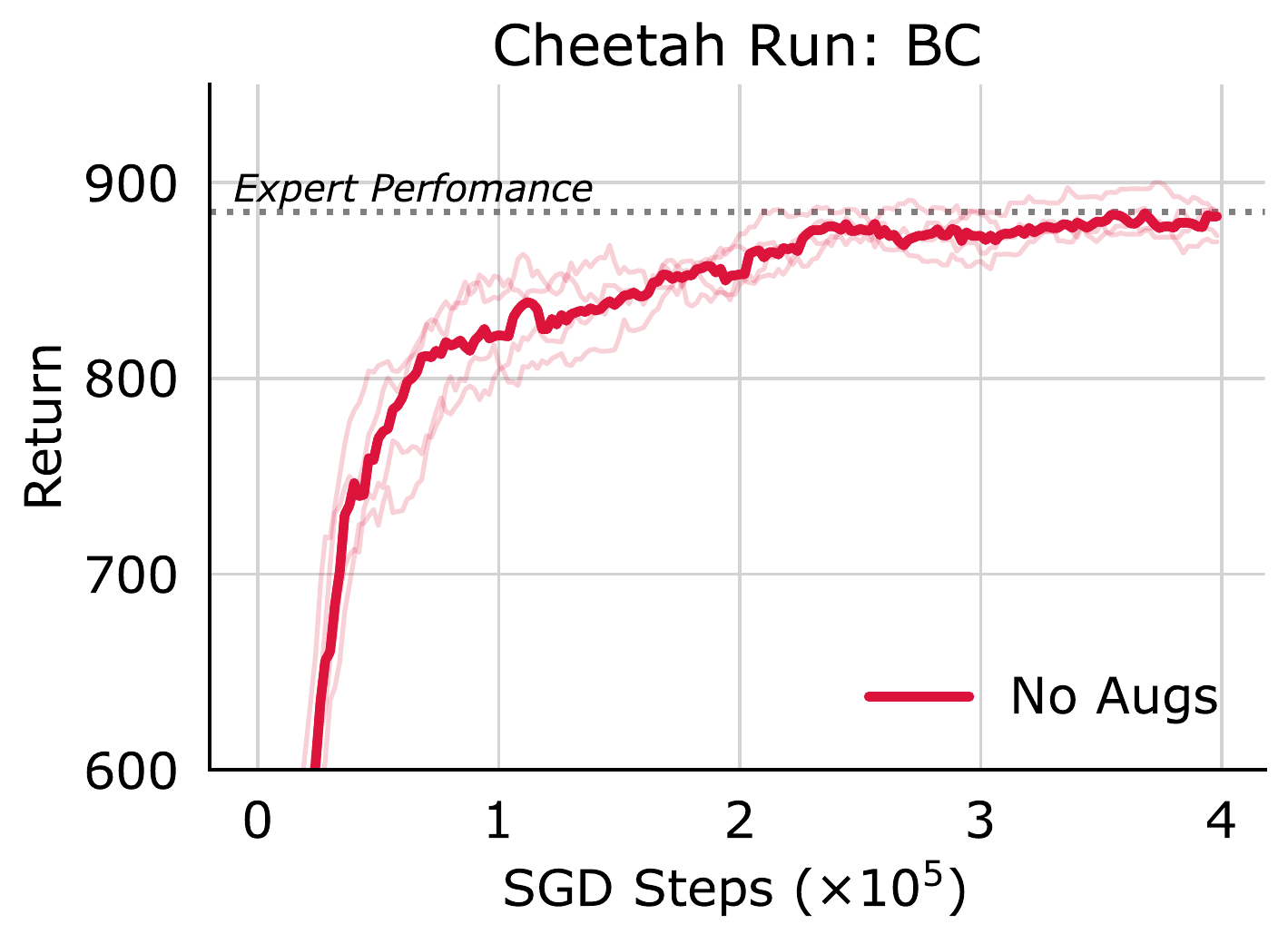} 
    \vspace{-5mm}
    \caption{\small{{Returns of agents over 5 seeds. Solid lines represent median performance, faded lines represent individual runs. The grey dotted horizontal line represents mean expert performance.}}}
    \vspace{-5mm}
    \label{fig:BC}
\end{figure}%

To illustrate that test time shift invariance is not required, we show that it is possible to learn a policy through supervised learning. To do this, we generate a pixel-based dataset of 500,000 timesteps under an expert policy in Cheetah Run, and jointly train a CNN encoder and policy using behavior cloning/supervised learning by minimizing the loss $\mathcal{L} = (a - \pi(o))^2$ until convergence, where $o$ follows the stacked frame image inputs of \cite{dqn}. We see that the pixel-based policy performs as well as the behavior agent, despite using both higher dimensional data and fewer than half the samples compared to existing expert offline RL benchmarks from \emph{proprioceptive} states \cite{d4rl}.

This provides clear evidence that shift invariance is not required at test time, and motivates us to find an alternative explanation for why random shift augmentations help the learning process in TD-learning. An alternative perspective is that when the learning signal is strong, as is the case for supervised learning (and later stages during online learning when target values are more accurate), the natural bias of CNNs to learn lower order representations acts as an implicit regularizer \cite{cnnspectralbias} that results in test-time generalization.

\subsection{Turning Off Augmentations}
\label{sec:turnoffaug}
\begin{figure}[H]
    \vspace{-2mm}
    \centering
    \includegraphics[width=0.6\linewidth]{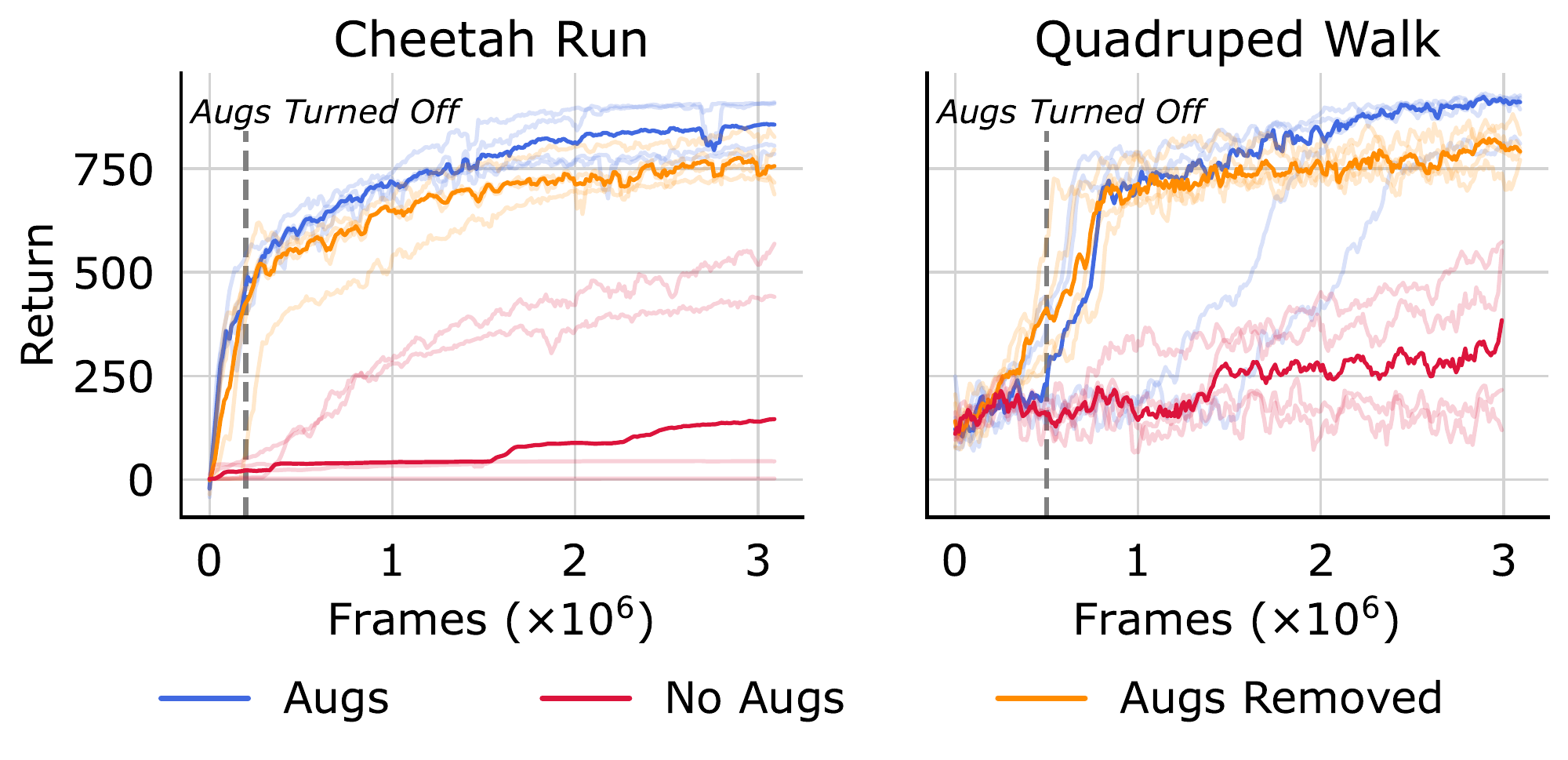} 
    \vspace{-6mm}
    \caption{\small{{Returns of agents over 5 seeds. Solid lines represent median performance, faded lines represent individual runs. The grey dashed line shows when augmentations are turned off.}}}
    \vspace{-3mm}
    \label{fig:turnoffaug2}
\end{figure}

We present more evidence showing that augmentations benefit learning the most at the beginning of training. In Fig.~\ref{fig:turnoffaug2} we show the effect of turning off augmentations at 200,000 steps in Cheetah Run, and at 500,000 in Quadruped Walk. In both instances, we see large improvements over not augmenting at all, and both nearly converge to the same value as DrQ-v2, showing further evidence that stability initially in learning is vital. We posit that turning off augmentations here did not yield similar benefits to Fig.~\ref{fig:turnoffaug} due to the fact that there is still high-frequency information in the targets (consider that the augmentations in Cheetah Run are switched off significantly earlier) that cause a marginal amount of self-overfitting, reducing the rate of learning due to feature space degeneration.

\subsection{Action-Value Surfaces}
\label{sec:actvalsurf}
Here we show the action-value surfaces of the offline agents' critics at various tuples sampled from the data. This provides us with an intuition over the loss landscape that the policies will be optimizing during the policy improvement, as accordingly the policy under the deterministic policy gradient~\citep{dpg} updates its own weights towards maximizing the action-values defined by the critic through the chain rule:
\begin{align}
    \nabla_{\phi} J_\pi &\approx \mathbb{E}_{s\sim E}\left[ \nabla_a Q_\theta(s,a)|_{a = f_\phi(s)} \nabla_{\phi} f_\phi(s) \right]
\end{align}
where $\phi$ and $\theta$ are policy and critic weights respectively. We hypothesize that self-overfitting reduces the sensitivity of the critic to actions, discarding important information regarding the causal link between actions and expected returns. To evaluate this, we sample state-action pairs from our replay buffer, and then visualize the action-value surface by sampling two random \emph{orthogonal} direction vectors from the action space $A$. We then normalize the direction vectors to have a 2-norm of 1, and then multiply each direction vector by scalars $\alpha, \beta \in [-2,2]$ respectively. We then plot the action-value surface as a result of adding the random vectors multiplied by their respective scalars onto the action sampled from offline dataset, giving us a 3-D surface. We clip actions to $a \in [-1,1]^{|A|}$ as actions are squashed to this range in the policy through a truncated normal distribution.

\begin{figure}[H]
    \vspace{-1mm}
    \centering
    \begin{subfigure}[t]{0.43\linewidth}
        \centering
        \includegraphics[width=0.999\linewidth]{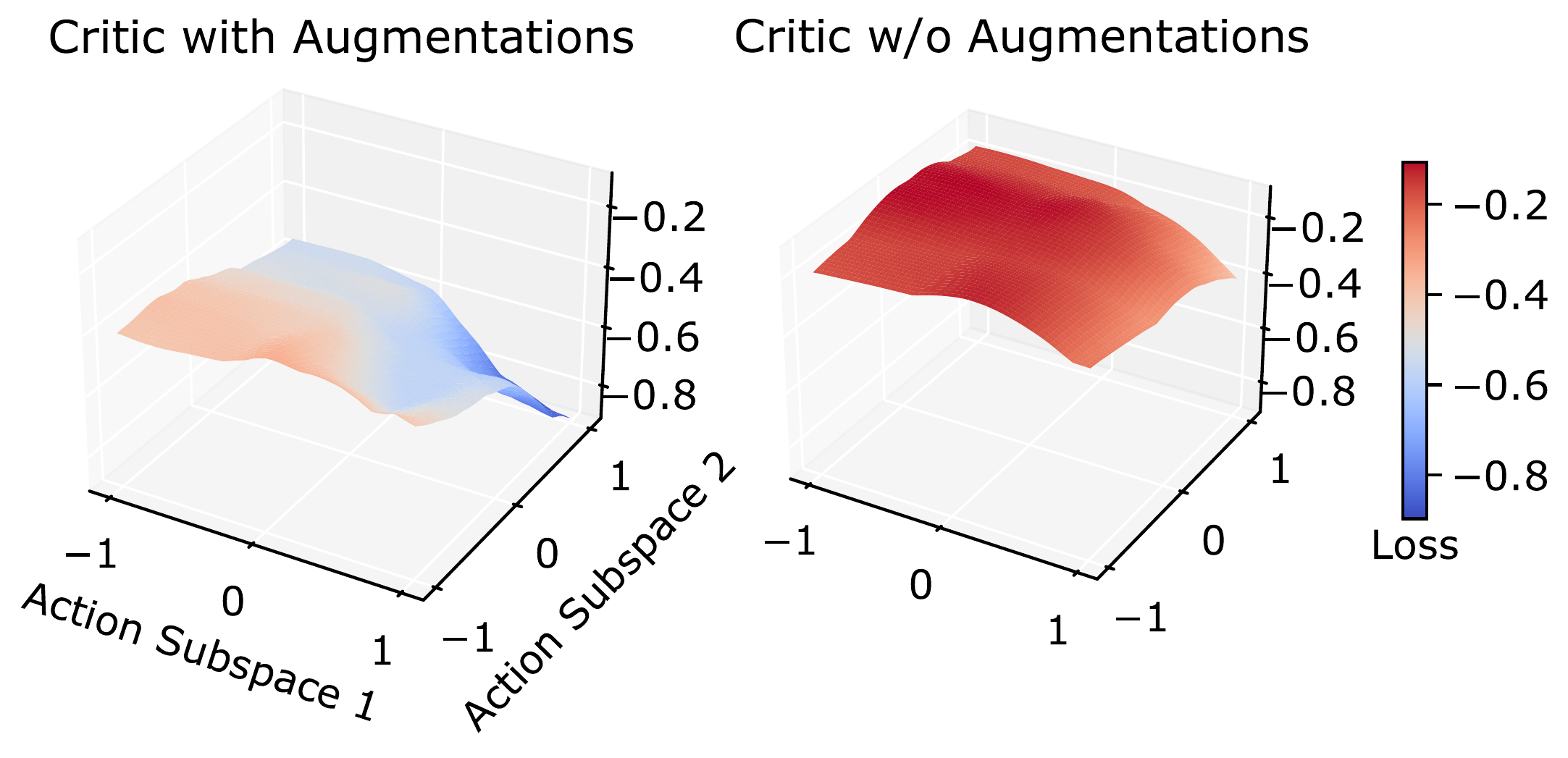} 
        \vspace{-5mm}
        \caption{\small{{Random Sampled State-Action Pair 1}}}
        % \label{fig:qvals}
    \end{subfigure}
    \hspace{6mm}
    \begin{subfigure}[t]{0.43\linewidth}
        \centering
        \includegraphics[width=0.999\linewidth]{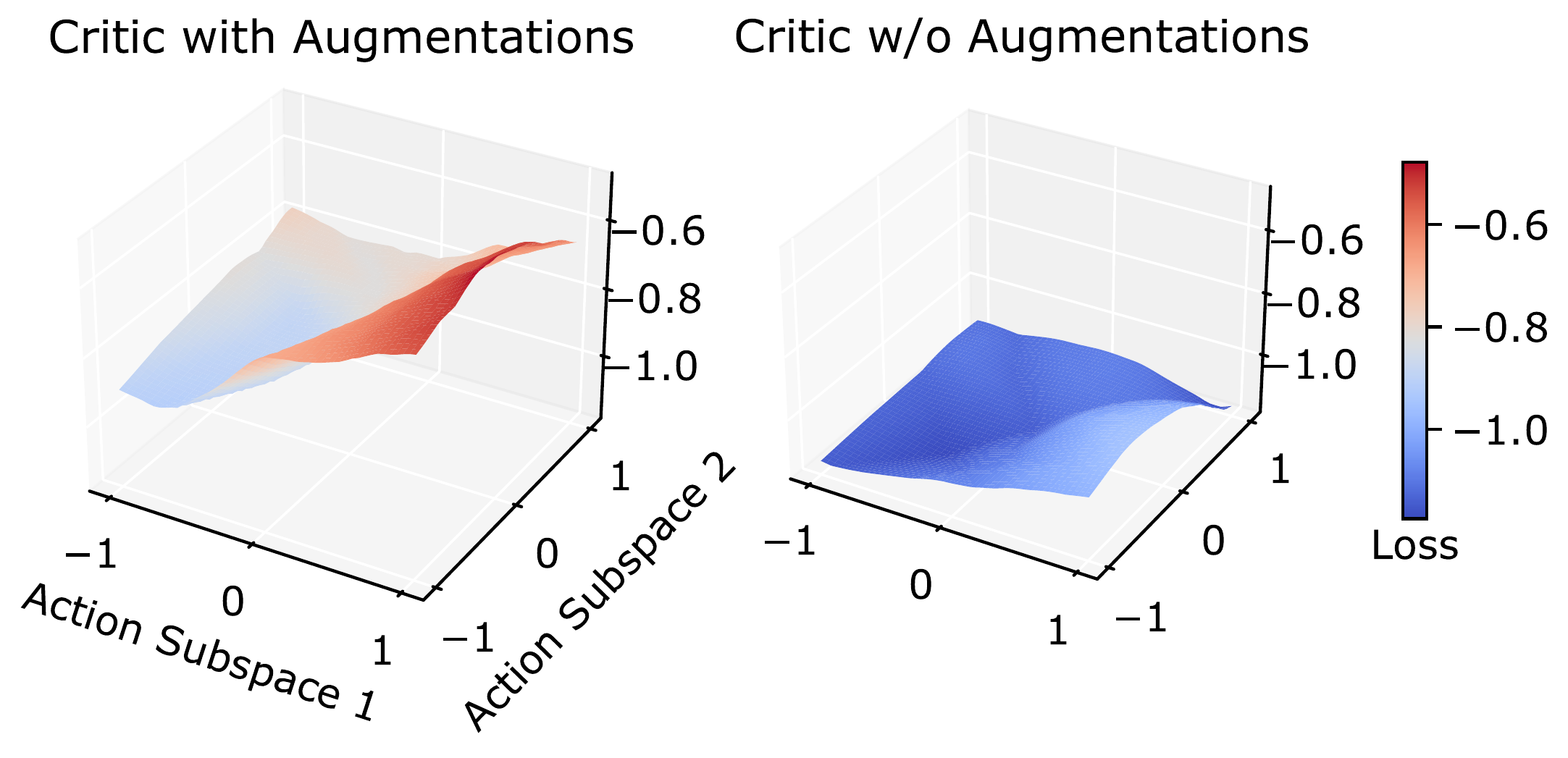} 
        \vspace{-5mm}
        \caption{\small{{Random Sampled State-Action Pair 2}}}
        % \label{fig:pearson}
    \end{subfigure}
    \begin{subfigure}[t]{0.43\linewidth}
        \centering
        \includegraphics[width=0.999\linewidth]{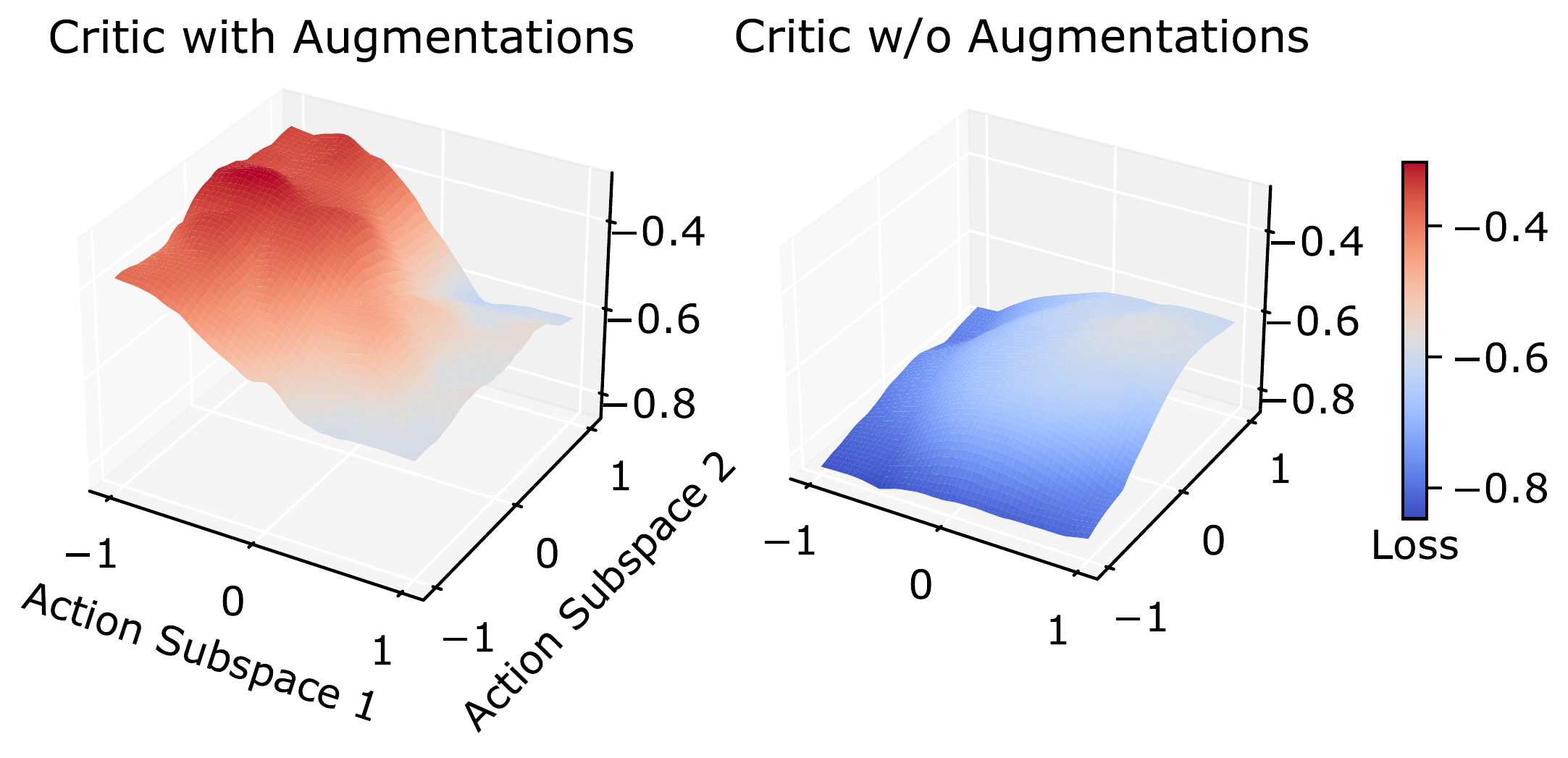} 
        \vspace{-5mm}
        \caption{\small{{Random Sampled State-Action Pair 4}}}
        \vspace{-2mm}
        % \label{fig:pearson}
    \end{subfigure}
    \hspace{6mm}
    \begin{subfigure}[t]{0.43\linewidth}
        \centering
        \includegraphics[width=0.999\linewidth]{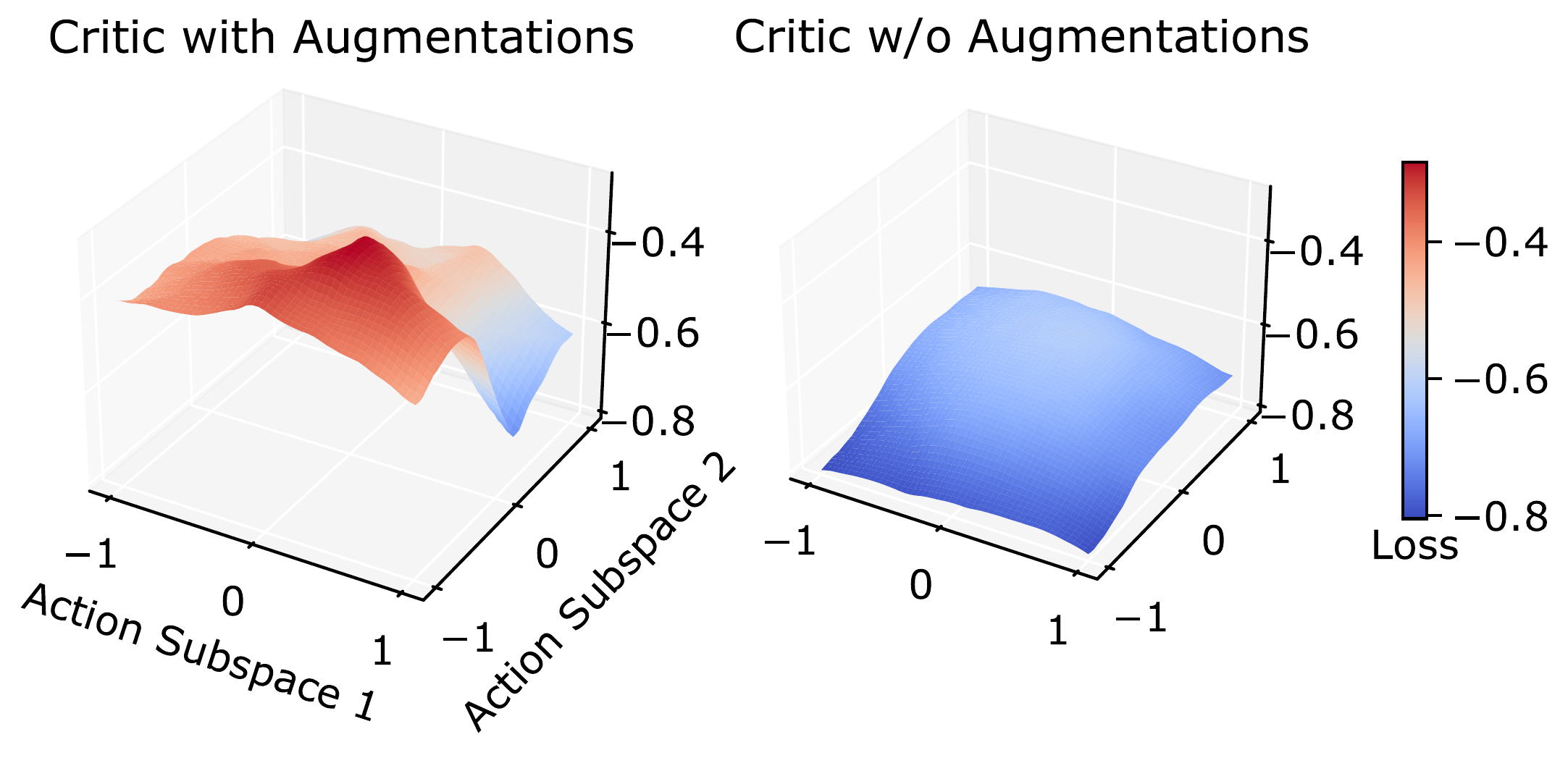} 
        \vspace{-5mm}
        \caption{\small{{Random Sampled State-Action Pair 4}}}
        \vspace{-2mm}
        % \label{fig:pearson}
    \end{subfigure}
    \vspace{-1mm}
    \caption{\small{Action-Value loss surface plotted with respect two orthogonal random directions sampled from the action space (i.e., $d_{{r}} \in A$ and $d_1 \perp d_2$).}}
    \vspace{-3mm}
\end{figure}%

We see that the critics learned by the augmented agents are more sensitive to changes in action. We believe this is due to the non-augmented agents overfitting to the observations, thus ignoring the lower-dimensional action inputs. To validate this, we sampled 128 random state-action tuples from the offline buffer, and calculated the average variance across the loss surfaces. We see a significant difference, with the augmented agent having an average loss surface variance of \textbf{0.0129}, whereas the non-augmented agent has an average loss surface variance of \textbf{0.0044}, suggestive of lower sensitivity.

\subsection{Evidence of Critic MLP Overfitting from High-Frequency Features}
\label{sec:checkerboard}
We provide further evidence that measuring high-frequency features through the ND score is vital to understanding overfit by showing how overfitting is able to occur in the fully-connected critic layers, which are usually stable under proprioceptive observations (see Table \ref{tab:offline}). To do this, we construct a pattern containing high frequency checkerboard noise $c \in \mathbb{R}^{H\times W}$, and produce as many patterns as there are channels $C$ in the final layer. To ensure consistency across each individual feature map, we normalize each checkerboard pattern by the maximum value in its respective feature map, and then divide by the width of the checkerboard. We then add this pattern multiplied by a scalar $\alpha$ onto each feature map.
\begin{figure}[H]
\vspace{-3mm}
    \centering
    \begin{subfigure}[b]{0.4\linewidth}
        \centering
        \includegraphics[width=0.999\linewidth]{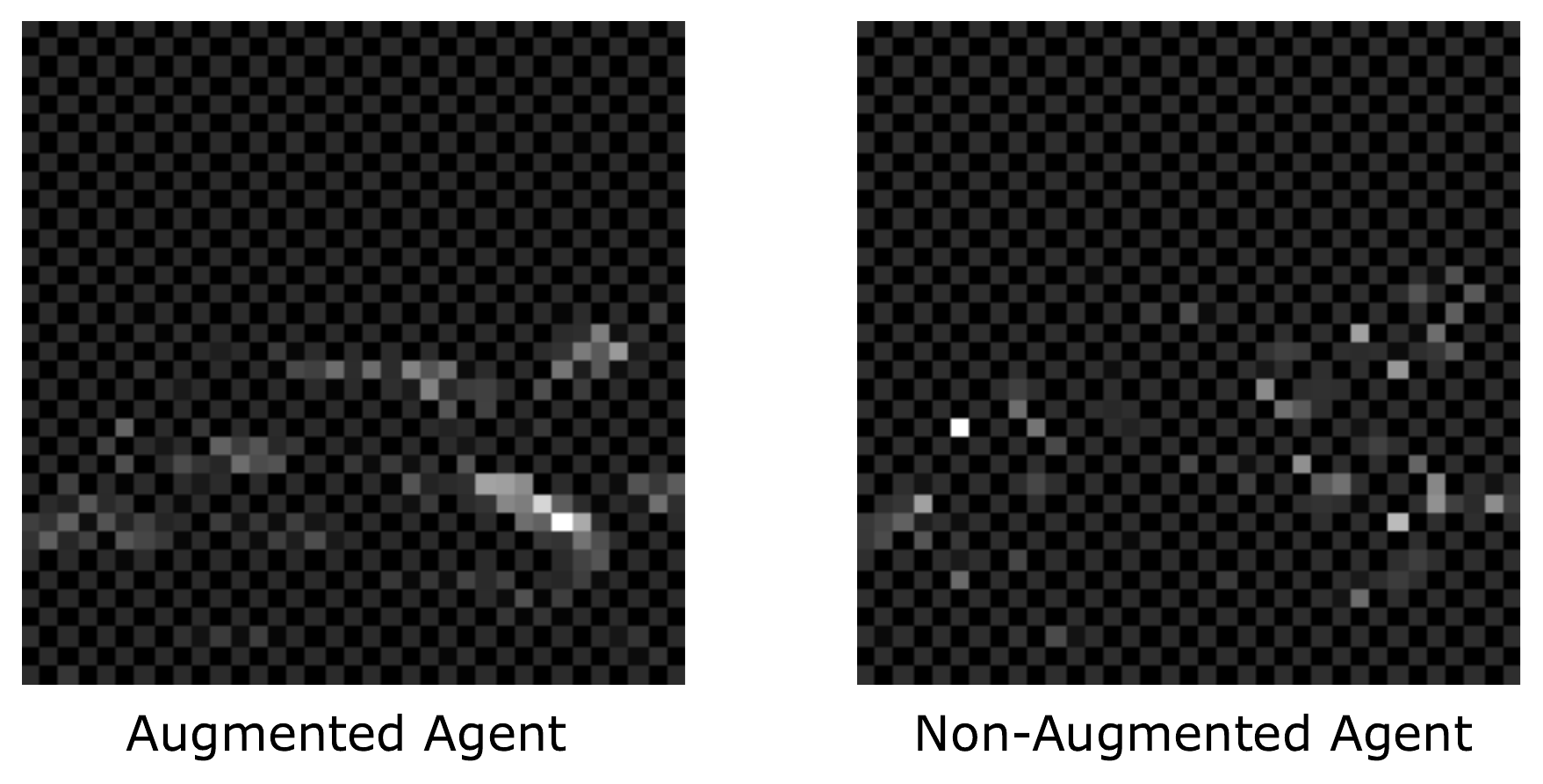} 
        \vspace{-6mm}
        \caption{\small{{Example checkerboard artefacts.}}}
        % \label{fig:qvals}
    \end{subfigure}
    \hspace{7mm}
    \begin{subfigure}[b]{0.4\linewidth}
        \centering
        \vspace{-1mm}
        \includegraphics[width=0.58\linewidth]{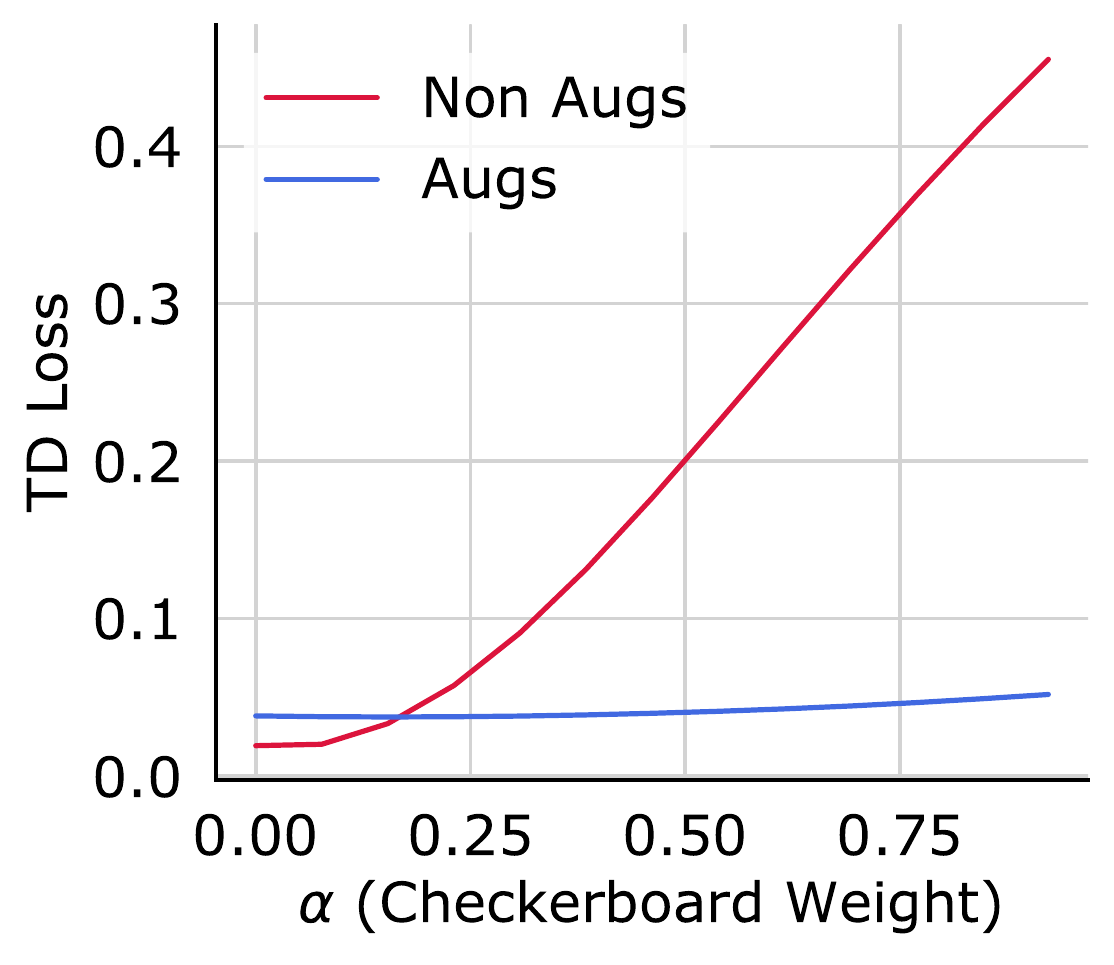} 
        \vspace{-2mm}
        \caption{\small{{Sensitivity of agents to checkerboard artifact weight}}}
        % \label{fig:pearson}
    \end{subfigure}
    \vspace{-4mm}
    \caption{\small{Effect of checkerboard artifacts on feature maps and resultant loss sensitivity. We see the non-augmented agent is significantly more sensitive to this high-frequency noise.}}
    \vspace{-5mm}
\end{figure}
As we see, the loss is significantly more sensitive to high-frequency perturbations in the non-augmented agent, justifying its reliance on high-frequency patterns in the feature maps to enable self-overfitting.

\subsection{Additional Loss Surfaces}
\label{sec:mlpsurface}
Here we show the loss surfaces of the offline agents under policy evaluation with at 1,000, 5,000, and 10,000 training steps. We also show the surfaces respect to only the MLP layers, again following the normalization approach of~\citet{visualloss}.
\begin{figure}[H]
    \vspace{-2mm}
    \centering
    \begin{subfigure}[t]{0.33\linewidth}
        \centering
        \includegraphics[width=0.999\linewidth]{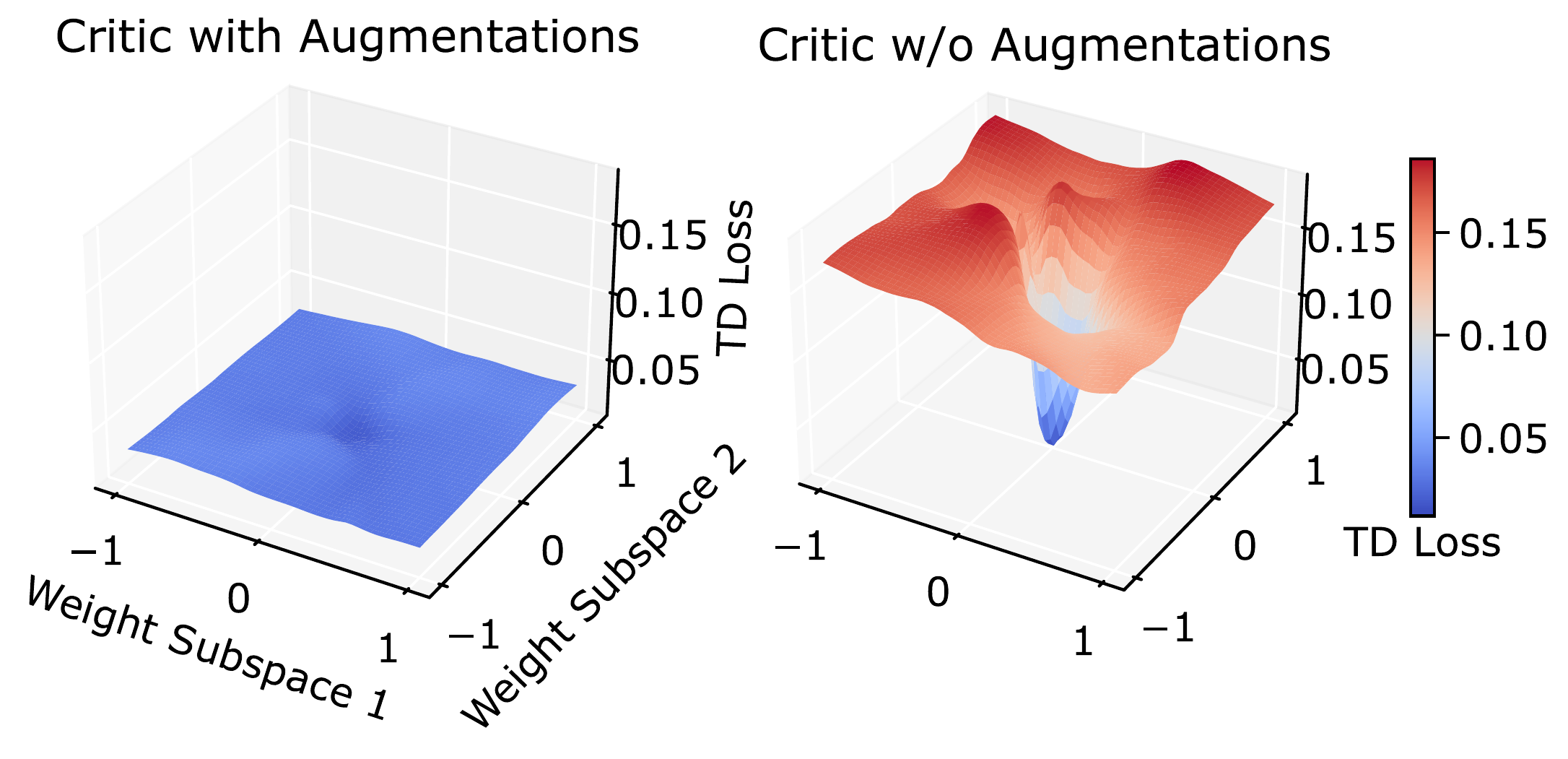} 
        \vspace{-5mm}
        \caption{\small{{1,000 SGD Steps}}}
        \vspace{-2mm}
        % \label{fig:qvals}
    \end{subfigure}
    \hfill
    \begin{subfigure}[t]{0.33\linewidth}
        \centering
        \includegraphics[width=0.999\linewidth]{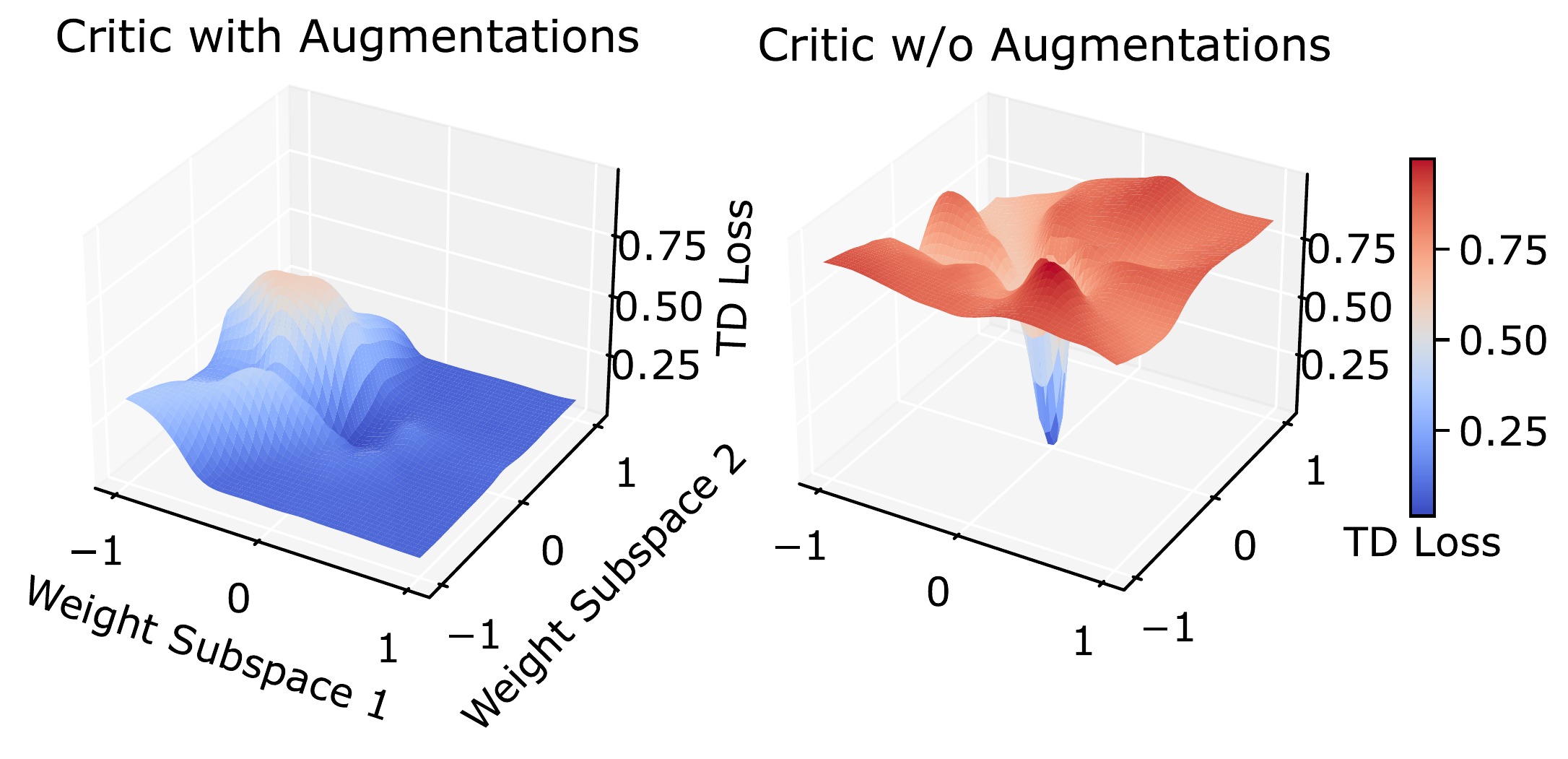} 
        \vspace{-5mm}
        \caption{\small{{5,000 SGD Steps}}}
        \vspace{-2mm}
        % \label{fig:pearson}
    \end{subfigure}
    \hfill
    \begin{subfigure}[t]{0.33\linewidth}
        \centering
        \includegraphics[width=0.999\linewidth]{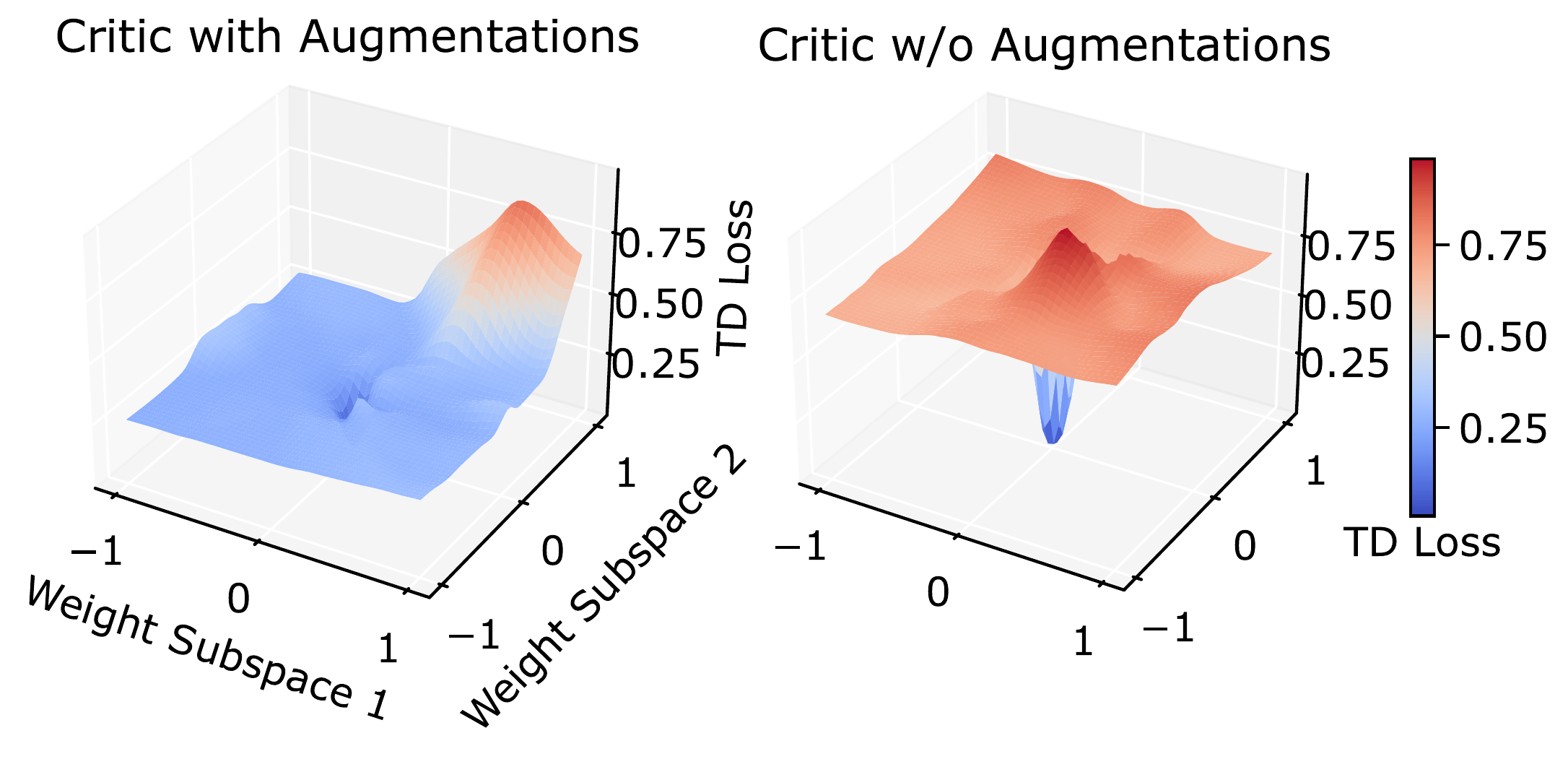} 
        \vspace{-5mm}
        \caption{\small{{10,000 SGD Steps}}}
        \vspace{-2mm}
        % \label{fig:pearson}
    \end{subfigure}
    \caption{\small{Loss surface plotted with respect to \emph{Encoder} parameters at various stages of training.}}
    \vspace{-3mm}
\end{figure}%
\begin{figure}[H]
    \vspace{-2mm}
    \centering
    \begin{subfigure}[t]{0.33\linewidth}
        \centering
        \includegraphics[width=0.999\linewidth]{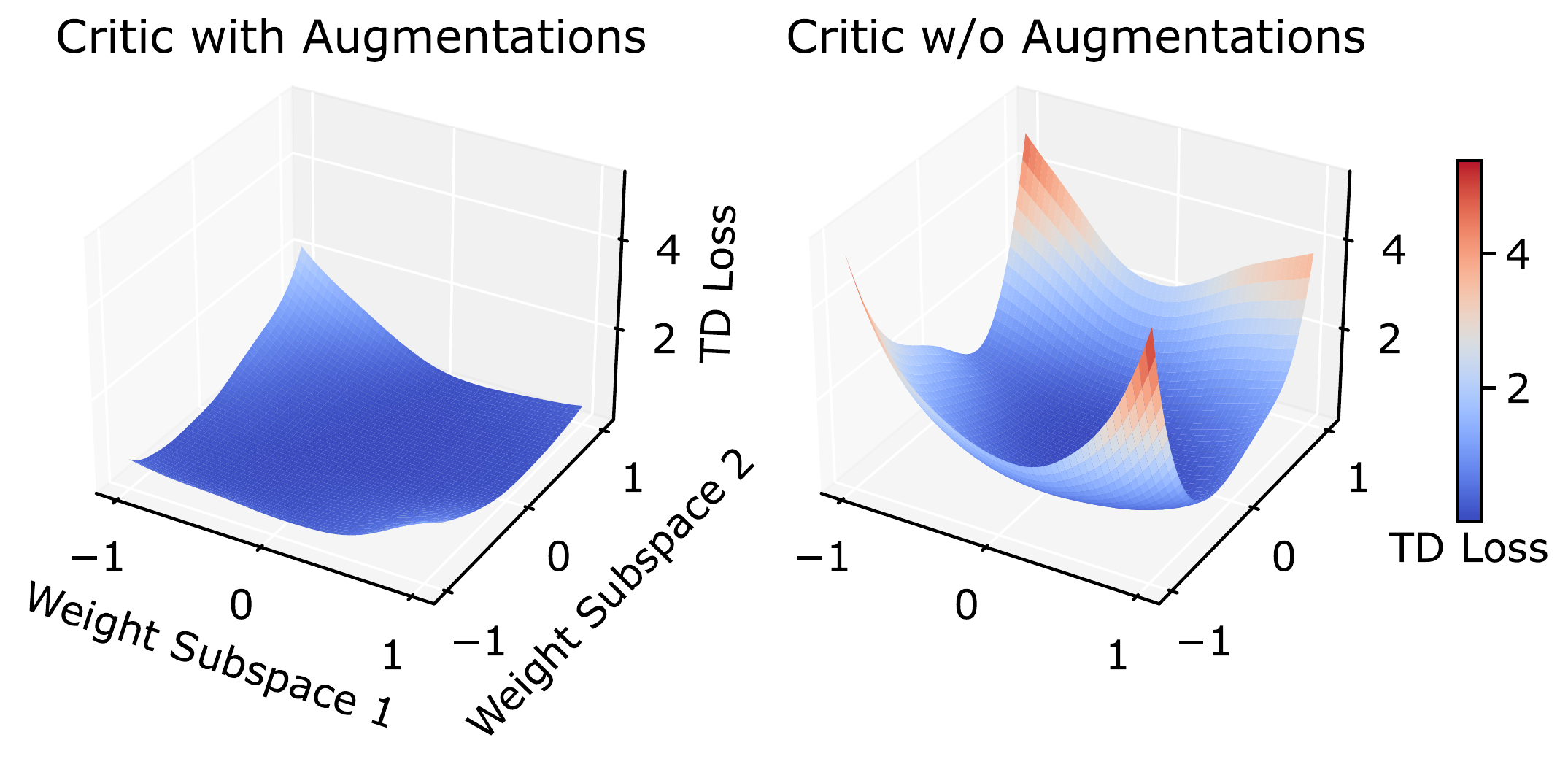} 
        \vspace{-5mm}
        \caption{\small{{1,000 SGD Steps}}}
        \vspace{-2mm}
        % \label{fig:qvals}
    \end{subfigure}
    \hfill
    \begin{subfigure}[t]{0.33\linewidth}
        \centering
        \includegraphics[width=0.999\linewidth]{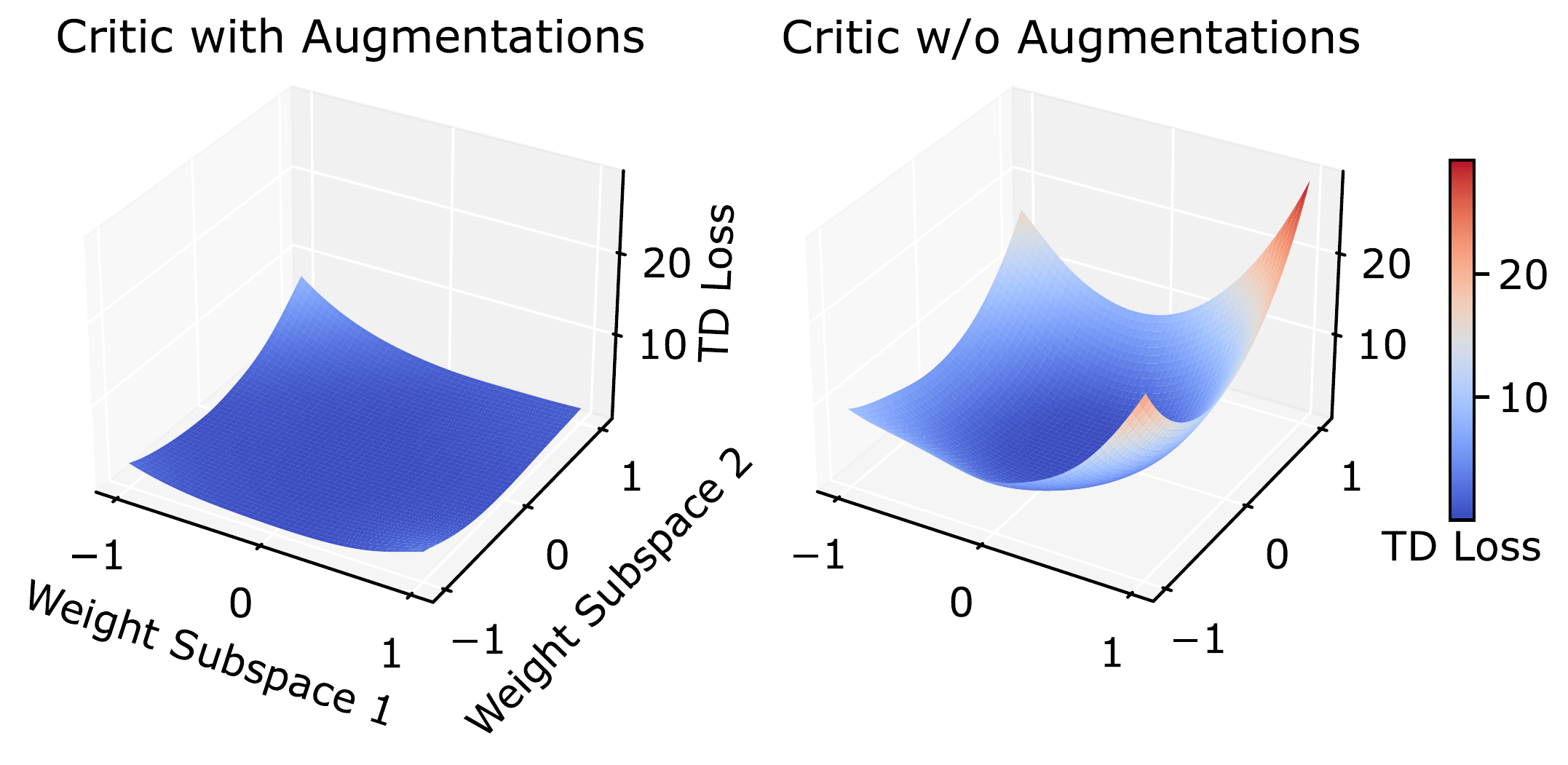} 
        \vspace{-5mm}
        \caption{\small{{5,000 SGD Steps}}}
        \vspace{-2mm}
        % \label{fig:pearson}
    \end{subfigure}
    \hfill
    \begin{subfigure}[t]{0.33\linewidth}
        \centering
        \includegraphics[width=0.999\linewidth]{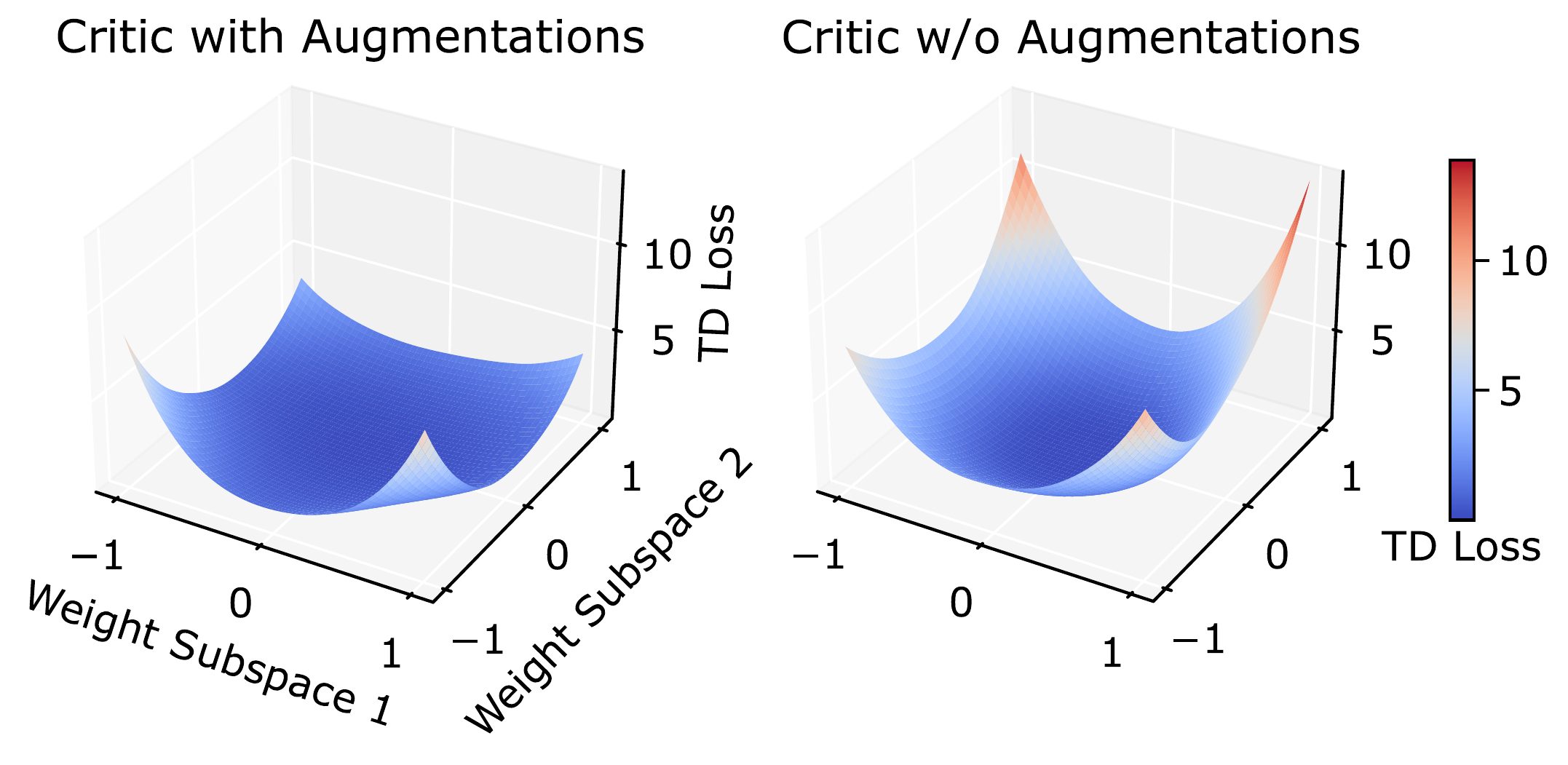} 
        \vspace{-5mm}
        \caption{\small{{10,000 SGD Steps}}}
        \vspace{-2mm}
        % \label{fig:pearson}
    \end{subfigure}
    \caption{\small{Loss surface plotted with respect to \emph{Critic MLP} parameters at various stages of training.}}
    \vspace{-5mm}
\end{figure}%
 As we see, the loss surface with respect to the MLP parameters is significantly less sharp, lending further evidence that self-overfitting is predominately a result of the flexibility of the CNN layers to learn high-frequency features.